\documentclass{article} %
\usepackage{iclr2027_conference,times}

\usepackage{amsmath,amsfonts,bm}

\def\eqref#1{equation~\ref{#1}}
\def\1{\bm{1}}

\DeclareMathAlphabet{\mathsfit}{\encodingdefault}{\sfdefault}{m}{sl}
\SetMathAlphabet{\mathsfit}{bold}{\encodingdefault}{\sfdefault}{bx}{n}

\usepackage{hyperref}
\usepackage{url}
\usepackage{xspace}
\usepackage{enumitem}
\usepackage{graphicx}
\usepackage{wrapfig}
\usepackage{placeins}
\usepackage{float}
\usepackage{subcaption}
\usepackage{booktabs}
\usepackage{array} 
\usepackage{xcolor}   %
\usepackage{pifont}   %
\usepackage{longtable}
\usepackage{placeins}
\usepackage{tabularx}
\usepackage{amsmath}
\usepackage{cleveref}
\usepackage{multirow}

\newcommand{\method}[0]{\textsc{NodeGround}\xspace}

\newcommand{\mypar}[1]{\textbf{#1}\hspace{4pt}}

\usepackage{acronym}
\acrodef{GNN}[GNN]{graph neural network}
\acrodef{HP}[HP]{hyper-parameter}
\acrodef{GT}[GT]{graph transformer}
\acrodef{MLP}[MLP]{multi-layer perceptron}
\acrodef{GFM}[GFM]{graph foundation model}
\acrodef{TFM}[TFM]{tabular foundation model}
\acrodef{ICL}[ICL]{in-context learning}
\acrodef{PFN}[PFN]{prior-data fitted network}

\title{\fontsize{15}{19}\selectfont\method: A Node Classification Benchmark\\in the Graph Foundation Model Era}

\author{
Jinmo Lee$^{1}$\thanks{These authors contributed equally.}\quad
Dooho Lee$^{1}$\footnotemark[1]\quad
Minho Jeong$^{1}$\footnotemark[1]\quad
Jaemin Yoo$^{1}$\\
$^{1}$Nums AI\\
\texttt{\{jinmo.lee, dooho, minho.jeong, jaemin\}@nums.world}
}

\iclrfinalcopy %
\begin{document}

\maketitle
\lhead{Preprint}

\begin{abstract}
Can a pretrained graph model replace training and tuning a separate predictor for each dataset?
Answering this requires evaluating prediction quality alongside computational cost.
We present \method{}, a node classification benchmark that puts graph foundation models (GFMs) and dataset-specific supervised learning under a common evaluation framework.
The benchmark spans 51 datasets and evaluates six GFMs alongside 15 supervised methods under two label-availability regimes.
Shared data partitions, validation-only model selection, controlled hyperparameter searches, and multiple predictive metrics make comparisons systematic, while workflow measurements account for adaptation, training, tuning, and inference.
The results favor carefully tuned graph neural networks overall.
GraphPFN reaches third place by Elo when more labels are available, yet its relative strengths vary substantially with dataset properties.
Efficiency comparisons further qualify the benefits of pretrained reuse: GVT and GraphPFN appear on the Pareto frontiers when supervised methods are represented by their default and fully tuned configurations.
Adding intermediate tuning budgets removes this advantage for GVT and leaves GraphPFN extending the estimated frontier in the label-rich setting alone.
Thus, reusing pretrained parameters does not yet provide a broadly reliable route to either stronger predictions or cheaper workflows.
We release the evaluation pipeline, run-level records, and an open leaderboard at \url{https://github.com/nums-ai/nodeground}.
\end{abstract}

\section{Introduction}

\begin{wrapfigure}[19]{r}{0.45\textwidth}
  \centering
  \vspace{-14pt}
  \includegraphics[width=\linewidth]{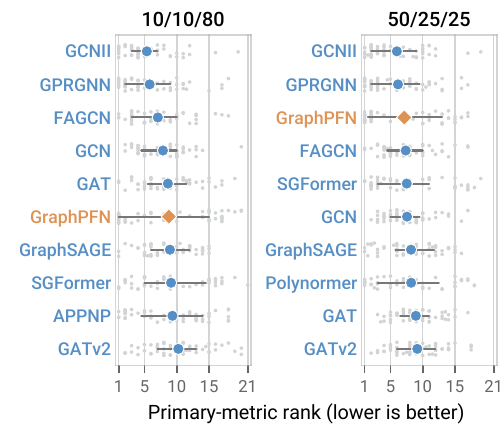}
  \caption{\textbf{Top ten methods under two label splits,}
  ranked over 51 datasets using binary AUROC and multiclass accuracy.
    Gray points show dataset ranks; lines show IQRs.
  }
  \label{fig:intro-results}
\end{wrapfigure}

Graph neural networks (GNNs) learn node representations by propagating and aggregating information along graph edges~\citep{kipf2016semi,hamilton2017inductive}.
They have shown strong performance across diverse domains, including social, scholarly, commerce, and transportation networks~\citep{hu2020open, bazhenov2026graphland, liang2026towards}.
Although this success has made GNNs a dominant approach to node classification,
achieving strong performance on a new graph typically requires training from scratch and extensive hyperparameter tuning, which demand substantial time and effort.

Recent \acfp{GFM} aim to reduce this repeated cost by transferring pretrained knowledge to new graphs with minimal adaptation~\citep{zhao2025fully,lee2025view,lee2026node4all}.
Existing studies report promising gains of \acp{GFM} over GNNs~\citep{choi2026learning,eremeev2025turning,eremeev2025graphpfn}, but the consistency and cost-effectiveness of those gains remain unclear.
It depends on two questions whether \acp{GFM} can serve as genuine alternatives to supervised methods trained and tuned separately for each dataset: (i) how consistently \acp{GFM} match or outperform supervised methods across diverse graph datasets, and (ii) how their adaptation cost compares with the cost of training and tuning a supervised method.

To answer these questions, we need a large systematic benchmark that compares \acp{GFM} with supervised methods across diverse datasets with controlled cost measurements.
Node classification is a natural setting for this comparison, as it has long served as a symbolic task in graph learning and requires models to fully leverage both features and graph topology.
However, existing benchmarks remain fragmented across dataset suites~\citep{hu2020open,pei2020geom,zhu2020beyond} and differ in data splits, implementations, and hyperparameter tuning~\citep{shchur2018pitfalls,errica2019fair}.
This leaves no unified framework that combines broad dataset coverage, controlled evaluation, and computational cost measurement to assess the performance--cost trade-offs.

To address these gaps, we introduce \method, a comprehensive node classification benchmark for evaluating \acp{GFM} with strong supervised methods.
Our contributions are summarized as follows:

\begin{itemize}[leftmargin=*, labelsep=0.6em, itemsep=0.15em, topsep=0pt,partopsep=0pt,
    parsep=0pt,]

\item \textbf{Broad benchmark coverage.} We curate 51 datasets across six domains, spanning diverse graph structures and feature specifications, and evaluate 15 supervised methods with six recent \acp{GFM}.

\item \textbf{Joint performance--cost measurement.} We jointly measure predictive performance along with the wall-clock time required to adapt each GFM and to train and tune each supervised method.

\item \textbf{Standardized evaluation protocol.} For every dataset, we evaluate all methods under two data-split regimes of 10/10/80 and 50/25/25, each comprising five fixed splits. All evaluations use documented method implementations, predefined hyperparameter search spaces, equal configuration-search budgets, and uniform model-selection criteria.

\end{itemize}
We further release the evaluation pipeline, exact data splits, raw results, and selected hyperparameters for each repeat.
These resources support reproducibility and enable experiments and analyses beyond those reported here.
To support continued GFM development, we also provide a leaderboard and submission system for evaluating new methods under the same protocol.

\mypar{Main findings.}
As shown in \Cref{fig:intro-results}, GCNII~\citep{chen2020simple} and GPRGNN~\citep{chien2020adaptive} lead the mean-rank comparison under both splits.
GraphPFN~\citep{eremeev2025graphpfn} is the strongest GFM, rising from sixth under 10/10/80 to third under 50/25/25  (\Cref{sec:rq1-overall-comparison}).
However, these leading methods exhibit complementary strengths across data properties (\Cref{sec:rq2-gfm-generality}).
GVT~\citep{lee2025view} and GraphPFN lie on the performance--cost Pareto frontier in both regimes, but
including intermediate tuning budgets leaves GraphPFN as the sole GFM on the frontier (\Cref{sec:rq3-gfm-efficiency}).
Current \acp{GFM} thus offer limited practical advantages, motivating future works to deliver stronger predictions across diverse graphs within the same budget as supervised training and tuning.

\section{Background}
\label{sec:related-work}

With foundation models, node classification has begun shifting from training a separate model for each graph to reusing a single pretrained model across graphs.
We review this shift, examine gaps in current evaluations, and identify key requirements for evaluating this emerging paradigm.

\mypar{Supervised node classification.}
Supervised \acp{GNN} have long dominated node classification and evolved through architectural advances in
neighborhood aggregation~\citep{kipf2016semi,hamilton2017inductive},
attention-based message passing~\citep{velivckovic2017graph,brody2021attentive},
decoupled transformation and propagation~\citep{gasteiger2018predict,wu2019simplifying,chien2020adaptive},
and scalable graph-wide interaction~\citep{wu2022nodeformer,wu2023sgformer,chen2022nagphormer,deng2024polynormer}.
However, strong performance typically requires training and extensive hyperparameter tuning for each graph dataset.
This repeated optimization motivates \acp{GFM}, which seek to replace per-dataset learning with pretrained knowledge reusable across graphs.

\mypar{Graph foundation models.}
\acp{GFM} move per-dataset learning into reusable pretrained components, with different expectations on \emph{what must be learned} from each new dataset.
GraphAny~\citep{zhao2025fully} fits LinearGNN predictors per graph and combines their predictions using a pretrained model. 
GVT~\citep{lee2025view} and Node4All~\citep{lee2026node4all} 
use pretrained encoders to produce node representations, on which a lightweight predictor is trained.
A different approach comes from tabular learning.
Prior-data fitted networks (PFNs)~\citep{muller2021transformers} are pretrained on synthetic tables to predict from labeled examples as context, without updating model parameters on the target dataset.
G2T-FM~\citep{eremeev2025turning} extends this paradigm to graphs by adding structural information to node features before passing them to a pretrained PFN. 
NodePFN~\citep{choi2026learning} and GraphPFN~\citep{eremeev2025graphpfn} incorporate graph structure more directly to the PFN architecture through graph-specific architectures and synthetic graph priors. 

\mypar{Limitations of current evaluations.}
Although these methods report matching or outperforming supervised \acp{GNN} in their original evaluations, the evidence remains limited in coverage or baseline strength.
NodePFN~\citep{choi2026learning} uses only graphs with fewer than 50,000 nodes, while G2T-FM~\citep{eremeev2025turning} and GraphPFN~\citep{eremeev2025graphpfn} consider fewer than 500 input features.
GraphAny~\citep{zhao2025fully} and NodePFN~\citep{choi2026learning} compare against only two \acp{GNN}; GVT~\citep{lee2025view} and Node4All~\citep{lee2026node4all} cover more baselines but omit the residual connection and layer normalization, which can strengthen \acp{GNN}~\citep{luo2024classic}. 

The computational advantages of \acp{GFM} also remain unclear, although reducing per-dataset adaptation costs is a central motivation for \acp{GFM}. 
Some methods require target-specific preprocessing or search~\citep{choi2026learning, lee2025view}, while dense interactions among labeled nodes can make PFN-based inference costly on large graphs~\citep{choi2026learning, eremeev2025graphpfn}. 
These costs have not been systematically compared with the training and hyper-parameter tuning costs of supervised \acp{GNN}.
Existing evaluations therefore leave unclear whether pretrained reuse offers consistent predictive gains at competitive computational costs over supervised methods.

\begin{table*}[t]
\begingroup

\newcommand{\ngyes}{%
  \textcolor{green!60!black}{\ding{109}}}
\newcommand{\ngpartial}{%
  \textcolor{orange!90!black}{%
    \raisebox{0.05ex}{\scalebox{0.9}{$\triangle$}}}}
\newcommand{\ngno}{%
  \textcolor{red!80!black}{\ensuremath{\boldsymbol{\times}}}}
\newcommand{\ngcost}[3]{%
  \makebox[1.25em][c]{#1}%
  \makebox[1.25em][c]{#2}%
  \makebox[1.25em][c]{#3}}

\caption{\textbf{Comparison of node-classification benchmarks and evaluation protocols.}
Columns correspond to the five criteria given in Section \ref{sec:related-work}.
Under \textbf{Cost accounting}, the three symbols indicate reporting of training, tuning, and inference costs, respectively.
\ngyes, \ngpartial, and \ngno\ denote full, partial, and no documented support.
A dagger marks an evaluation protocol rather than a benchmark.
Appendix~\ref{app:benchmark-comparison-evidence} defines the criteria and documents the evidence for each assessment.}
\label{tab:benchmark-comparison}

\centering
\small
\setlength{\tabcolsep}{1.5pt}
\renewcommand{\arraystretch}{1.10}

\resizebox{\textwidth}{!}{%
\begin{tabular}{@{}lcccccccc@{}}
\toprule
&
\multicolumn{1}{c}{\textbf{Dataset coverage}}
&
\multicolumn{2}{c}{\textbf{Comparison control}}
&
\multicolumn{2}{c}{\textbf{Eval. robustness}}
&
\multicolumn{1}{c}{\textbf{Cost accounting}}
&
\multicolumn{2}{c}{\textbf{Transparency}}
\\
\cmidrule(lr){2-2}
\cmidrule(lr){3-4}
\cmidrule(lr){5-6}
\cmidrule(lr){7-7}
\cmidrule(lr){8-9}

\textbf{Benchmark}
&
\shortstack{\textbf{Number of}\\\textbf{datasets}}
&
\shortstack{\textbf{HPO}\\\textbf{control}}
&
\shortstack{\textbf{Val.}\\\textbf{selection}}
& 
\makebox[5em][c]{\shortstack{\textbf{Data}\\\textbf{splits}}}
& 
\makebox[5em][c]{\shortstack{\textbf{Multi-}\\\textbf{metric}}}
&
\shortstack{\textbf{Compute}\\\textbf{cost}}
&
\shortstack{\textbf{Raw}\\\textbf{results}}
&
\shortstack{\textbf{Leader-}\\\textbf{board}}
\\
\midrule

\cite{yang2016revisiting}$^{\dagger}$
& 4 & \ngno & \ngyes & 1 & \ngno
& \ngcost{\ngno}{\ngno}{\ngno} & \ngno & \ngno \\

\cite{shchur2018pitfalls}$^{\dagger}$
& 8 & \ngyes & \ngpartial & 100 & \ngno
& \ngcost{\ngno}{\ngno}{\ngno} & \ngno & \ngno \\

\cite{pei2020geom}$^{\dagger}$
& 9 & \ngyes & \ngyes & 10 & \ngno
& \ngcost{\ngyes}{\ngno}{\ngno} & \ngno & \ngno \\

\cite{platonov2026fair}$^{\dagger}$
& 4 & \ngyes & \ngyes & 1 & \ngno
& \ngcost{\ngyes}{\ngyes}{\ngyes} & \ngpartial & \ngno \\

\midrule

\cite{you2020design}
& 18 & \ngyes & \ngpartial & 3 & \ngno
& \ngcost{\ngyes}{\ngpartial}{\ngpartial} & \ngno & \ngno \\

\cite{hu2020open}
& 5 & \ngno & \ngyes & 1 & \ngno
& \ngcost{\ngno}{\ngno}{\ngno} & \ngno & \ngyes \\

\cite{lim2021large}
& 7 & \ngpartial & \ngyes & 5 & \ngno
& \ngcost{\ngpartial}{\ngno}{\ngno} & \ngno & \ngno \\

\cite{dwivedi2023benchmarking}
& 3 & \ngpartial & \ngpartial & $\{1,20\}$ & \ngno
& \ngcost{\ngyes}{\ngno}{\ngno} & \ngno & \ngno \\

\cite{platonov2023critical}
& 5 & \ngpartial & \ngyes & 10 & \ngno
& \ngcost{\ngno}{\ngno}{\ngno} & \ngno & \ngno \\

\cite{luo2024classic}
& 18 & \ngyes & \ngyes & $\{1,10\}$ & \ngno
& \ngcost{\ngno}{\ngno}{\ngno} & \ngno & \ngno \\

\cite{bazhenov2026graphland}
& 7 & \ngpartial & \ngyes & $\{2,3\}$ & \ngno
& \ngcost{\ngpartial}{\ngno}{\ngno} & \ngno & \ngno \\

\cite{yu2026evaluating}
& 15 & \ngpartial & \ngno & 50 & \ngyes
& \ngcost{\ngpartial}{\ngno}{\ngpartial} & \ngno & \ngno \\

\midrule

\textbf{\method}
& \textbf{51} & \ngyes & \ngyes & \textbf{5} & \ngyes
& \ngcost{\ngyes}{\ngyes}{\ngyes} & \ngyes & \ngyes \\

\bottomrule
\end{tabular}%
}

\endgroup
\end{table*}

\mypar{Node-classification benchmarks.}
Existing node-classification benchmarks were primarily designed to compare and assess the power of supervised \acp{GNN}~\citep{shchur2018pitfalls,you2020design,luo2024classic} or to include diverse graph datasets~\citep{pei2020geom,lim2021large,hu2020open,platonov2023critical,bazhenov2026graphland}.
Evaluating \acp{GFM} poses a different question: whether pretrained reuse works broadly, remains competitive with strong supervised methods, and reduces per-dataset computation.
Answering this requires
(i)~\textbf{dataset coverage} across diverse graphs and domains,
(ii)~\textbf{comparison control} through consistent tuning and model selection,
(iii)~\textbf{evaluation robustness} across splits and predictive metrics,
(iv)~\textbf{computational cost} covering training, tuning, adaptation, and inference, and
(v)~\textbf{transparency} through released run-level results.

As summarized in \Cref{tab:benchmark-comparison}, no benchmark in our comparison covers all five aspects.
\method{} brings them together with 51 datasets, two label regimes, controlled training and tuning, repeated multi-metric evaluation, comprehensive cost measurement, and released run-level results.
Detailed definitions, scoring rules, and evidence for each entry are provided in Appendix~\ref{app:benchmark-comparison-evidence}.
\section{\method Dataset Collection}
\label{sec:dataset-collection}

\method focuses on transductive, single-label node classification on small- and medium-scale homogeneous attributed graphs.
Directed and temporal sources are converted into static, undirected graphs.
Multilabel classification, heterogeneous graphs, and web-scale training are outside the current scope.
In this section, we describe how we curate 51 datasets within this scope (\Cref{sec:dataset-selection}) and characterize their structural, feature, and label properties (\Cref{sec:data-regime-coverage}).

\subsection{Dataset Selection Protocol}
\label{sec:dataset-selection}

We collect 117 candidate datasets from four sources: datasets available through PyTorch Geometric (PyG)~\citep{fey2019fast}, the Open Graph Benchmark (OGB)~\citep{hu2020open}, the heterophilous graph collection introduced by \citet{platonov2023critical},\footnote{\url{https://github.com/yandex-research/heterophilous-graphs}} and the Text-Attributed-Graphs (TAG) collection introduced by \citet{wang2025generalization}.\footnote{\url{https://huggingface.co/datasets/Graph-COM/Text-Attributed-Graphs}}

\begin{wrapfigure}{r}{0.37\textwidth}
    \vspace{-8.5pt}
    \centering
    \includegraphics[width=\linewidth]{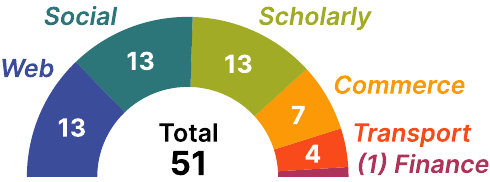}
    \vspace{-12pt}
    \caption{Application domains of the 51 retained datasets.}
    \label{fig:dataset-composition}
    \vspace{-12pt}
\end{wrapfigure}

\mypar{Selection criteria.}
We screen candidates using five sequential checks. \textbf{(1)} The task must be node classification on a homogeneous graph. \textbf{(2)} Each labeled node must have a single class label. \textbf{(3)} The dataset must provide node attributes that are neither identity-only, topology-only, nor directly derived from target labels. \textbf{(4)} The graph and features must meet the specified scale limits. \textbf{(5)} The dataset must not duplicate one already retained. These checks successively reduce the pool to 101, 94, 79, 67, and 51 candidates.

As shown in \Cref{fig:dataset-composition}, the retained datasets span six application domains.
Web, social, and scholarly datasets are equally represented, with 13 datasets each, alongside smaller collections from commerce (7), transport (4), and finance (1).
No single domain accounts for more than approximately one quarter of the collection.
Appendix~\ref{app:candidate-ledger} details the selection criteria, stage-wise screening outcomes, and how duplicate datasets were handled. Appendix~\ref{app:dataset-catalog} lists each retained dataset's source, domain, prediction target, citation, and upstream license evidence.

\subsection{Coverage of Dataset Properties}
\label{sec:data-regime-coverage}

Having established the sources and domains represented in \method, we examine the range of its structural, feature, and target
properties and how these properties combine within each dataset.

\begin{figure*}[h]
    \centering
    \vspace{-2mm}
    \includegraphics[width=\textwidth]{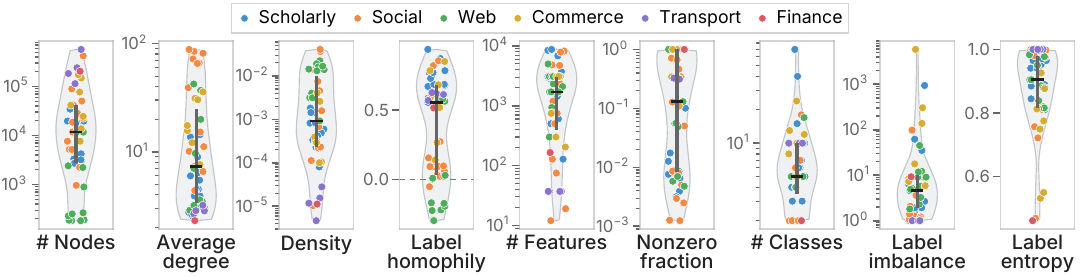}
    \vspace{-5mm}
    \caption{\textbf{Individual property coverage of \method.}
Panels summarize graph structure, node features, and targets. Each point represents one dataset, with colors indicating application domains. Gray boxes show interquartile ranges and black lines show medians.} 
    \label{fig:marginal-coverage}
\end{figure*}
\begin{figure*}[t]
    \centering
    \vspace{-2mm}
    \includegraphics[width=\textwidth]{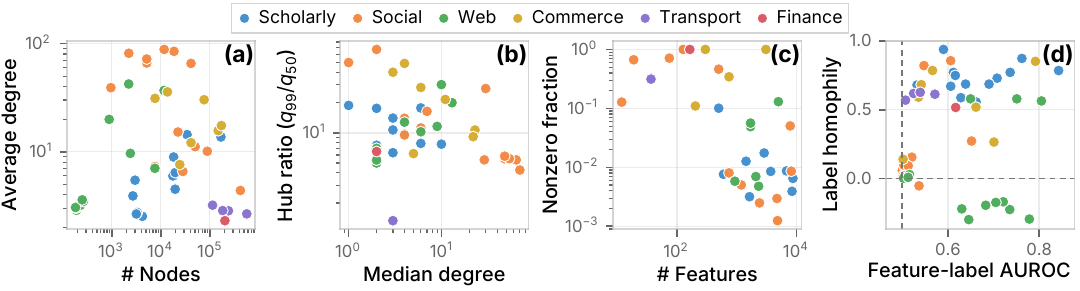}
    \vspace{-5mm}
    \caption{\textbf{Joint property coverage of \method.}
(a) Node count versus average degree.
(b) Median nonzero degree \(q_{50}\) versus hub concentration \(q_{99}/q_{50}\).
(c) Feature dimension versus nonzero fraction.
(d) Class-balanced feature-similarity AUROC versus adjusted homophily.
Dashed lines in panel~(d) mark chance-level feature separation and zero adjusted homophily.}
    \label{fig:joint-coverage}
\end{figure*}

\mypar{Individual property coverage.}
\Cref{fig:marginal-coverage} summarizes the dataset properties from left to right in terms of graph structure, node features, and prediction targets.
The structural properties span several orders of magnitude in node count, average degree, and graph density.
To measure label alignment with graph structure, we use adjusted homophily, which is the label-based form of Newman's assortativity coefficient~\citep{newman2003mixing,platonov2023characterizing}.
The resulting values range from strongly heterophilous to strongly homophilous graphs.
The feature matrices also cover both compact dense representations and high-dimensional sparse inputs.
The tasks range from binary to high-cardinality classification, with class distributions ranging from balanced to highly imbalanced.
These properties form distinct profiles across application domains.
Transportation graphs are among the largest but have low average degree and density, while scholarly graphs tend to be homophilous and have high-dimensional, sparse features.
Appendix~\ref{app:dataset-statistics} reports the exact values for every dataset.

\mypar{Joint property coverage.}
The marginal ranges above show how each property varies, but not which combinations a model encounters. \Cref{fig:joint-coverage} reveals differences that a single statistic would miss.
(a) Graphs with similar node counts can have very different average degrees, while (b) similar median degrees can hide large differences in hub concentration, measured as the ratio of the 99th-percentile to the median nonzero degree.
(c) High-dimensional features can be either sparse or dense. (d) In eight datasets, features can distinguish labels despite negative adjusted homophily.
Overall, datasets that look alike on one axis can present quite different learning conditions.
\section{Standardized Evaluation Protocol}
\label{sec:evaluation-protocol}

\method standardizes data representation, label splits, hyperparameter optimization and model selection. \Cref{tab:model-coverage} lists all evaluated methods, and \Cref{app:method-implementations} provides implementation details.

\begin{table*}[h]
\centering
\small
\caption{\textbf{Methods evaluated in \method.}
We evaluate 15 supervised models trained from scratch for each dataset and split, alongside six recent \acp{GFM}.}
\label{tab:model-coverage}
\vspace{-3pt}
\fontsize{8.5pt}{10pt}\selectfont
\setlength{\tabcolsep}{5pt}
\begin{tabularx}{\textwidth}{@{}l>{\raggedright\arraybackslash}X@{}}
\toprule
\textbf{Group} & \textbf{Methods} \\
\midrule
\begin{tabular}[t]{@{}l@{}}Supervised\\\end{tabular} & \mbox{MLP (feature-only)},
  \mbox{GCN~\citep{kipf2016semi}},
  \mbox{GraphSAGE~\citep{hamilton2017inductive}},
  \mbox{GAT~\citep{velivckovic2017graph}},
  \mbox{APPNP~\citep{gasteiger2018predict}},
  \mbox{SGC~\citep{wu2019simplifying}},
  \mbox{GCNII~\citep{chen2020simple}},
  \mbox{FAGCN~\citep{bo2021beyond}},
  \mbox{GPRGNN~\citep{chien2020adaptive}},
  \mbox{LINKX~\citep{lim2021large}},
  \mbox{GATv2~\citep{brody2021attentive}},
  \mbox{NodeFormer~\citep{wu2022nodeformer}},
  \mbox{SGFormer~\citep{wu2023sgformer}},
  \mbox{NAGphormer~\citep{chen2022nagphormer}},
  \mbox{Polynormer~\citep{deng2024polynormer}} \\
\midrule
Foundation & \mbox{G2T-FM~\citep{eremeev2025turning}},
  \mbox{GraphPFN~\citep{eremeev2025graphpfn}},
  \mbox{NodePFN~\citep{choi2026learning}}
  \mbox{GraphAny~\citep{zhao2025fully}},
  \mbox{GVT~\citep{lee2025view}},
  \mbox{Node4All~\citep{lee2026node4all}}\\
\bottomrule
\end{tabularx}
\vspace{-8pt}
\end{table*}

\subsection{Model Optimization and Data Protocol.}
\label{sec:model-hpo-protocol}

\mypar{Data representation and splits.}
We represent each dataset as a PyG \texttt{Data} object, coalesce repeated edges, make the graph undirected, and add a self-loop per node.
Method-specific preprocessing subsequently modifies this representation (Appendix~\ref{app:checkpoint-audit}).

We use two shared, class-stratified train/validation/test ratios. The
10/10/80 split represents limited label availability, while the 50/25/25
split represents a label-rich setting. 
For each ratio, we use five fixed experimental indices, where each identifies one shared data partition and deterministically sets the random seeds for all stochastic components of the evaluation.
Appendix~\ref{app:splits-randomness} explains how the indices determine the shared partitions and random seeds.

\mypar{Hyperparameter optimization.}
Each supervised method is evaluated in both its \textit{default} and \textit{tuned} forms.
The default evaluation uses the released or author-recommended configuration. The tuned result is selected from a pool containing the default and up to 200 unique configurations sampled from a method-specific search space.
We train a model separately for each configuration and select the final checkpoint and configuration based on validation cross-entropy.
GFMs retain their released pretrained components and optimize only the target-specific preprocessing, label-conditioning, or head-training choices exposed by their interfaces.
Appendix~\ref{app:search-spaces} reports the default settings, search spaces, executed trial counts, and method-level selection outcomes.

\subsection{Evaluation and Reporting}
\label{sec:evaluation-design}

We first average each method's scores over the five experimental indices for dataset-level summaries. The main analysis then weights all 51 datasets equally, while missing results are replaced with the default-GCN result. 
Appendix~\ref{app:aggregation-imputation} provides details on the full aggregation and imputation.

\mypar{Predictive performance.}
We use accuracy as the primary metric for multiclass and AUROC for binary datasets. We also report macro-F1 and cross-entropy for multiclass datasets, as well as accuracy, positive-class precision, and binary cross-entropy for binary datasets.
For overall comparisons, we report Elo ratings derived from pairwise method outcomes on these primary metrics.

\mypar{Computational cost.}
In terms of costs, we record training-and-tuning time, prediction time, their sum (as total workflow time), and peak allocated GPU memory during fitting and prediction.
For the default configuration, training-and-tuning time includes preprocessing and representation construction within the overall training process.
For the tuned method, it includes all executed search trials. 
GFM measurements exclude upstream pretraining and previously cached downloads.
Appendix~\ref{app:compute-environment} specifies the measurement boundaries
and compute environments. 
\begin{figure*}[t]
    \centering

    \begin{subfigure}[t]{0.95\textwidth}
        \centering
        \includegraphics[width=\linewidth]
        {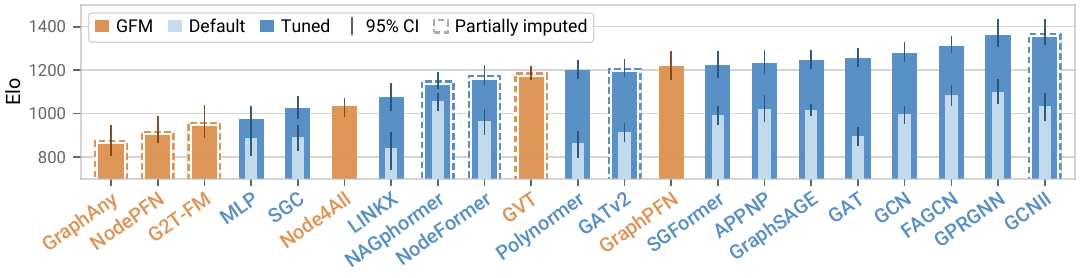}
        \vspace{-16pt}
        \caption{10/10/80 split.}
    \end{subfigure}

    \vspace{4pt}

    \begin{subfigure}[t]{0.95\textwidth}
        \centering
        \includegraphics[width=\linewidth]
        {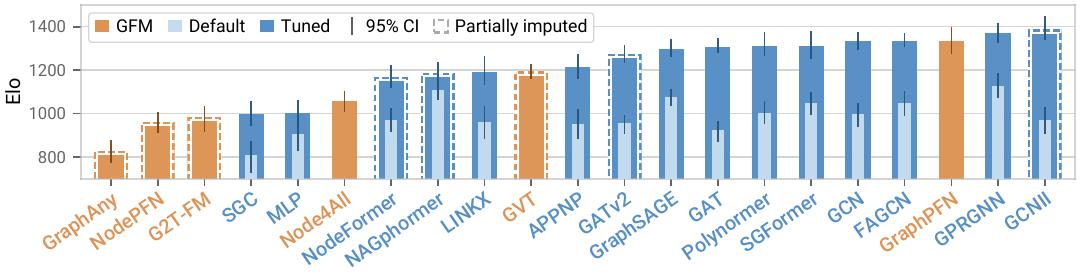}
        \vspace{-16pt}
        \caption{50/25/25 split.}
    \end{subfigure}
    \caption{\textbf{Overall comparison of GFMs and supervised methods.}
    Primary-metric Elo scores (binary AUROC and multiclass accuracy) across all 51 datasets under (a) 10/10/80 and (b) 50/25/25. 
    Orange denotes GFMs, while light and dark blue denote the default and tuned supervised methods.}
    \label{fig:nongfm-elo}
    \vspace{-12pt}
\end{figure*}

\section{Experimental Results}
\label{sec:results}

We present the first systematic comparison of six GFMs and 15 supervised methods across 51 node classification datasets, assessing both predictive generality and computational efficiency. 

\subsection{How Do GFMs Compare with Supervised Methods (RQ1)?}
\label{sec:rq1-overall-comparison}

RQ1 asks whether the cross-dataset reuse offered by GFMs translates into predictive performance competitive with methods trained and tuned for each dataset. 

\mypar{Tuned supervised methods remain strongest overall.}
Figure~\ref{fig:nongfm-elo} reports primary-metric Elo scores across all 51 datasets.
GVT and GraphPFN score above all default supervised configurations under both splits.
After tuning, however, GCNII and GPRGNN occupy the two leading positions in both regimes.
GraphPFN remains the strongest GFM, ranking below several tuned methods under 10/10/80 but rising to third under 50/25/25.
The remaining GFMs fall below the leading tuned methods in both regimes.
Thus, gains over default baselines do not necessarily persist against well-tuned supervised methods, although GraphPFN remains competitive with sufficient labeled nodes.

\begin{figure*}[t]
    \centering

    \includegraphics[
        width=0.249\textwidth,
        trim=67 2 2 2,
        clip
    ]{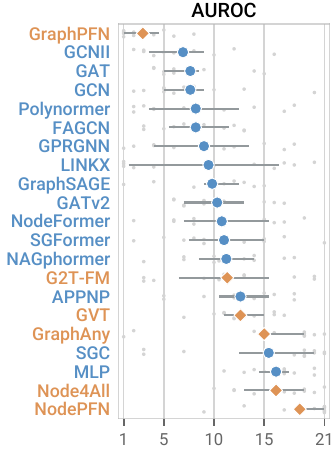}
    \hspace{-2mm}
    \includegraphics[
        width=0.249\textwidth,
        trim=67 2 2 2,
        clip
    ]{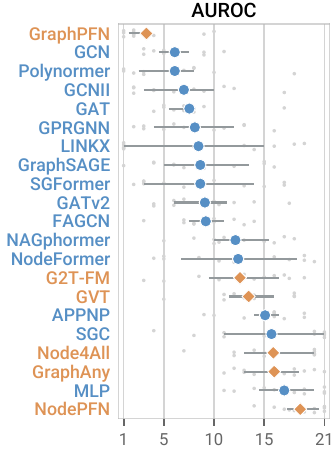}
    \hspace{-2mm}
    \includegraphics[
        width=0.249\textwidth,
        trim=67 2 2 2,
        clip
    ]{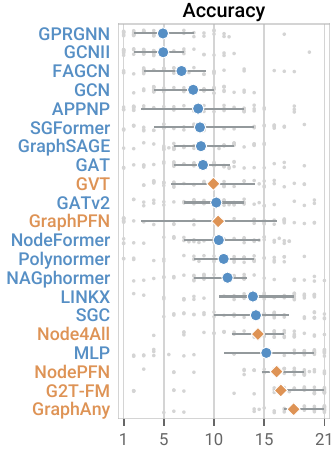}
    \hspace{-2mm}
    \includegraphics[
        width=0.249\textwidth,
        trim=67 2 2 2,
        clip
    ]{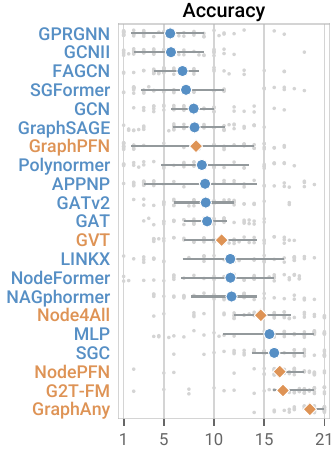}

    \caption{\textbf{Primary-metric ranks of all methods.}
    From left to right: binary, 10/10/80; binary, 50/25/25; multiclass,
    10/10/80; and multiclass, 50/25/25. Gray points show dataset ranks,
    gray horizontal lines show interquartile ranges, and large
    markers show mean ranks (blue circles for supervised methods and
    orange diamonds for GFMs).}
    \label{fig:nongfm-accuracy-ranks}
    \vspace{-12pt}
\end{figure*}

\mypar{GraphPFN leads on binary tasks but trails on multiclass tasks.}
Figure~\ref{fig:nongfm-accuracy-ranks} separates the overall comparison into binary and multiclass tasks.
On binary tasks, GraphPFN achieves the best mean AUROC rank under both splits, ahead of the tuned supervised methods, whereas the other GFMs generally rank lower.
On multiclass tasks, this advantage disappears. 
GPRGNN and GCNII lead under both splits, while the GFMs are generally less competitive.
Thus, GraphPFN's strong overall position under 50/25/25 split Elo leaderboard reflects a pronounced advantage on binary tasks together with improved multiclass performance as more labeled nodes become available.

\subsection{How Broadly Do GFMs Transfer Across Dataset Conditions (RQ2)?}
\label{sec:rq2-gfm-generality}

\begin{figure*}[t]
    \centering
    \begin{subfigure}[t]{0.45\textwidth}
        \centering
        \includegraphics[width=\linewidth]{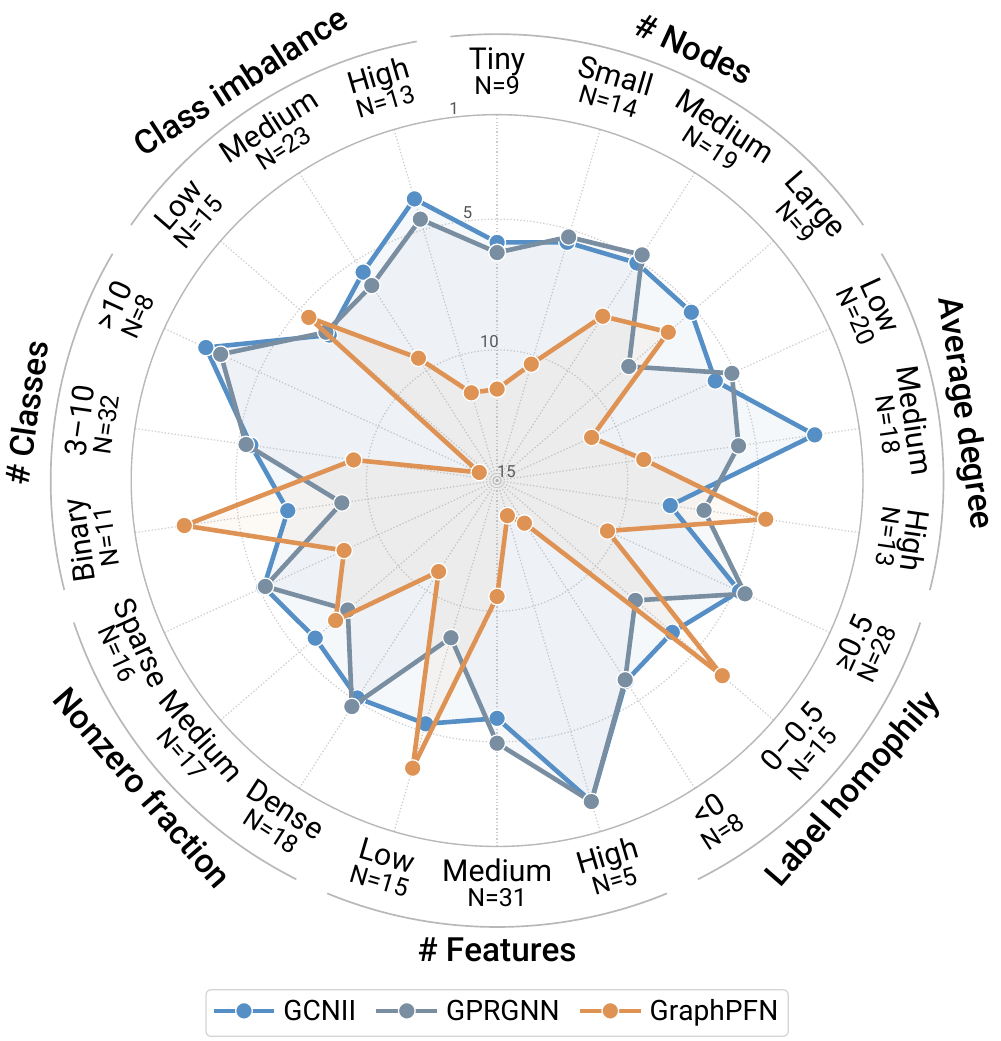}
        \caption{10/10/80 split.}
    \end{subfigure}\hfill
    \begin{subfigure}[t]{0.45\textwidth}
        \centering
        \includegraphics[width=\linewidth]{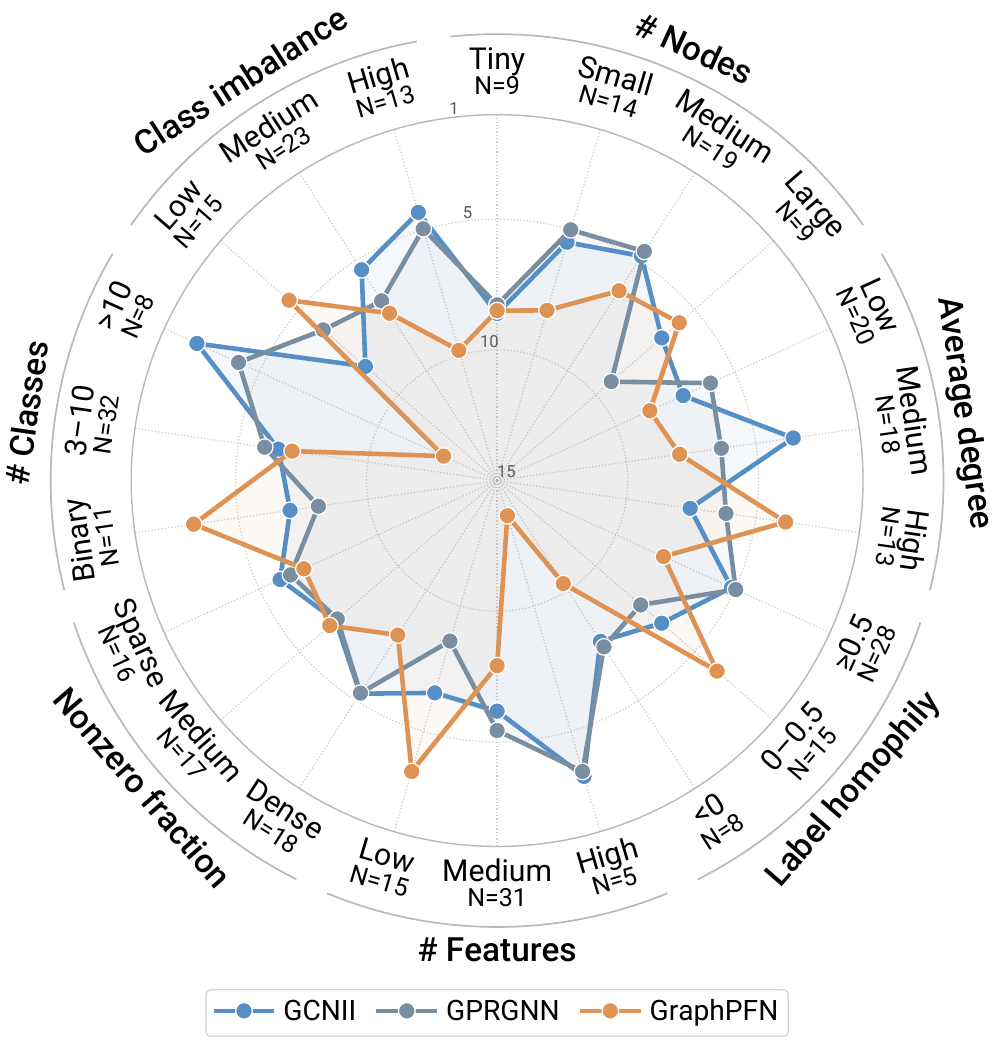}
        \caption{50/25/25 split.}
    \end{subfigure}
    \caption{
    \textbf{Primary-metric ranks across dataset conditions.}
    GCNII, GPRGNN, and GraphPFN are compared across 22 subgroups.
    The radial scale is truncated at rank 15.
    Farther outward indicates a better (lower) rank; $N$ denotes the number of datasets in each subgroup.
    }
    \label{fig:gfm-groupwise-ranks}
\end{figure*}

GFMs aim to reuse pretrained knowledge across graphs with different structures, features, and prediction tasks.
We examine how broadly this promise holds by comparing performance across dataset conditions.
We focus on GraphPFN, the highest-ranked GFM in our benchmark, and compare it with well-tuned GCNII and GPRGNN.
Figure~\ref{fig:gfm-groupwise-ranks} reports mean within-dataset ranks across 22 overlapping subgroups defined by seven properties.
Ranks use each dataset's primary metric and include all 21 methods.
Appendices~\ref{app:dataset-definitions} and~\ref{app:dataset-statistics} define the properties and report their per-dataset values.

\mypar{Transfer remains uneven across graph conditions.}
GraphPFN ranks ahead of both GNNs on binary tasks, high-degree graphs, and graphs with intermediate label homophily under the 10/10/80 splits.
However, it fails under strong homophily, on high-dimensional features, and on highly imbalanced targets.
Tasks with more than ten classes also remain unfavorable.
Thus, even the highest-ranked GFM does not provide consistently competitive results across dataset conditions, indicating that its success in some settings does not translate into broad transfer across diverse graphs.

\mypar{More labeled context does not resolve the gaps.}
Under the 50/25/25 split, GraphPFN ranks ahead of both GNNs on large graphs and datasets with intermediate feature nonzero fractions.
Yet its weaknesses on high-dimensional features, highly imbalanced targets, and tasks with more than ten classes persist.
Additional context therefore does not suffice to make its predictions competitive.
These persistent gaps suggest limitations in adapting pretrained representations to a new graph.
Moreover, tasks with more than ten classes exceed the output space of GraphPFN's tabular foundation model backbone and require mechanisms such as output coding~\citep{eremeev2025graphpfn}.
Feature adaptation and larger label spaces are therefore concrete priorities for extending its applicability.

\subsection{Do GFMs Improve the Performance--Cost Frontier (RQ3)?}
\label{sec:rq3-gfm-efficiency}

\begin{figure*}[t]
\centering

\begin{subfigure}[t]{0.45\textwidth}
    \centering
    \includegraphics[width=\linewidth]{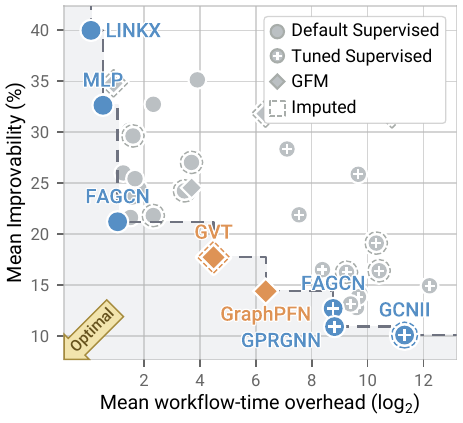}
    \caption{10/10/80 split.}
\end{subfigure}
\hfill
\begin{subfigure}[t]{0.45\textwidth}
    \centering
    \includegraphics[width=\linewidth]{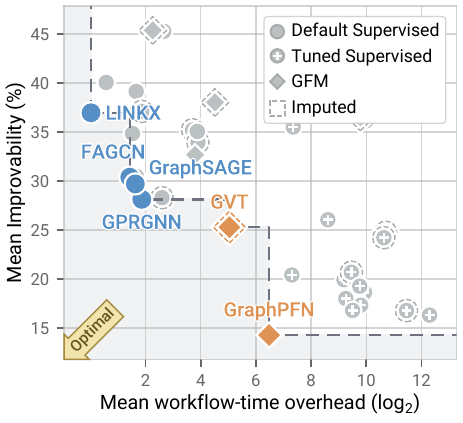}
    \caption{50/25/25 split.}
\end{subfigure}

\caption{\textbf{Performance--cost frontiers under both label splits.}
Mean Improvability is plotted against mean workflow-time overhead for GFMs and the default and tuned supervised methods.
Dashed lines mark the Pareto frontiers; lower is better on both axes.}
\label{fig:rq3-intro-results}
\vspace{-12pt}
\end{figure*}
GFMs can be useful without outperforming supervised methods if they offer competitive predictions at lower cost.
We measure predictive quality using \emph{Mean Improvability}, the average fraction of error removable by matching the best method on each dataset~\citep{erickson2026tabarena}.
We introduce \emph{Mean Overhead} as a general measure of computational cost, defined as the $\log_2$ geometric-mean cost ratio relative to the least costly method on each dataset; a value of $k$ means $2^k$ times the reference cost. We use workflow time as the metric of cost for this RQ.
Appendix~\ref{app:overhead-pareto} provides formal definitions.

\mypar{Two GFMs occupy the endpoint frontiers.}
\Cref{fig:rq3-intro-results} compares the performance--cost frontiers under (a) 10/10/80 and (b) 50/25/25, plotting mean workflow-time overhead against Mean Improvability.
GVT and GraphPFN lie on the Pareto frontier under both splits.
GVT reaches the frontier through lower workflow time but retains a larger predictive gap.
GraphPFN achieves stronger predictions at greater cost and, under 50/25/25, attains the lowest Mean Improvability.
Its mean workflow-time overhead of approximately $6.4$, however, corresponds to roughly $84\times$ the reference cost.
All other GFMs are dominated by methods with both better performance and lower cost.

\begin{figure*}[t]
\centering

\begin{subfigure}[t]{0.45\textwidth}
    \centering
    \includegraphics[width=\linewidth]{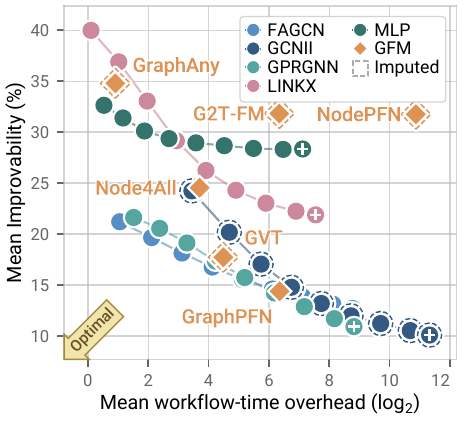}
    \caption{10/10/80 split.}
\end{subfigure}
\hfill
\begin{subfigure}[t]{0.45\textwidth}
    \centering
    \includegraphics[width=\linewidth]{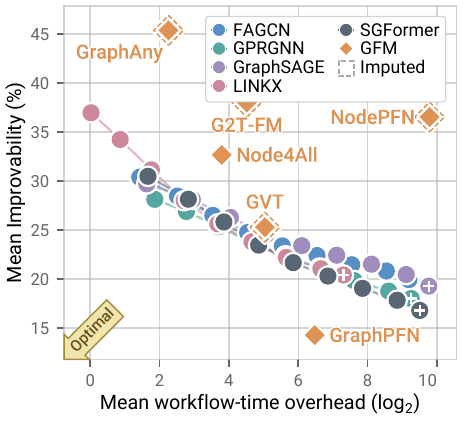}
    \caption{50/25/25 split.}
\end{subfigure}

\caption{\textbf{Performance--cost trajectories across tuning budgets.}
Lines trace supervised methods from 1 to 201 configurations, while diamonds mark GFMs, under (a) 10/10/80 and (b) 50/25/25.}
\label{fig:tuning-trajectories}
\vspace{-8pt}
\end{figure*}

\mypar{Only GraphPFN reaches the frontier against partially tuned methods.}
The above comparisons can overstate the effectiveness of GFMs because they represent each supervised method only before and after a full tuning.
The stricter test is whether a GFM outperforms the best supervised configuration attainable with the same or less workflow time.
Figure~\ref{fig:tuning-trajectories} enables this comparison by tracing supervised methods over budgets of 1, 2, 4, 8, 16, 32, 64, 128, and 201 configurations.
Intermediate workflow costs are interpolated between the default and full-search endpoints (Appendix~\ref{app:overhead-pareto}).

Under this new setting, partially tuned supervised methods overtake both GVT and GraphPFN under 10/10/80, so no GFM expands the aggregate performance--cost frontier.
Under 50/25/25, GraphPFN is the sole exception.
It achieves lower Mean Improvability than every supervised trajectory and extends the estimated aggregate frontier, whereas all remaining GFMs are dominated.

\mypar{Avoiding target-specific training often shifts cost to inference.}
PFN-based GFMs such as G2T-FM, NodePFN, and GraphPFN rely on self-attention with quadratically growing cost to the context length. Their avoidance of target-specific optimization can therefore be offset by expensive adaptation and inference. In contrast, GFMs with lighter mechanisms run faster but generally underperform partially tuned supervised methods at comparable costs. 
Together with RQ1 and RQ2, this result sets a concrete goal for future GFMs: they should outperform the best supervised method attainable within the same workflow-time budget, consistently across label regimes and graph conditions.
\section{Conclusion}

The idea of \acfp{GFM} is that knowledge acquired from pretraining should reduce the work needed to work on a new graph.
\textsc{NodeGround} tests how far current \acp{GFM} fulfill this promise through a unified comparison of 21 methods across 51 node-classification datasets.
The results expose a gap between reusing a pretrained model and obtaining broadly competitive predictions.
Tuned GNNs remain the strongest overall, while GraphPFN's strengths depend on the prediction task, graph conditions, and available labels.
Avoiding model parameter updates also does not guarantee an inexpensive workflow, as adaptation and inference can absorb the computation.

These findings establish a definite target for future GFMs.
Pretrained knowledge should support competitive predictions across varied graphs within the time a supervised method would need to train and tune.
Achieving this requires progress in both transfer across feature and label spaces and the cost of using the pretrained model.
By releasing shared splits, implementations, trial-level results, and a public leaderboard, \textsc{NodeGround} makes these requirements testable.

\mypar{Limitations and future work.}
Our evaluation focuses on node classification under fixed label budgets and shared data splits.
Although the datasets span diverse graph conditions, these protocols do not fully capture non-IID settings, where labeled and query nodes may come from systematically different communities or time periods.
Evaluating such shifts would clarify whether the benefits of pretrained knowledge persist under distribution changes.
The conclusions also do not extend to edge-level or graph-level tasks, such as link prediction and graph classification.
Extending the benchmark to these settings would provide a broader assessment of graph foundation models.

\bibliography{iclr2027_conference}
\bibliographystyle{iclr2027_conference}

\appendix
\clearpage

\vspace{1em}
{\large\bfseries Table of Contents.\par}
\vspace{1.25em}

\begingroup
\setlength{\parindent}{0pt}
\setlength{\parskip}{0pt}

\newcommand{\apptocsection}[3]{%
  \par\addvspace{0.85em}%
  \noindent
  \makebox[2.2em][l]{\textbf{#1}}%
  \textbf{#2}%
  \nobreak\leaders\hbox{\kern0.22em.\kern0.22em}\hfill
  \nobreak\makebox[2.3em][r]{\textbf{\pageref{#3}}}%
  \par\addvspace{0.25em}%
}

\newcommand{\apptocsubsection}[3]{%
  \noindent\hspace*{2.2em}%
  \makebox[3.2em][l]{#1}%
  #2%
  \nobreak\leaders\hbox{\kern0.22em.\kern0.22em}\hfill
  \nobreak\makebox[2.3em][r]{\pageref{#3}}%
  \par\addvspace{0.18em}%
}

\apptocsection{A}{Additional Analyses and Robustness}{app:additional-results}
\apptocsubsection{A.1}{Analysis Cohorts and Sensitivity to Evaluation Choices}{app:complete-case-results}
\apptocsubsection{A.2}{Performance--Memory Tradeoffs}{app:gfm-cost-analyses}
\apptocsubsection{A.3}{Pairwise Win Rates and Statistical Significance}{app:pairwise-significance}
\apptocsubsection{A.4}{Tuning Requirements and Model-Selection Robustness}{app:tuning-robustness}
\apptocsection{B}{Dataset Curation and Catalog}{app:datasets}
\apptocsubsection{B.1}{Candidate Sources}{app:dataset-selection}
\apptocsubsection{B.2}{Ordered Screening, Canonicalization, and Task Identity}{app:candidate-ledger}
\apptocsubsection{B.3}{Retained Dataset Catalog}{app:dataset-catalog}
\apptocsubsection{B.4}{Dataset Properties}{app:dataset-definitions}
\apptocsubsection{B.5}{Per-Dataset Statistics}{app:dataset-statistics}
\apptocsubsection{B.6}{Benchmark Comparison Criteria and Evidence}{app:benchmark-comparison-evidence}

\apptocsection{C}{Methods, Artifacts, and Optimization}{app:methods}
\apptocsubsection{C.1}{Method Scope and GFM Definition}{app:method-scope}
\apptocsubsection{C.2}{Implementations and Target-Dataset Adaptation}{app:method-implementations}
\apptocsubsection{C.3}{Pretrained Artifacts, Preprocessing, and Overlap}{app:checkpoint-audit}
\apptocsubsection{C.4}{Optimization Protocol, Search Spaces, and Outcomes}{app:search-spaces}

\apptocsection{D}{Evaluation and Computational Accounting}{app:evaluation-accounting}
\apptocsubsection{D.1}{Splits and Randomness}{app:splits-randomness}
\apptocsubsection{D.2}{Predictive Metrics and Model Selection}{app:metric-selection}
\apptocsubsection{D.3}{Aggregation, Missing Results, and Uncertainty}{app:aggregation-imputation}
\apptocsubsection{D.4}{Computational Measurement}{app:compute-environment}
\apptocsubsection{D.5}{Mean Improvability, Mean Overhead, and Pareto Frontiers}{app:overhead-pareto}

\apptocsection{E}{Additional Dataset-Level and Configuration Results}{app:complete-results}
\apptocsubsection{E.1}{Dataset-Level Predictive Outcomes}{app:complete-predictive-results}
\apptocsubsection{E.2}{Configuration-Level Rankings}{app:configuration-results}
\apptocsubsection{E.3}{Predictive Coverage}{app:run-status}

\endgroup
\clearpage

\crefalias{section}{appendix}

\section{Additional Analyses and Robustness}
\label{app:additional-results}
\raggedbottom

The main analysis reveals a clear trade-off between predictive performance and computational cost among the 21 evaluated methods.
Tuned GNNs are strongest overall, while GraphPFN is the only GFM to join the leading group; its strongest results occur in the label-rich regime and require substantial computation.
This appendix examines the robustness and practical implications of these findings.
We first check whether the conclusions change when we use different subsets of datasets and methods, handle missing results differently, or change the predictive metrics and aggregation rules.
We then compare alternative measures of computational cost, examine how often each method outperforms another across datasets, and quantify how tuning budgets affect performance and cost.

\subsection{Analysis Cohorts and Sensitivity to Evaluation Choices}
\label{app:complete-case-results}

Section~\ref{sec:results} compares all 21 methods across the full set of 51 datasets. Because some method--dataset results are unavailable, the primary rankings replace them with the corresponding GCN-Default results. We assess the sensitivity of the conclusions to this choice by recomputing the rankings under three fixed cohorts that vary dataset coverage, method coverage, and imputation.

\textbf{Cohort V1 (full benchmark)} follows the main-text policy of replacing each unavailable result with the corresponding GCN-Default result. \textbf{Cohort V2 (complete case)} retains the 28 datasets with observed results for all 21 methods and uses no predictive imputation. \textbf{Cohort V3 (restricted method)} retains all 51 datasets but
limits the comparison to 15 near-complete methods. Its only imputed entries are GraphAny and GVT on \texttt{ogbn\_arxiv}, whose observed results are excluded because of pretraining overlap. Table~\ref{tab:result-cohorts} reports the details for each cohort.

\begin{table*}[h]
\centering
\caption{\textbf{Predictive-result cohorts used throughout the paper and release.} A run cell is one dataset--method--split--index record. Cohort V3 is near-complete rather than a strict complete-case cohort. Its 20 imputed run cells are GraphAny and GVT on \texttt{ogbn\_arxiv}.}
\label{tab:result-cohorts}
\scriptsize
\setlength{\tabcolsep}{2pt}
\renewcommand{\arraystretch}{1.08}
\begin{tabular*}{\textwidth}{@{\extracolsep{\fill}}lp{0.31\textwidth}lrrrrr@{}}
\toprule
Cohort & Definition & Split & Datasets & Methods & Idx. & Cells & Imputed \\
\midrule
V1 & Full benchmark & 10/10/80 & 51 & 21 & 5 & 5355 & 144 \\
V1 & Full benchmark & 50/25/25 & 51 & 21 & 5 & 5355 & 244 \\
V2 & Complete case & 10/10/80 & 28 & 21 & 5 & 2940 & 0 \\
V2 & Complete case & 50/25/25 & 28 & 21 & 5 & 2940 & 0 \\
V3 & Restricted method & 10/10/80 & 51 & 15 & 5 & 3825 & 10 \\
V3 & Restricted method & 50/25/25 & 51 & 15 & 5 & 3825 & 10 \\
\bottomrule
\end{tabular*}
\end{table*}

\begin{figure*}[!p]
    \centering
    \begin{subfigure}[t]{0.98\textwidth}
        \centering
        \includegraphics[width=\linewidth,height=0.125\textheight,keepaspectratio]
        {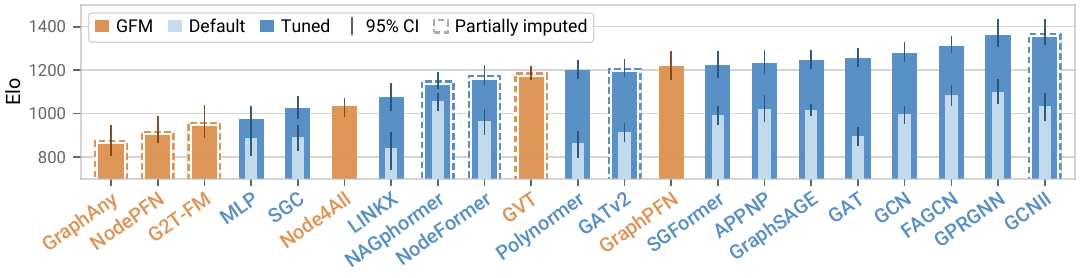}
        \caption{Cohort V1, 10/10/80.}
    \end{subfigure}
    \vspace{0.1em}

    \begin{subfigure}[t]{0.98\textwidth}
        \centering
        \includegraphics[width=\linewidth,height=0.125\textheight,keepaspectratio]
        {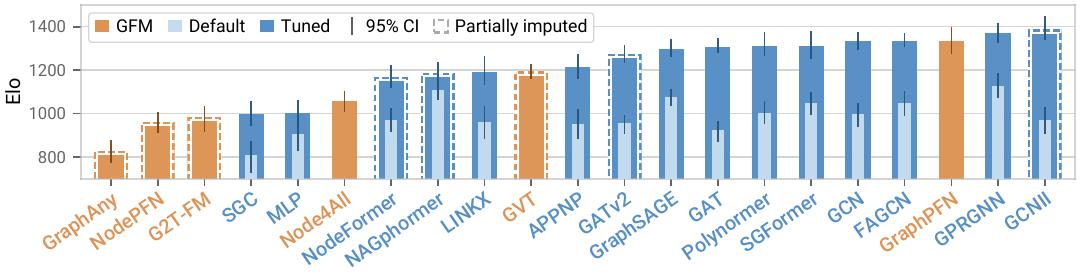}
        \caption{Cohort V1, 50/25/25.}
    \end{subfigure}
    \vspace{0.1em}

    \begin{subfigure}[t]{0.98\textwidth}
        \centering
        \includegraphics[width=\linewidth,height=0.125\textheight,keepaspectratio]
        {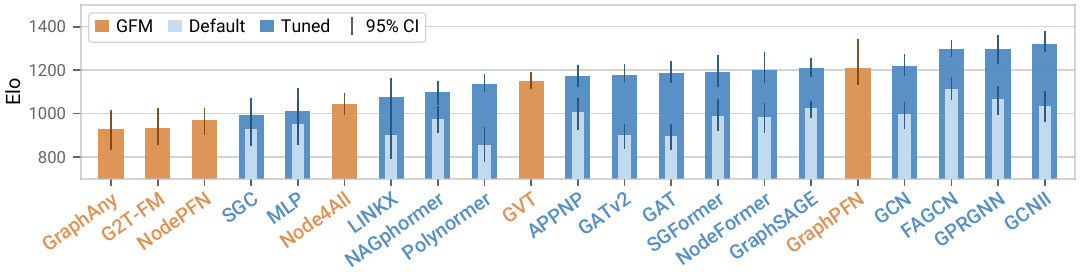}
        \caption{Cohort V2, 10/10/80.}
    \end{subfigure}
    \vspace{0.1em}

    \begin{subfigure}[t]{0.98\textwidth}
        \centering
        \includegraphics[width=\linewidth,height=0.125\textheight,keepaspectratio]
        {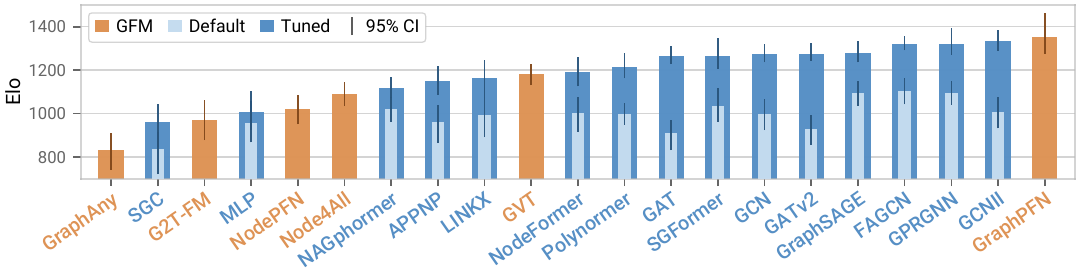}
        \caption{Cohort V2, 50/25/25.}
    \end{subfigure}
    \vspace{0.1em}

    \begin{subfigure}[t]{0.98\textwidth}
        \centering
        \includegraphics[width=\linewidth,height=0.125\textheight,keepaspectratio]
        {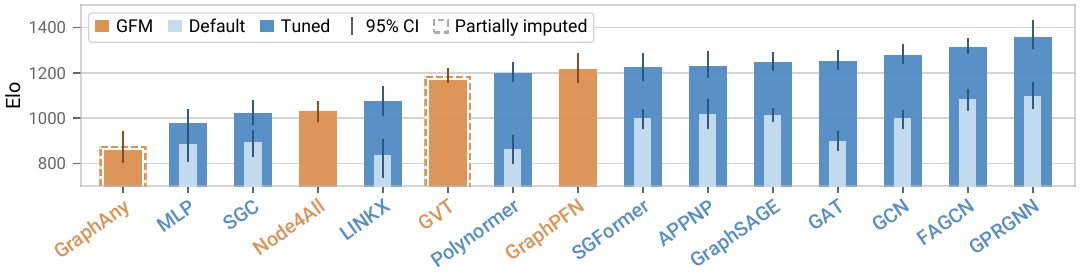}
        \caption{Cohort V3, 10/10/80.}
    \end{subfigure}
    \vspace{0.1em}

    \begin{subfigure}[t]{0.98\textwidth}
        \centering
        \includegraphics[width=\linewidth,height=0.125\textheight,keepaspectratio]
        {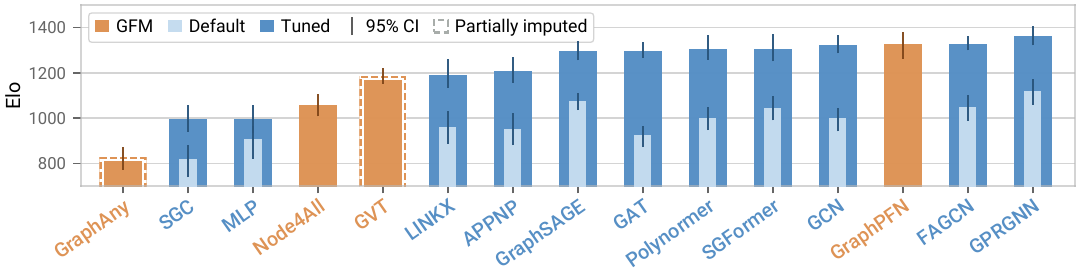}
        \caption{Cohort V3, 50/25/25.}
    \end{subfigure}

    \caption{\textbf{Elo leaderboards across analysis cohorts and label splits.}
    Cohort V1 evaluates all 21 methods on all 51 datasets using the benchmark
    replacement policy; Cohort V2 uses the 28 complete-case datasets; and
    Cohort V3 evaluates 15 near-complete methods on all 51 datasets. Dashed outlines identify partially imputed results, and error bars show 95\% confidence intervals.}
    \label{fig:app-cohort-elo}
\end{figure*}

\mypar{The leading methods remain similar across evaluation subsets.}
\Cref{fig:app-cohort-elo} compares Elo rankings under both label splits using all 51 datasets (V1), the 28 datasets with complete results (V2), and all 51 datasets with a reduced set of methods (V3).
Under 10/10/80, GCNII leads V1 and V2, while GPRGNN leads V3, which excludes GCNII.
Under 50/25/25, GraphPFN rises from third in V1 to first in V2, whereas GPRGNN leads V3.
GraphPFN's exact position therefore depends on which datasets and competitors are included, but it remains the sole GFM among the leading methods.
Across all three comparisons, the strongest competition comes from tuned GNNs.

\begin{figure*}[!htb]
    \centering
    \begin{subfigure}[t]{0.98\textwidth}
        \centering
        \includegraphics[width=\linewidth]
        {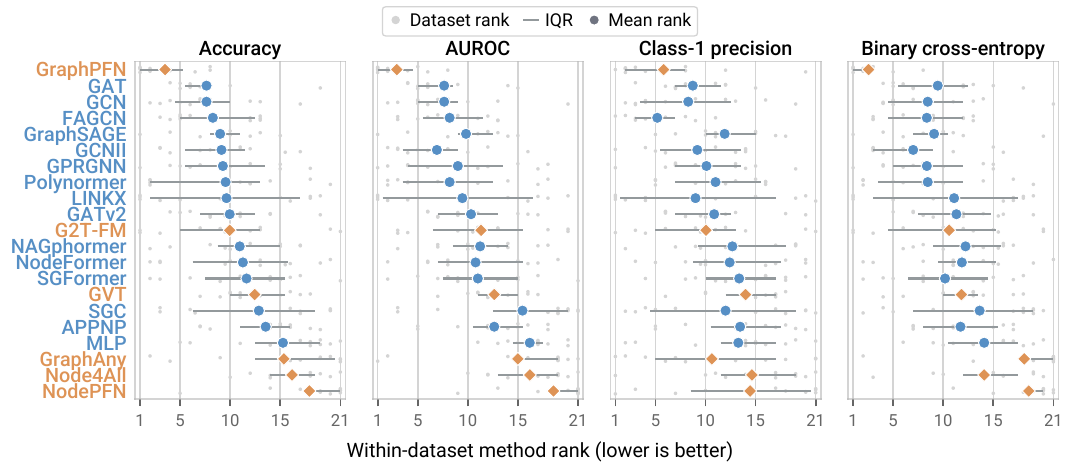}
        \caption{Cohort V1.}
    \end{subfigure}
    \vspace{-0.25em}

    \begin{subfigure}[t]{0.98\textwidth}
        \centering
        \includegraphics[width=\linewidth]
        {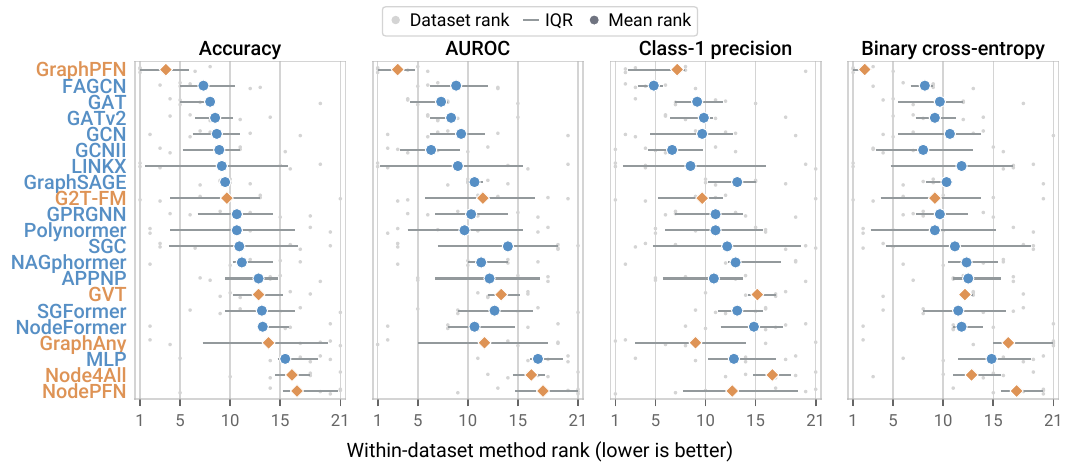}
        \caption{Cohort V2.}
    \end{subfigure}
    \vspace{-0.25em}

    \begin{subfigure}[t]{0.98\textwidth}
        \centering
        \includegraphics[width=\linewidth]
        {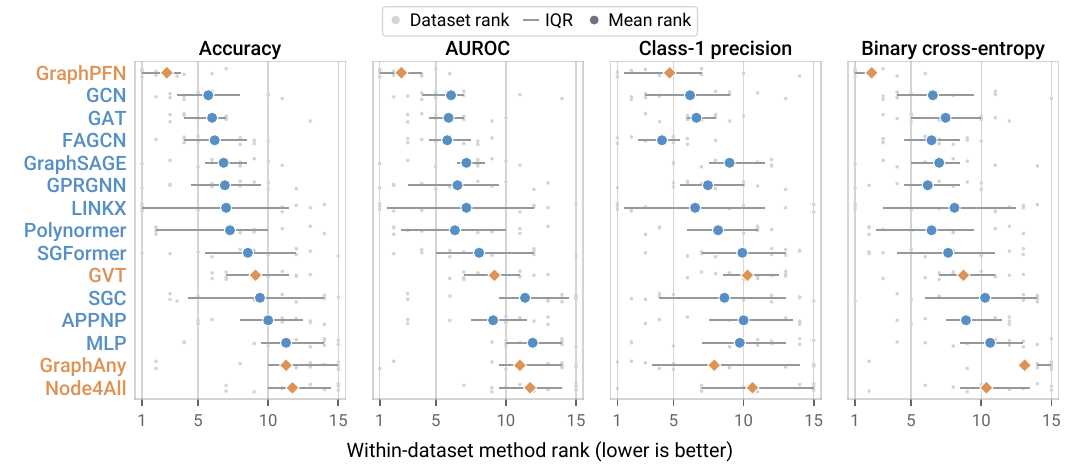}
        \caption{Cohort V3.}
    \end{subfigure}

    \caption{\textbf{Binary-metric ranks across analysis cohorts under the
    10/10/80 split.} Gray points show dataset-level ranks, and gray horizontal lines show interquartile ranges. Large blue circles and orange diamonds
    show mean ranks for supervised methods and GFMs, respectively.
    Lower ranks are better.}
    \label{fig:app-cohort-binary-metrics-low-label}
\end{figure*}

\begin{figure*}[!htb]
    \centering
    \begin{subfigure}[t]{0.98\textwidth}
        \centering
        \includegraphics[width=\linewidth]
        {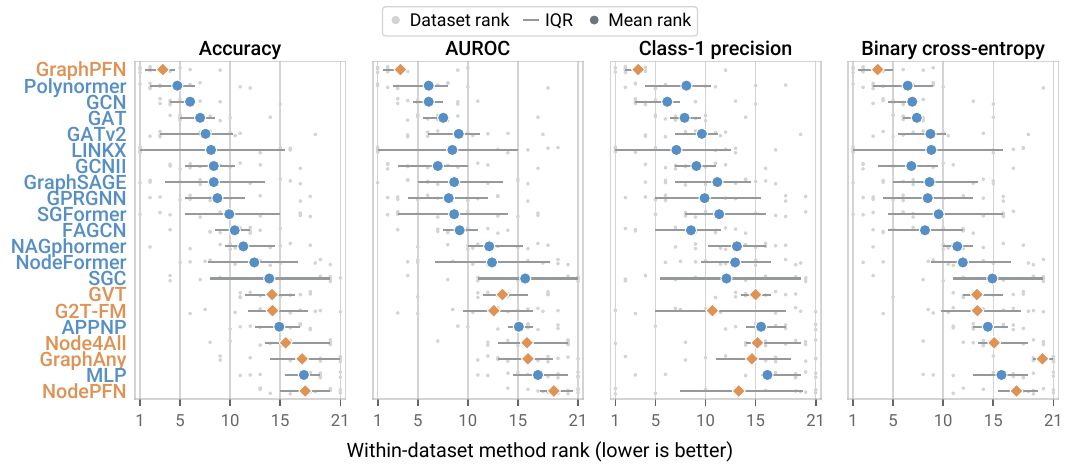}
        \caption{Cohort V1.}
    \end{subfigure}
    \vspace{-0.25em}

    \begin{subfigure}[t]{0.98\textwidth}
        \centering
        \includegraphics[width=\linewidth]
        {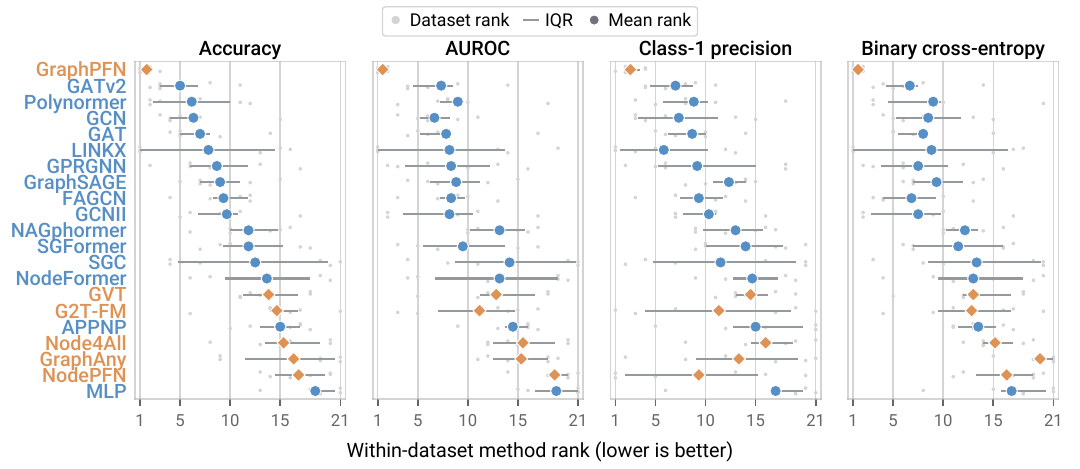}
        \caption{Cohort V2.}
    \end{subfigure}
    \vspace{-0.25em}

    \begin{subfigure}[t]{0.98\textwidth}
        \centering
        \includegraphics[width=\linewidth]
        {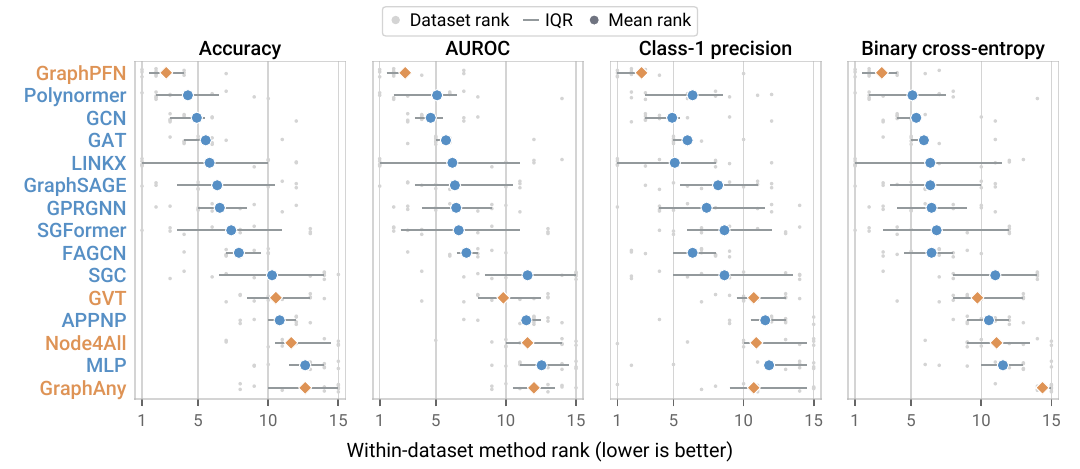}
        \caption{Cohort V3.}
    \end{subfigure}

    \caption{\textbf{Binary-metric ranks across analysis cohorts under the
    50/25/25 split.} Panels and visual encodings follow
    Figure~\ref{fig:app-cohort-binary-metrics-low-label}.}
    \label{fig:app-cohort-binary-metrics-label-rich}
\end{figure*}

\begin{figure*}[!htb]
    \centering
    \begin{subfigure}[t]{0.98\textwidth}
        \centering
        \includegraphics[width=\linewidth]
        {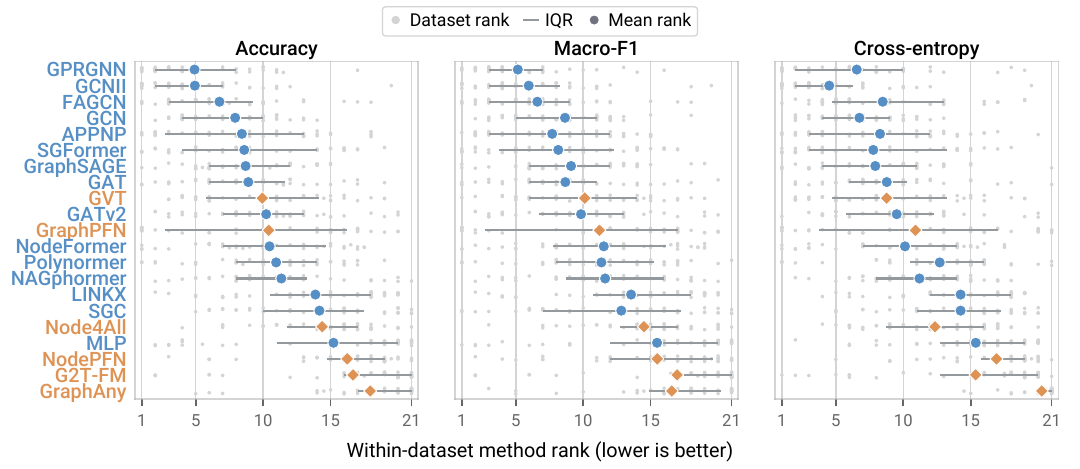}
        \caption{Cohort V1.}
    \end{subfigure}
    \vspace{-0.25em}

    \begin{subfigure}[t]{0.98\textwidth}
        \centering
        \includegraphics[width=\linewidth]
        {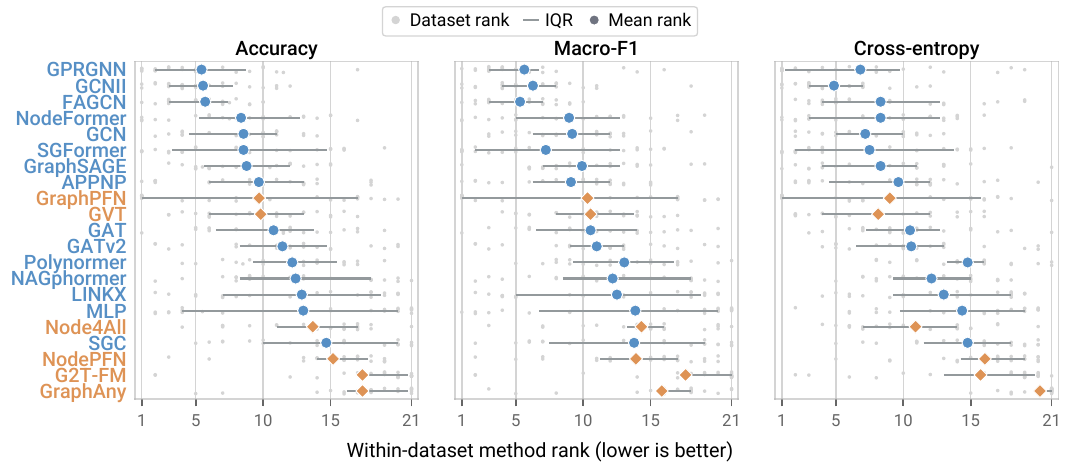}
        \caption{Cohort V2.}
    \end{subfigure}
    \vspace{-0.25em}

    \begin{subfigure}[t]{0.98\textwidth}
        \centering
        \includegraphics[width=\linewidth]
        {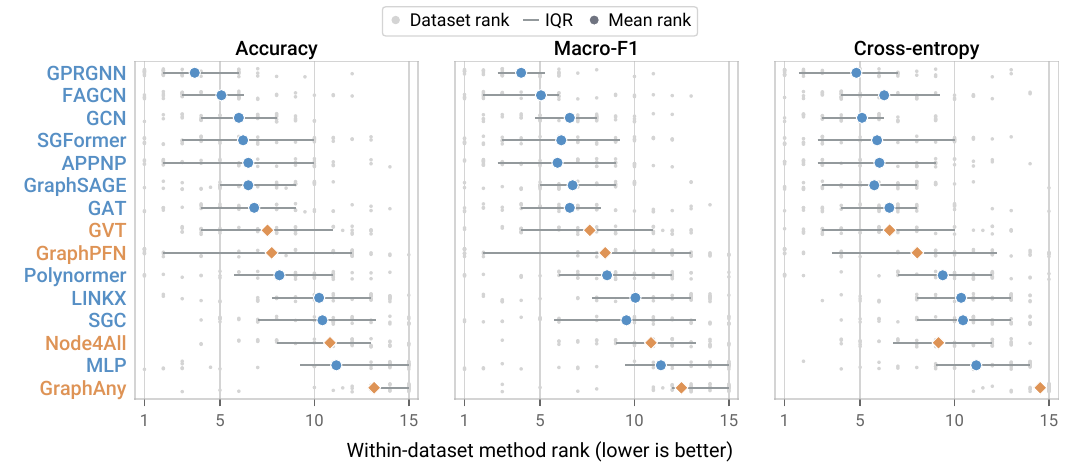}
        \caption{Cohort V3.}
    \end{subfigure}

    \caption{\textbf{Multiclass-metric ranks across analysis cohorts under the
    10/10/80 split.} Gray points show dataset-level ranks, and gray horizontal lines show interquartile ranges. Large blue circles and orange diamonds
    show mean ranks for supervised methods and GFMs, respectively.
    Lower ranks are better.}
    \label{fig:app-cohort-multiclass-metrics-low-label}
\end{figure*}

\begin{figure*}[!htb]
    \centering
    \begin{subfigure}[t]{0.98\textwidth}
        \centering
        \includegraphics[width=\linewidth]
        {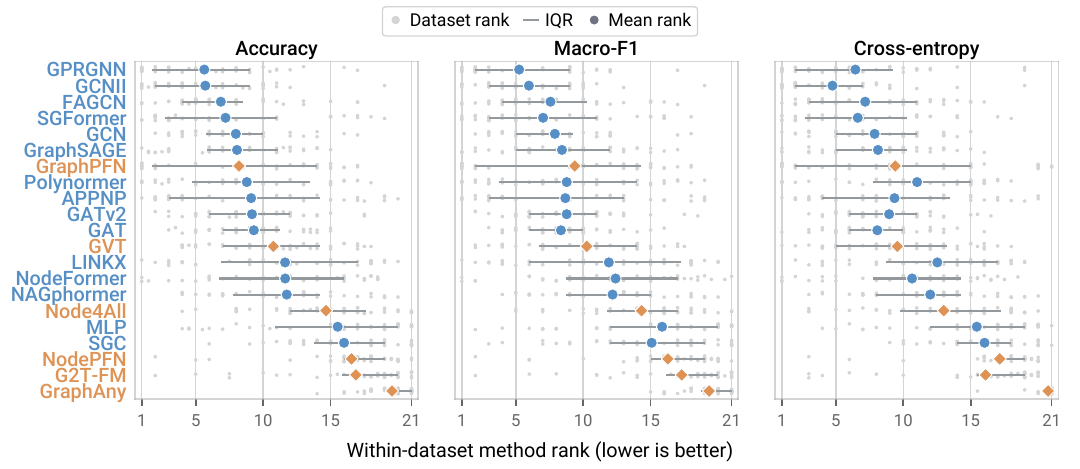}
        \caption{Cohort V1.}
    \end{subfigure}
    \vspace{-0.25em}

    \begin{subfigure}[t]{0.98\textwidth}
        \centering
        \includegraphics[width=\linewidth]
        {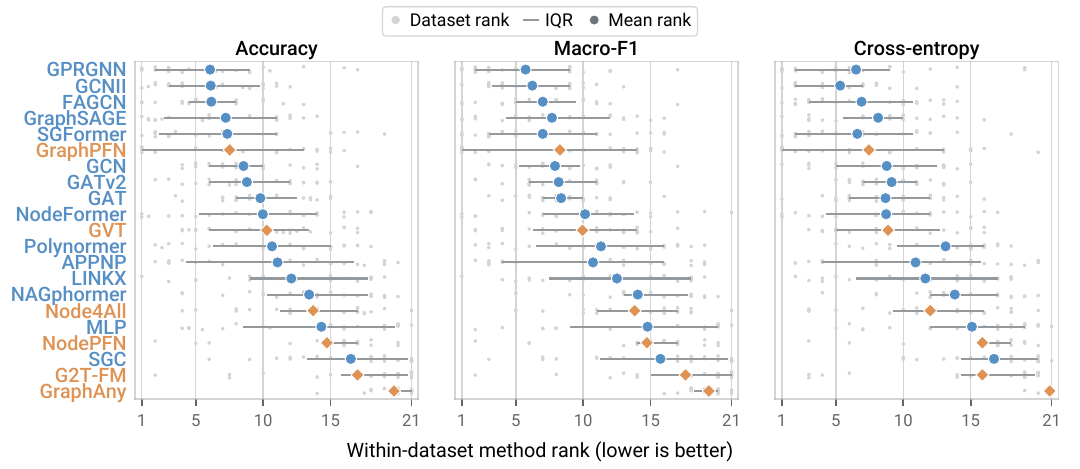}
        \caption{Cohort V2.}
    \end{subfigure}
    \vspace{-0.25em}

    \begin{subfigure}[t]{0.98\textwidth}
        \centering
        \includegraphics[width=\linewidth]
        {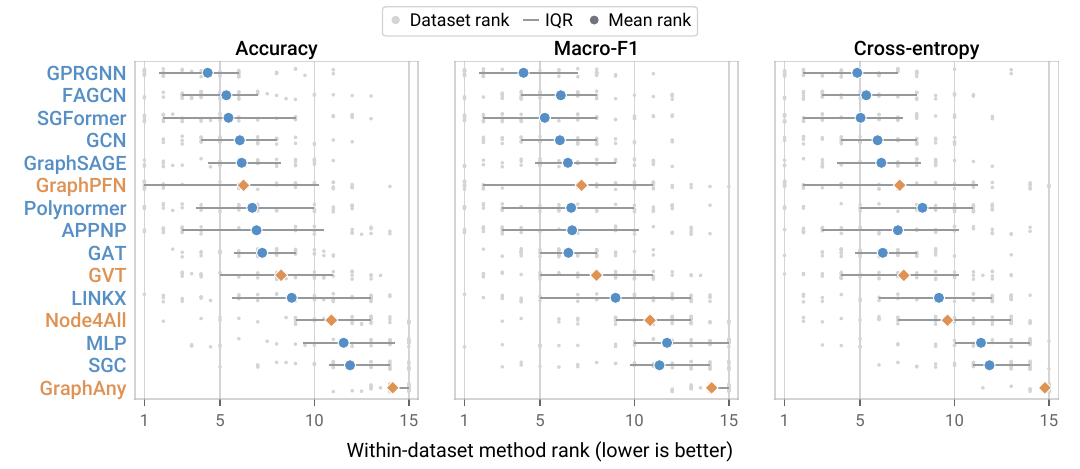}
        \caption{Cohort V3.}
    \end{subfigure}

    \caption{\textbf{Multiclass-metric ranks across analysis cohorts under the
    50/25/25 split.} Panels and visual encodings follow
    Figure~\ref{fig:app-cohort-multiclass-metrics-low-label}.}
    \label{fig:app-cohort-multiclass-metrics-label-rich}
\end{figure*}

\mypar{Metric-specific strengths persist across cohorts.}
Figures~\ref{fig:app-cohort-binary-metrics-low-label}--\ref{fig:app-cohort-multiclass-metrics-label-rich} compare within-dataset ranks for every reported predictive metric. Across all three cohorts, GraphPFN leads binary accuracy, AUROC, and binary cross-entropy under both splits, whereas positive-class precision produces a different ordering. The multiclass results consistently favor GPRGNN and FAGCN, together with GCNII in the
cohorts that include it. Accuracy and Macro-F1 yield similar leading groups; cross-entropy moves several methods but does not make any GFM a consistent multiclass leader. The main contrast between GraphPFN's binary strength and the GNN advantage on multiclass datasets is therefore not created by imputation or by a single predictive metric.

\FloatBarrier

\begin{figure}[H]
    \centering
    \begin{subfigure}[t]{0.49\textwidth}
        \centering
        \includegraphics[width=\linewidth]
        {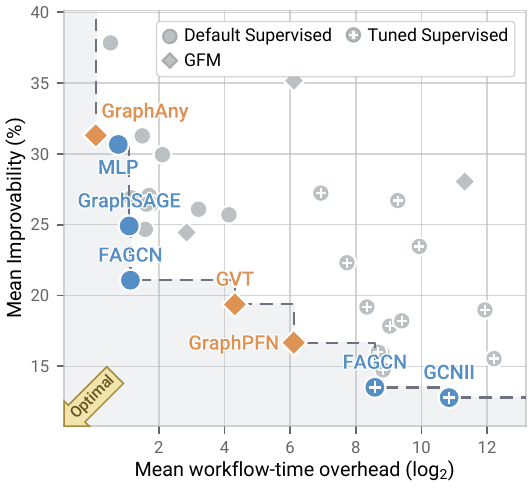}
        \caption{Cohort V2, 10/10/80 split.}
    \end{subfigure}
    \hfill
    \begin{subfigure}[t]{0.49\textwidth}
        \centering
        \includegraphics[width=\linewidth]
        {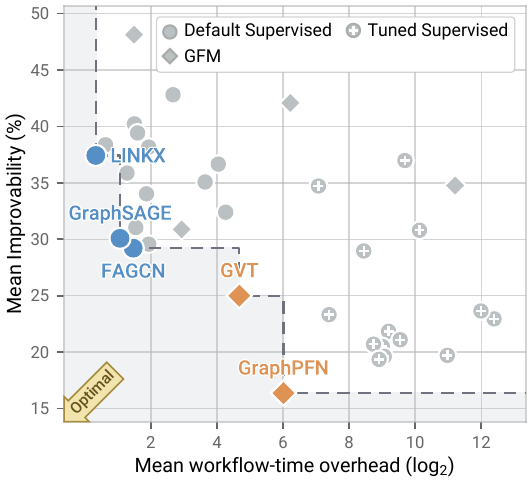}
        \caption{Cohort V2, 50/25/25 split.}
    \end{subfigure}

    \vspace{0.3em}

    \begin{subfigure}[t]{0.49\textwidth}
        \centering
        \includegraphics[width=\linewidth]
        {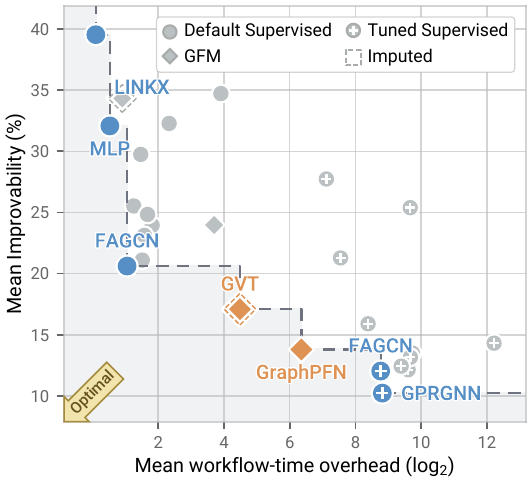}
        \caption{Cohort V3, 10/10/80 split.}
    \end{subfigure}
    \hfill
    \begin{subfigure}[t]{0.49\textwidth}
        \centering
        \includegraphics[width=\linewidth]
        {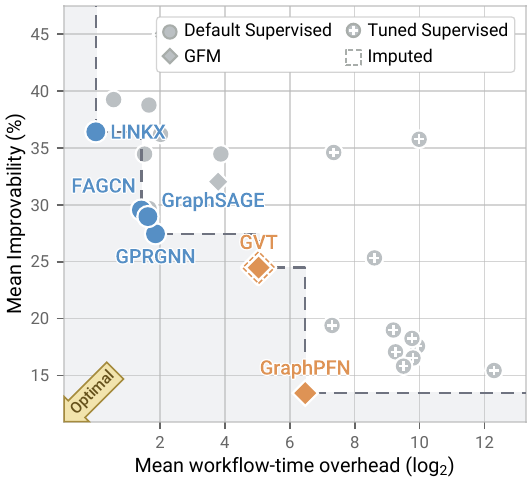}
        \caption{Cohort V3, 50/25/25 split.}
    \end{subfigure}

    \caption{\textbf{Workflow-time performance--cost frontiers under
    alternative analysis cohorts.}
    Mean Improvability is plotted against Mean Overhead from total
    target-dataset time; lower is better on both axes. Dashed lines mark
    Pareto frontiers. The Cohort V1 analyses appear in Figure~\ref{fig:rq3-intro-results}.}
    \label{fig:app-cohort-workflow-pareto}
\end{figure}

\mypar{The performance--cost conclusion also survives cohort changes.}
Figure~\ref{fig:app-cohort-workflow-pareto} repeats the endpoint performance--cost analysis for V2 and V3 and complements the V1 results in Figure~\ref{fig:rq3-intro-results}. Cohort choice changes some efficient low-cost points: GraphAny, for example, joins the low-overhead end of the V2 frontier under 10/10/80. It does not change the central tradeoff. In every cohort, GraphPFN trails the strongest tuned GNNs under 10/10/80 but attains the lowest Mean Improvability under 50/25/25, always at substantial workflow
overhead. GVT remains cheaper but leaves a larger predictive gap. Apart from the low-cost GraphAny exception noted above, the remaining GFMs are dominated. Thus neither complete-case filtering nor the restricted method universe produces a GFM that is both broadly strong and computationally inexpensive.

\mypar{Changing the metric can change which methods rank highest.}
Using the 28 datasets with complete results (V2), \Cref{tab:app-ranking-agreement} checks whether different predictive metrics give similar method rankings.
For multiclass tasks, switching from accuracy to Macro-F1 barely changes the ordering: the Spearman correlations are 0.978 and 0.989 under 10/10/80 and 50/25/25, respectively.
For binary tasks, AUROC and positive-class precision give less consistent rankings, with Kendall correlations of 0.519 and 0.664.
A method that ranks positive examples above negative ones well may therefore not be the best at avoiding false positives among its positive predictions.
The preferred method can depend on which of these objectives matters for the task.

\mypar{Changing how results are combined preserves the winner, but not every position.}
\Cref{tab:app-design-sensitivity} combines the same V2 results in three ways.
Mean rank summarizes where each method places on each dataset, Elo summarizes how often it beats other methods, and Mean Improvability measures how much of its prediction error could be removed by using the best configuration.
All three place GCNII first under 10/10/80 and GraphPFN first under 50/25/25.
The overall orderings are also similar, with pairwise Spearman correlations ranging from 0.955 to 0.993.
However, individual methods move by up to four places under 10/10/80 and six under 50/25/25, and the set of top-four methods changes.
Thus, the choice of summary measure does not change the winner in either split, but it does affect comparisons among the remaining methods.

\begin{table*}[!tbp]
\centering
\caption{\textbf{Selected ranking correlations on Cohort V2.} The released correlation table contains every metric, split, and aggregation comparison.}
\small
\setlength{\tabcolsep}{7pt}
\begin{tabular}{@{}lcccl@{}}
\toprule
Comparison & Coefficient & 10/10/80 & 50/25/25 & Interpretation \\
\midrule
Multiclass accuracy vs. Macro-F1 & Spearman & 0.978 & 0.989 & Strong agreement \\
Binary AUROC vs. positive precision & Kendall & 0.519 & 0.664 & Metric-sensitive ordering \\
\bottomrule
\end{tabular}
\label{tab:app-ranking-agreement}
\end{table*}

\begin{table*}[!tbp]
    \centering
    \caption{\textbf{Agreement among cross-dataset aggregation rules on Cohort V2.}
    Rank denotes mean within-dataset rank and MI denotes Mean Improvability. Each coefficient is the Spearman correlation between the indicated pair of complete method rankings. Leader is the method ranked first by all three rules. Max. shift is the largest change in any method's position across the three rankings, and Same top four indicates whether they retain the same set of four leading methods.}
    \small
    \setlength{\tabcolsep}{8pt}
    \begin{tabular}{@{}lcccccc@{}}
        \toprule
        Split
        & Rank--Elo
        & Rank--MI
        & Elo--MI
        & Leader
        & Max. shift
        & \shortstack{Same\\top four} \\
        \midrule
        10/10/80
        & 0.990
        & 0.962
        & 0.969
        & GCNII
        & 4
        & No \\
        50/25/25
        & 0.993
        & 0.972
        & 0.955
        & GraphPFN
        & 6
        & No \\
        \bottomrule
          \end{tabular}
    \label{tab:app-design-sensitivity}
\end{table*}

\subsection{Performance--Memory Tradeoffs}
\label{app:gfm-cost-analyses}

We examine whether the performance--cost tradeoff in RQ3 also holds
when computational cost is measured by peak GPU memory rather than
workflow time. Using Cohort V1, we compare 15 supervised-method defaults,
15 tuned endpoints, and six GFMs across all 51 datasets.
Missing performance and memory values are replaced with the corresponding
GCN-Default results, following the main-text policy.

\begin{figure}[t]
    \centering
    \begin{subfigure}[t]{0.49\textwidth}
        \centering
        \includegraphics[width=\linewidth]
        {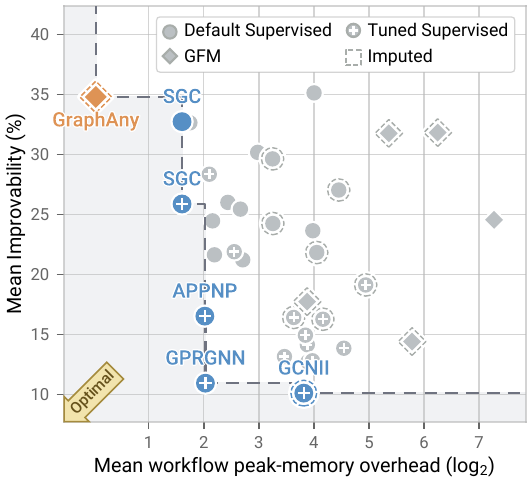}
        \caption{10/10/80 split.}
    \end{subfigure}
    \hfill
    \begin{subfigure}[t]{0.49\textwidth}
        \centering
        \includegraphics[width=\linewidth]
        {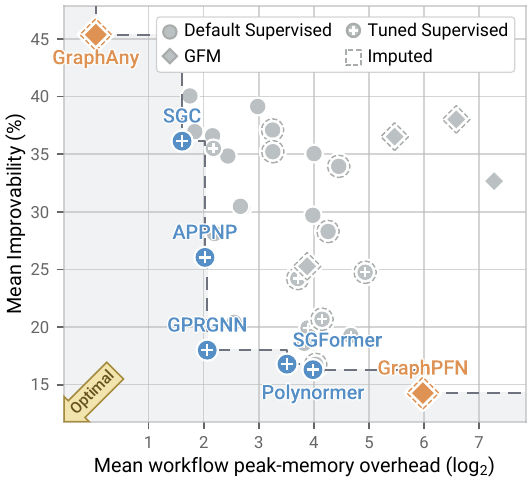}
        \caption{50/25/25 split.}
    \end{subfigure}
    \caption{\textbf{Workflow-peak-memory performance--cost frontiers on Cohort V1.}
    Mean Improvability is plotted against Mean Overhead from the larger of
    fit/search and prediction peak GPU memory. Dashed lines mark Pareto
    frontiers; lower is better on both axes.}
    \label{fig:app-workflow-memory-pareto}
\end{figure}

\mypar{Low memory use and strong predictions favor different GFMs.}
Figure~\ref{fig:app-workflow-memory-pareto} plots Mean Improvability
against memory overhead.
Workflow peak memory is the larger of the fit/search and prediction
peaks, not their sum, because these stages run sequentially.
For GraphAny, which fits on CPU, we use its prediction-stage GPU peak;
this measure does not include CPU memory.

GraphAny anchors the low-memory end of the frontier under both splits,
but leaves a large predictive gap.
GraphPFN is dominated under 10/10/80 and reaches the 50/25/25 frontier
through stronger predictions, at roughly $2^6$ times the reference
memory on a geometric-mean basis.
Thus, the two GFMs reach the frontier for different reasons:
GraphAny minimizes GPU memory use, whereas GraphPFN trades substantially
more memory for predictive quality.
The memory analysis therefore reinforces the workflow-time finding:
current GFMs do not consistently combine strong predictions with
low resource requirements across both label regimes.

\subsection{Pairwise Win Rates and Statistical Significance}
\label{app:pairwise-significance}

Overall rankings do not show how often one method outperforms a particular competitor.
We first compare every pair of methods across all 51 datasets (V1), using the same replacement policy for missing results as in the main analysis.
We then test whether the observed performance differences are statistically supported using the 28 datasets with complete results.

\begin{figure*}[!p]
    \centering
    \begin{subfigure}[t]{0.85\textwidth}
        \centering
        \includegraphics[width=\linewidth]
        {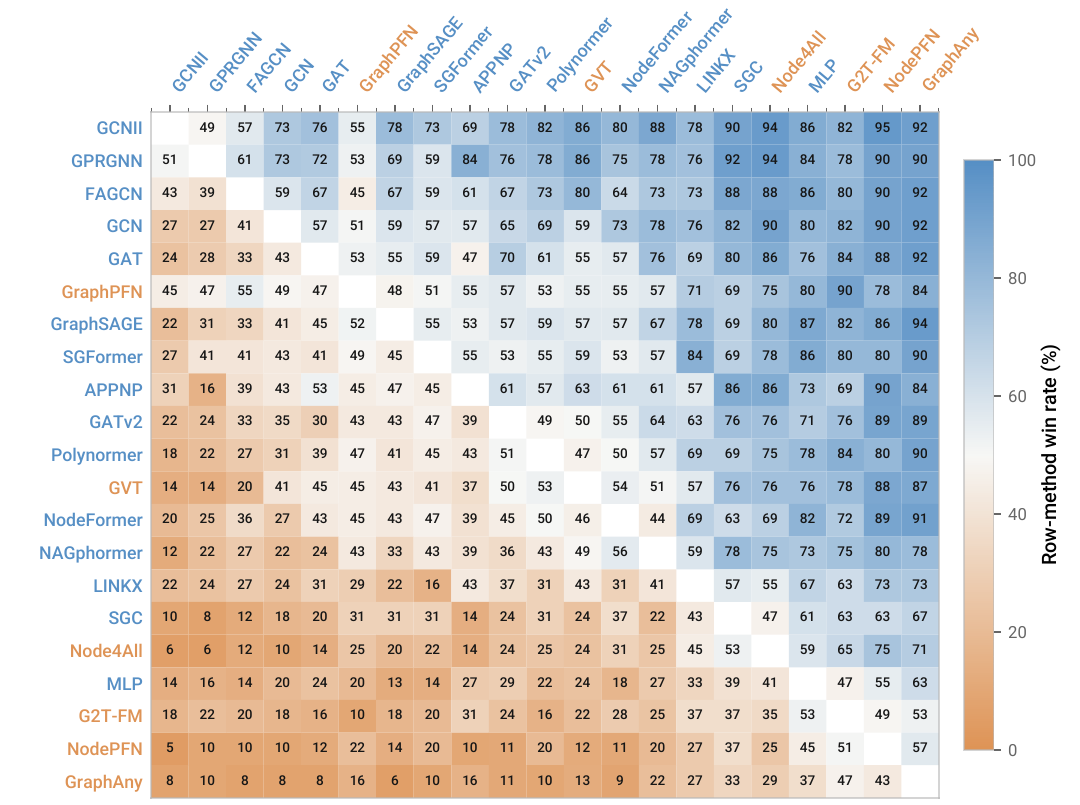}
        \caption{10/10/80 split.}
    \end{subfigure}

    \vspace{0.4em}

    \begin{subfigure}[t]{0.85\textwidth}
        \centering
        \includegraphics[width=\linewidth]
        {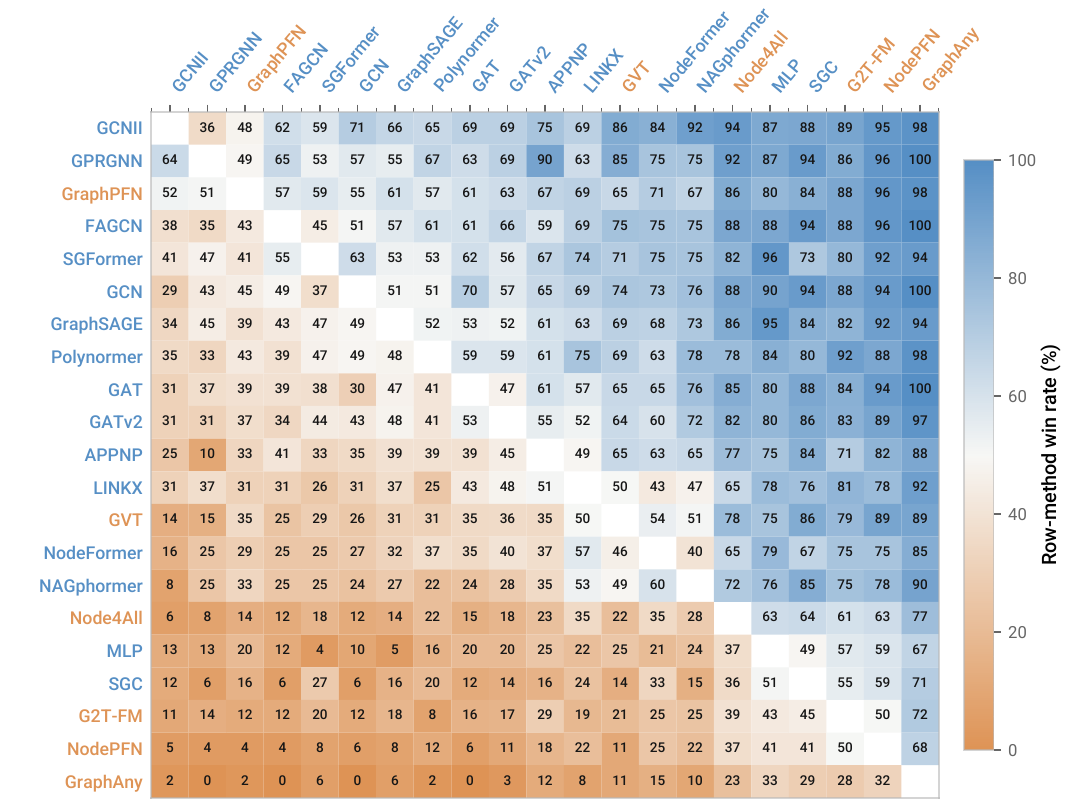}
        \caption{50/25/25 split.}
    \end{subfigure}

    \caption{\textbf{Pairwise primary-metric win rates on Cohort V1.}
    Each cell gives the percentage of the 51 datasets on which the row method
    beats the column method. Methods are ordered separately for each split by
    increasing mean within-dataset primary-metric rank.}
    \label{fig:app-pairwise-significance}
\end{figure*}

\mypar{The Elo leader does not win most often against every competitor.}
\Cref{fig:app-pairwise-significance} shows how often each row method outperforms each column method across datasets.
A win rate above 50\% means that the row method wins more often than its competitor.
Although GCNII ranks first by Elo under both label splits, it does not have a majority of wins against every method.
Under 10/10/80, GPRGNN wins more often than all 20 competitors, including GCNII.
Under 50/25/25, GraphPFN ranks third by Elo but wins more often than every competitor, including GCNII and GPRGNN.

These results reflect different ways of summarizing performance.
Elo fits a single rating to pairwise outcomes across datasets and experimental indices, whereas the win-rate analysis compares scores averaged over the five indices within each dataset.
A narrow majority against a strong competitor can therefore coexist with a lower overall Elo rating.
The pairwise results complement the leaderboard by showing which direct comparisons depart from its overall ordering.

\mypar{Winning more often does not establish a reliable performance advantage.}
A win rate counts a small improvement and a large improvement equally.
A method can therefore win on more datasets while losing by larger margins on others.
To assess whether the performance differences provide evidence of an advantage, we compare each pair's scores on the same datasets using a paired Wilcoxon signed-rank test.
This test considers both the direction and the relative size of the differences without assuming that they follow a normal distribution.
We use the 28 datasets with complete results so that these tests rely entirely on observed scores.

Within each label split, we test all $\binom{21}{2}=210$ method pairs.
Testing so many pairs increases the chance of declaring an advantage by chance.
We therefore apply Holm's correction to keep the probability of any false-positive conclusion across the 210 tests at no more than 5\%, under the test assumptions.
All tests are two-sided, and exact ties are handled using the Pratt procedure.

\begin{table}[t]
\centering
\caption{\textbf{Holm-corrected pairwise comparisons.} Counts are computed on the 28-dataset predictive complete-case cohort, with Holm correction applied jointly across all 210 method pairs within each split. Counts summarize the 20 paired method comparisons for each row method. A win requires a positive median paired effect and adjusted $p<0.05$.}
\small
\setlength{\tabcolsep}{4pt}
\begin{tabular}{@{}lrrrr@{}}
\toprule
& \multicolumn{2}{c}{10/10/80} & \multicolumn{2}{c}{50/25/25} \\
\cmidrule(lr){2-3}\cmidrule(l){4-5}
Method & Significant wins & Significant losses & Significant wins & Significant losses \\
\midrule
GCNII & 11 & 0 & 8 & 0 \\
GraphPFN & 1 & 0 & 7 & 0 \\
GPRGNN & 6 & 0 & 6 & 0 \\
FAGCN & 6 & 0 & 8 & 0 \\
GCN & 5 & 0 & 7 & 0 \\
GraphSAGE & 4 & 0 & 7 & 0 \\
SGFormer & 3 & 0 & 4 & 0 \\
NodeFormer & 4 & 0 & 3 & 0 \\
GAT & 1 & 1 & 5 & 0 \\
GATv2 & 0 & 1 & 6 & 0 \\
GVT & 3 & 1 & 4 & 0 \\
Polynormer & 0 & 2 & 3 & 0 \\
APPNP & 1 & 1 & 1 & 4 \\
LINKX & 0 & 0 & 2 & 0 \\
NAGphormer & 0 & 1 & 1 & 7 \\
MLP & 0 & 3 & 0 & 6 \\
Node4All & 0 & 6 & 1 & 8 \\
NodePFN & 0 & 8 & 0 & 11 \\
SGC & 0 & 4 & 0 & 9 \\
G2T-FM & 0 & 10 & 0 & 12 \\
GraphAny & 0 & 7 & 0 & 16 \\
\bottomrule
\end{tabular}
\label{tab:app-pairwise-significance-summary}
\end{table}

\Cref{tab:app-pairwise-significance-summary} counts the statistically significant wins and losses for each method.
A significant win requires a positive median score difference and a Holm-adjusted $p<0.05$.
Under 10/10/80, GCNII records 11 significant wins and no losses, while GraphPFN records one significant win.
Under 50/25/25, GraphPFN records seven significant wins and no losses; GCNII and FAGCN each record eight, and GPRGNN records six.
These counts use 28 datasets, so they are not directly comparable to the win-rate counts over all 51 datasets above.
GraphPFN has more statistically supported advantages when more labels are available, but no method significantly outperforms every competitor.
The absence of a significant difference also does not establish that two methods perform equally.

\subsection{Tuning Requirements and Model-Selection Robustness}
\label{app:tuning-robustness}

Our main experiments compare GFMs with supervised methods trained and tuned separately for each dataset.
For each supervised method, the search evaluates up to 201 configurations and selects one using validation data.
We now examine how much of this search is needed to obtain strong test performance, and whether improvements on validation nodes carry over to test nodes.
This analysis covers the 15 supervised methods on the 28 datasets with complete results.

We reuse stored trials to evaluate budgets of 1, 2, 5, 10, 25, 50, 100, 150, and 201 attempted configurations.
Each search starts with the default configuration.
We randomly reorder the remaining 200 configurations 20 times and evaluate progressively larger portions of each order.
At every budget, validation cross-entropy or binary cross-entropy selects the configuration.
Failed trials count toward the budget but cannot be selected.
We then measure the selected configuration's performance on both validation and test nodes, using accuracy for multiclass tasks and AUROC for binary tasks.

\mypar{Some methods benefit much more from tuning than others.}
We summarize tuning with two quantities.
The first is the test improvement from the default to the full 201-configuration search.
The second is the smallest evaluated budget that comes within one percentage point of the full-search result.
Letting $P_m(k)$ denote method $m$'s mean test score at budget $k$, these quantities are
\[
    \Delta_m=P_m(201)-P_m(1),
    \qquad
    B_m^{1\mathrm{pp}}
    =
    \min\left\{
        k:P_m(k)\geq P_m(201)-1\text{ percentage point}
    \right\}.
\]

\Cref{tab:app-tuning-efficiency} shows substantial differences between methods.
Under 10/10/80, tuning improves SGC by 1.2 percentage points but Polynormer by 9.4.
GCN, GPRGNN, and Polynormer need 50 configurations to come within one point of their full-search results.
Under 50/25/25, GAT and GATv2 improve by about ten points and reach this threshold at 50 configurations.
MLP improves by only 1.4 points and reaches it after two.
A small tuning budget therefore captures most of the benefit for some methods while understating the performance of others.

\begin{table*}[!tbp]
\centering
\caption{\textbf{Tuning dependence by method.} $\Delta$ is the test-performance gain from 1 to 201 configurations. $B^{1\mathrm{pp}}$ is the smallest sampled budget within one percentage point of the 201-configuration result. Results average the 28-dataset predictive complete-case cohort, five experimental indices, and 20 nested configuration samples. Cost is the workflow multiplier relative to the default, estimated by linear interpolation between the measured default and 201-configuration endpoints.}
\small
\setlength{\tabcolsep}{5.5pt}
\begin{tabular}{@{}lrrr rrr@{}}
\toprule
& \multicolumn{3}{c}{10/10/80} & \multicolumn{3}{c}{50/25/25} \\
\cmidrule(lr){2-4}\cmidrule(l){5-7}
Method & $\Delta$ (pp) & $B^{1\mathrm{pp}}$ & Cost at $B$ ($\times$) & $\Delta$ (pp) & $B^{1\mathrm{pp}}$ & Cost at $B$ ($\times$) \\
\midrule
APPNP & 4.5 & 25 & 14.0 & 6.0 & 25 & 14.8 \\
FAGCN & 3.0 & 10 & 5.6 & 2.9 & 25 & 14.8 \\
GAT & 7.3 & 25 & 13.9 & 10.0 & 50 & 29.4 \\
GATv2 & 6.9 & 25 & 10.2 & 10.0 & 50 & 20.1 \\
GCN & 5.5 & 50 & 27.7 & 8.2 & 25 & 14.7 \\
GCNII & 5.9 & 25 & 13.1 & 5.3 & 25 & 13.9 \\
GPRGNN & 5.1 & 50 & 28.0 & 3.3 & 25 & 14.8 \\
GraphSAGE & 3.6 & 25 & 13.9 & 2.9 & 25 & 14.7 \\
LINKX & 7.3 & 25 & 13.9 & 4.6 & 25 & 14.7 \\
MLP & 1.7 & 5 & 2.8 & 1.4 & 2 & 1.3 \\
NAGphormer & 2.0 & 5 & 2.7 & 2.6 & 25 & 13.8 \\
NodeFormer & 3.3 & 25 & 7.5 & 2.8 & 25 & 7.9 \\
Polynormer & 9.4 & 50 & 28.0 & 4.9 & 25 & 14.8 \\
SGC & 1.2 & 5 & 2.8 & 2.0 & 10 & 5.9 \\
SGFormer & 3.5 & 25 & 14.0 & 2.8 & 25 & 14.7 \\
\bottomrule
\end{tabular}
\label{tab:app-tuning-efficiency}
\end{table*}

The table also reports estimated workflow cost relative to the default.
For example, the 50-configuration budgets above cost approximately $28\times$ the default under 10/10/80.
Intermediate costs are estimated by linear interpolation between the measured default and full-search costs,
\[
    C_k=C_1+\frac{k-1}{200}(C_{201}-C_1).
\]
They are not direct measurements of cumulative trial time.
The main-text performance--cost trajectories use a different set of budgets, as specified in Appendix~\ref{app:overhead-pareto}.

\begin{figure*}[!t]
    \centering
    \includegraphics[width=\linewidth]
    {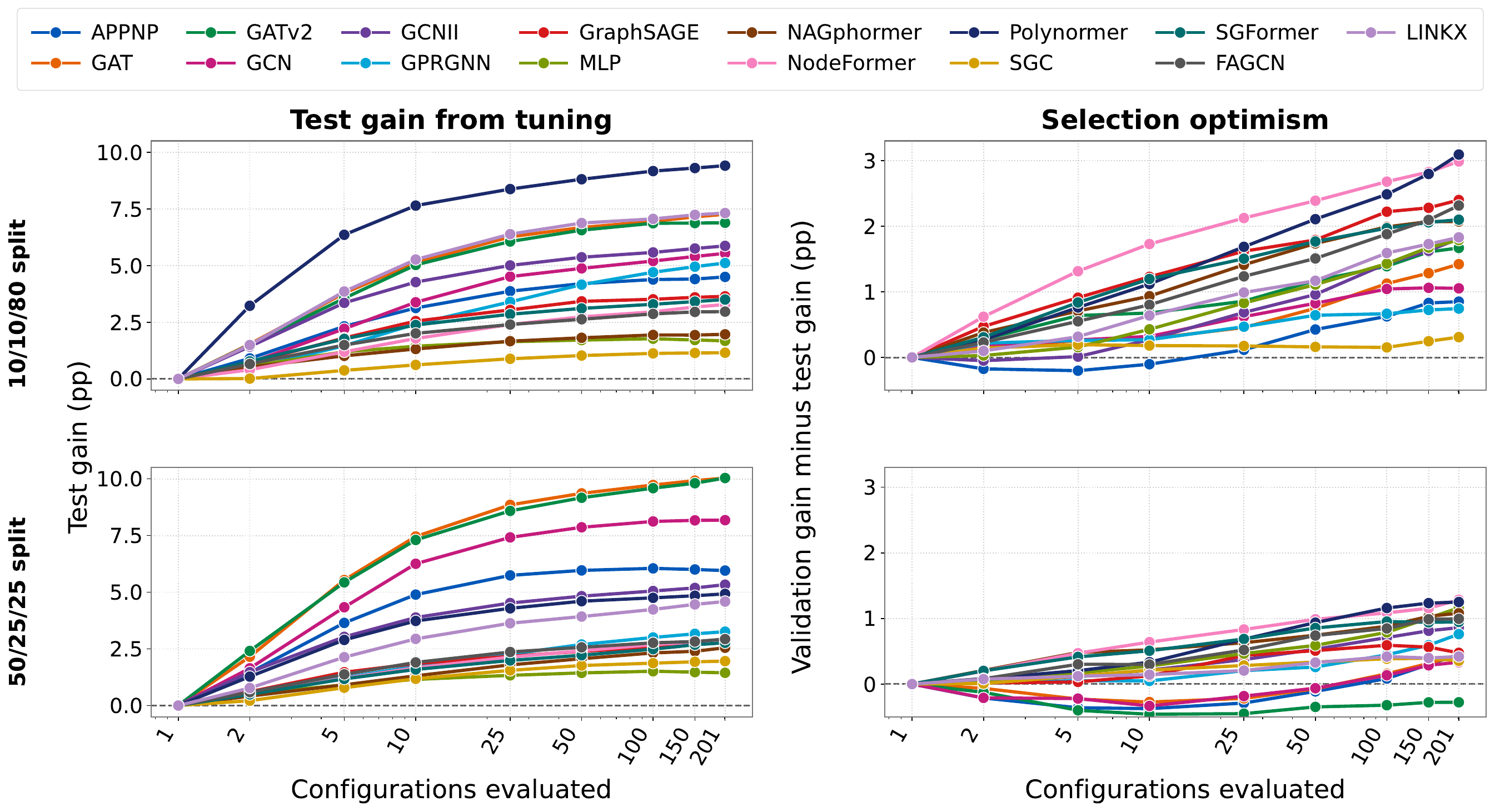}
    \caption{\textbf{How tuning improves test performance and changes the gap between validation and test gains.}
    Left: test-score improvement over the default configuration.
    Right: validation improvement minus test improvement, each measured relative to the default.
    Positive values mean that tuning appears more beneficial on validation nodes than on test nodes.
    All gains are in percentage points (pp).
    Configurations are selected by validation loss; gains use multiclass accuracy or binary AUROC.
    Curves average over 28 datasets, five shared splits, and 20 search orders.}
    \label{fig:app-validation-test-trajectories}
\end{figure*}

\mypar{Most test gains appear within 25--50 configurations.}
The left panels of \Cref{fig:app-validation-test-trajectories} show how test performance improves as more configurations are tried.
Polynormer benefits most under 10/10/80, while GAT and GATv2 benefit most under 50/25/25.
For every method, extending the search from 50 to 201 configurations adds less than one percentage point on average.
Much of the full-search cost therefore buys relatively little additional improvement.
This pattern supports comparing GFMs with partially tuned supervised methods, whose performance can approach the full-search result at substantially lower cost.

\mypar{With fewer labels, validation gains exceed test gains by more.}
The right panels check whether the improvement observed during model selection is also realized on test nodes.
We subtract the test improvement from the validation improvement, both relative to the default.
For example, a five-point validation gain and a two-point test gain produce a gap of three points.
This quantity is called selection optimism:
\[
    O_m(k)
    =
    \bigl[V_m(k)-V_m(1)\bigr]
    -
    \bigl[P_m(k)-P_m(1)\bigr],
\]
where $V_m(k)$ is the mean validation primary metric of the configuration selected at budget $k$.
Zero means that validation and test improve equally; negative values mean that test improves more.

Under 10/10/80, this gap generally grows with the search budget and reaches approximately three points.
Under 50/25/25, it stays below 1.3 points and is sometimes negative.
Thus, with fewer labels, the improvement observed on validation nodes can substantially overstate the improvement on unseen test nodes.
Together, the two columns show why larger searches need careful interpretation: test gains can level off while validation gains continue to grow.

\section{Dataset Curation and Catalog}
\label{app:datasets}

This section documents how the benchmark dataset collection was constructed. It identifies the four candidate sources, applies a fixed sequence of eligibility checks, records every canonicalization decision, and enumerates the 51 retained tasks. It then defines the dataset properties and group assignments used in the main analyses.

\subsection{Candidate Sources}
\label{app:dataset-selection}

We assembled 117 loader records from PyTorch Geometric (PyG), the Hugging Face Text-Attributed-Graphs collection (HF TAG), the Heterophilous Graphs collection, and OGB. Table~\ref{tab:selection-by-channel} reports the initial records, the records remaining after the four eligibility checks, and the canonical tasks retained from each channel.

\begin{table*}[!htbp]
\centering
\caption{\textbf{Candidate records and retained benchmark tasks by discovery channel.} Eligible records pass the scope, target, feature, and scale checks. Final tasks are counted after canonicalization.}
\label{tab:selection-by-channel}
\scriptsize
\setlength{\tabcolsep}{5pt}
\renewcommand{\arraystretch}{0.96}
\begin{tabular}{lrrr}
\toprule
Source channel & Candidate records & Eligible records & Retained tasks \\
\midrule
PyG & 94 & 48 & 32 \\
HF TAG & 11 & 11 & 11 \\
Heterophilous & 11 & 7 & 7 \\
OGB & 1 & 1 & 1 \\
\midrule
Total & 117 & 67 & 51 \\
\bottomrule
\end{tabular}
\end{table*}

\subsection{Ordered Screening, Canonicalization, and Task Identity}
\label{app:dataset-canonicalization}
\label{app:candidate-ledger}

Every candidate passes through five checks in a fixed order. The scope check requires homogeneous, single-domain, single-graph node classification. The target check requires one verified categorical label per evaluated node. The feature check requires observable, label-independent node information supplied upstream or produced by a documented transformation. Missing, identity-based, and topology-only features are excluded. The scale check enforces the graph and feature limits. Finally, canonicalization merges eligible loader records
that represent the same benchmark task. Table~\ref{tab:selection-funnel} defines each decision and shows how the pool decreases from 117 records to 51 tasks.

\begin{table*}[!htbp]
\centering
\caption{\textbf{Ordered candidate-screening criteria and resulting funnel.} Each record is assigned to the first check that it fails.}
\label{tab:selection-funnel}
\scriptsize
\setlength{\tabcolsep}{2.8pt}
\renewcommand{\arraystretch}{0.96}
\begin{tabularx}{\textwidth}{l>{\raggedright\arraybackslash}Xrr>{\raggedright\arraybackslash}p{0.25\textwidth}}
\toprule
Check & Criterion & Excluded & Retained & Typical exclusions or merges \\
\midrule
Scope & Homogeneous, single-domain, single-graph node classification. & 16 & 101 & PPI, WikipediaNetwork:Crocodile, Twitch:DE \\
Target & Verified categorical single label per evaluated node. & 7 & 94 & Yelp, AttributedGraphDataset:PPI, AttributedGraphDataset:Facebook \\
Feature & Eligible observable node attributes; no missing, identity-only, topology-only, or disallowed legacy representations. & 15 & 79 & KarateClub, Roman-empire, Minesweeper \\
Scale & $F\leq10{,}000$; $|V|+|E|\leq1.7\times10^6$; $|V||E|\leq5\times10^{11}$. & 12 & 67 & NELL, Reddit, Reddit2 \\
Canonicalization & Merge eligible loader records representing the same benchmark task. & 16 & 51 & CitationFull:Cora, CitationFull:PubMed, AttributedGraphDataset:Cora \\
\bottomrule
\end{tabularx}
\end{table*}

The first four checks retain 67 eligible records. Fifteen alias groups then consolidate 16 redundant records, with most groups joining two loader records and \texttt{pubmed} joining three. Records are merged only when their graph structure, node attributes, and target semantics identify the same task. Table~\ref{tab:alias-groups} lists every merged alias. Since some datasets are exposed through multiple loaders, we merge records only when they provide the same graph, node features, and labels. We do not merge \texttt{cora}, \texttt{pubmed}, and \texttt{wiki\_cs} with their respective TAG variants because each pair shares the same graph topology but provides different node features. They therefore remain separate benchmark tasks, yielding 51 tasks over 48 distinct graph topologies.

\begin{table*}[t]
\centering
\caption{\textbf{Eligible loader-record aliases consolidated into canonical benchmark tasks.}}
\label{tab:alias-groups}
\scriptsize
\setlength{\tabcolsep}{2.5pt}
\renewcommand{\arraystretch}{0.92}
\begin{tabularx}{\textwidth}{lr>{\raggedright\arraybackslash}X}
\toprule
Canonical task & Records & Merged loader records \\
\midrule
actor & 2 & Actor; heterophilous\_graph:actor \\
amazon\_ratings & 2 & HeterophilousGraphDataset:Amazon-ratings; heterophilous\_graph:amazon\_ratings \\
chameleon & 2 & WikipediaNetwork:Chameleon; heterophilous\_graph:chameleon \\
citeseer & 2 & Planetoid:CiteSeer; AttributedGraphDataset:CiteSeer \\
cora & 2 & Planetoid:Cora; AttributedGraphDataset:Cora \\
cornell & 2 & WebKB:Cornell; heterophilous\_graph:cornell \\
deezer & 2 & DeezerEurope; local\_raw\_fallback:deezer \\
elliptic\_bitcoin & 2 & EllipticBitcoinDataset; EllipticBitcoinTemporalDataset \\
facebook\_large & 2 & FacebookPagePage; local\_raw\_fallback:facebook\_large \\
full\_cora & 2 & CitationFull:Cora; CoraFull \\
lastfm\_asia & 2 & LastFMAsia; local\_raw\_fallback:lastfm\_asia \\
pubmed & 3 & Planetoid:PubMed; CitationFull:PubMed; AttributedGraphDataset:PubMed \\
squirrel & 2 & WikipediaNetwork:Squirrel; heterophilous\_graph:squirrel \\
texas & 2 & WebKB:Texas; heterophilous\_graph:texas \\
wisconsin & 2 & WebKB:Wisconsin; heterophilous\_graph:wisconsin \\
\bottomrule
\end{tabularx}
\end{table*}

\subsection{Retained Dataset Catalog}
\label{app:dataset-catalog}
\begin{table*}[t]
\centering
\caption{\textbf{Catalog of the 51 retained node-classification tasks.} Channel labels link to upstream sources. Citations identify the corresponding data or benchmark publications. License entries reproduce terms stated by the linked upstream repository or data release. ``Not stated'' means that no explicit license was found at that source; it does not imply that research use is prohibited. \method does not redistribute or relicense these datasets; our pipeline downloads data from upstream sources, preprocesses them locally, and reports evaluation results. The four road-network datasets derive from \href{https://www.openstreetmap.org/copyright}{OpenStreetMap contributors (ODbL 1.0)}; their processed release states no separate license.}
\label{tab:dataset-catalog}
\tiny
\setlength{\tabcolsep}{1.6pt}
\renewcommand{\arraystretch}{0.78}
\begin{tabularx}{\textwidth}{lll>{\raggedright\arraybackslash}Xll}
\toprule
Dataset & Channel & Domain & Task & Citation & License \\
\midrule
actor & \href{https://github.com/yandex-research/heterophilous-graphs}{Heteroph.} & Web & Actor category classification & \citep{pei2020geom} & \href{https://github.com/yandex-research/heterophilous-graphs/blob/main/LICENCE.txt}{MIT} \\
amazon\_computer & \href{https://github.com/shchur/gnn-benchmark}{PyG} & Commerce & Product category classification & \citep{mcauley2015image,shchur2018pitfalls} & \href{https://github.com/shchur/gnn-benchmark/blob/master/LICENSE}{MIT} \\
amazon\_photo & \href{https://github.com/shchur/gnn-benchmark}{PyG} & Commerce & Product category classification & \citep{mcauley2015image,shchur2018pitfalls} & \href{https://github.com/shchur/gnn-benchmark/blob/master/LICENSE}{MIT} \\
amazon\_ratings & \href{https://github.com/yandex-research/heterophilous-graphs}{Heteroph.} & Commerce & Binned mean product-rating classification & \citep{platonov2023critical} & \href{https://github.com/yandex-research/heterophilous-graphs/blob/main/LICENCE.txt}{MIT} \\
amherst41 & \href{https://github.com/CUAI/Non-Homophily-Large-Scale}{PyG} & Social & Student gender classification & \citep{traud2012social,lim2021large} & \href{https://github.com/CUAI/Non-Homophily-Large-Scale}{not stated} \\
artnet-exp & \href{https://zenodo.org/records/16895532}{PyG} & Social & Explicit-content creator classification & \citep{bazhenov2026graphland} & \href{https://zenodo.org/records/16895532}{Apache-2.0} \\
blogcatalog & \href{https://pytorch-geometric.readthedocs.io/en/latest/generated/torch_geometric.datasets.AttributedGraphDataset.html}{PyG} & Social & Blogger topic classification & \citep{li2015unsupervised,huang2017label} & \href{https://pytorch-geometric.readthedocs.io/en/latest/generated/torch_geometric.datasets.AttributedGraphDataset.html}{not stated} \\
chameleon & \href{https://github.com/yandex-research/heterophilous-graphs}{Heteroph.} & Web & Page-traffic quantile classification & \citep{rozemberczki2021multi,platonov2023critical} & \href{https://github.com/yandex-research/heterophilous-graphs/blob/main/LICENCE.txt}{MIT} \\
citation\_citeseer & \href{https://github.com/abojchevski/graph2gauss}{PyG} & Scholarly & Scholarly topic or field classification & \citep{bojchevski2017deep} & \href{https://github.com/abojchevski/graph2gauss/blob/master/LICENSE}{MIT} \\
citeseer & \href{https://github.com/kimiyoung/planetoid}{PyG} & Scholarly & Scholarly topic or field classification & \citep{sen2008collective,yang2016revisiting} & \href{https://github.com/kimiyoung/planetoid/blob/master/LICENSE}{MIT} \\
city-reviews & \href{https://zenodo.org/records/16895532}{PyG} & Commerce & Fraudulent-reviewer classification & \citep{bazhenov2026graphland} & \href{https://zenodo.org/records/16895532}{Apache-2.0} \\
coauthor\_cs & \href{https://github.com/shchur/gnn-benchmark}{PyG} & Scholarly & Scholarly topic or field classification & \citep{shchur2018pitfalls} & \href{https://github.com/shchur/gnn-benchmark/blob/master/LICENSE}{MIT} \\
coauthor\_physics & \href{https://github.com/shchur/gnn-benchmark}{PyG} & Scholarly & Scholarly topic or field classification & \citep{shchur2018pitfalls} & \href{https://github.com/shchur/gnn-benchmark/blob/master/LICENSE}{MIT} \\
cora & \href{https://github.com/kimiyoung/planetoid}{PyG} & Scholarly & Scholarly topic or field classification & \citep{sen2008collective,yang2016revisiting} & \href{https://github.com/kimiyoung/planetoid/blob/master/LICENSE}{MIT} \\
cora\_ml & \href{https://github.com/abojchevski/graph2gauss}{PyG} & Scholarly & Scholarly topic or field classification & \citep{bojchevski2017deep} & \href{https://github.com/abojchevski/graph2gauss/blob/master/LICENSE}{MIT} \\
cornell & \href{https://github.com/yandex-research/heterophilous-graphs}{Heteroph.} & Web & Web-page category classification & \citep{craven1998learning,pei2020geom} & \href{https://github.com/yandex-research/heterophilous-graphs/blob/main/LICENCE.txt}{MIT} \\
cornell5 & \href{https://github.com/CUAI/Non-Homophily-Large-Scale}{PyG} & Social & Student gender classification & \citep{traud2012social,lim2021large} & \href{https://github.com/CUAI/Non-Homophily-Large-Scale}{not stated} \\
dblp & \href{https://github.com/abojchevski/graph2gauss}{PyG} & Scholarly & Scholarly topic or field classification & \citep{bojchevski2017deep} & \href{https://github.com/abojchevski/graph2gauss/blob/master/LICENSE}{MIT} \\
deezer & \href{https://snap.stanford.edu/data/feather-deezer-social.html}{PyG} & Social & User gender classification & \citep{rozemberczki2020characteristic} & \href{https://snap.stanford.edu/data/feather-deezer-social.html}{not stated} \\
elliptic\_bitcoin & \href{https://www.kaggle.com/datasets/ellipticco/elliptic-data-set}{PyG} & Finance & Licit versus illicit transaction classification & \citep{weber2019anti} & \href{https://www.kaggle.com/datasets/ellipticco/elliptic-data-set}{CC BY-NC-ND 4.0} \\
facebook\_large & \href{https://github.com/benedekrozemberczki/MUSAE}{PyG} & Social & Facebook page category classification & \citep{rozemberczki2021multi} & \href{https://github.com/benedekrozemberczki/MUSAE/blob/master/LICENSE}{GPL-3.0} \\
flickr & \href{https://github.com/GraphSAINT/GraphSAINT}{PyG} & Social & Image-tag classification & \citep{zeng2019graphsaint} & \href{https://github.com/GraphSAINT/GraphSAINT/blob/master/LICENSE}{MIT} \\
full\_cora & \href{https://github.com/abojchevski/graph2gauss}{PyG} & Scholarly & Scholarly topic or field classification & \citep{bojchevski2017deep} & \href{https://github.com/abojchevski/graph2gauss/blob/master/LICENSE}{MIT} \\
genius & \href{https://github.com/CUAI/Non-Homophily-Large-Scale}{PyG} & Social & Marked-user-account classification & \citep{lim2021large} & \href{https://github.com/CUAI/Non-Homophily-Large-Scale}{not stated} \\
johnshopkins55 & \href{https://github.com/CUAI/Non-Homophily-Large-Scale}{PyG} & Social & Student gender classification & \citep{traud2012social,lim2021large} & \href{https://github.com/CUAI/Non-Homophily-Large-Scale}{not stated} \\
la & \href{https://github.com/LeonResearch/City-Networks}{PyG} & Transport & 10-quantile road-junction eccentricity classification & \citep{liang2026towards} & \href{https://github.com/LeonResearch/City-Networks}{not stated} \\
lastfm\_asia & \href{https://snap.stanford.edu/data/feather-lastfm-social.html}{PyG} & Social & User home-country classification & \citep{rozemberczki2020characteristic} & \href{https://snap.stanford.edu/data/feather-lastfm-social.html}{not stated} \\
london & \href{https://github.com/LeonResearch/City-Networks}{PyG} & Transport & 10-quantile road-junction eccentricity classification & \citep{liang2026towards} & \href{https://github.com/LeonResearch/City-Networks}{not stated} \\
ogbn\_arxiv & \href{https://ogb.stanford.edu/docs/nodeprop/\#ogbn-arxiv}{OGB} & Scholarly & Scholarly topic or field classification & \citep{hu2020open} & \href{https://ogb.stanford.edu/docs/nodeprop/\#ogbn-arxiv}{ODC-BY 1.0} \\
paris & \href{https://github.com/LeonResearch/City-Networks}{PyG} & Transport & 10-quantile road-junction eccentricity classification & \citep{liang2026towards} & \href{https://github.com/LeonResearch/City-Networks}{not stated} \\
penn94 & \href{https://github.com/CUAI/Non-Homophily-Large-Scale}{PyG} & Social & Student gender classification & \citep{traud2012social,lim2021large} & \href{https://github.com/CUAI/Non-Homophily-Large-Scale}{not stated} \\
pubmed & \href{https://github.com/kimiyoung/planetoid}{PyG} & Scholarly & Scholarly topic or field classification & \citep{namata2012query,yang2016revisiting} & \href{https://github.com/kimiyoung/planetoid/blob/master/LICENSE}{MIT} \\
reed98 & \href{https://github.com/CUAI/Non-Homophily-Large-Scale}{PyG} & Social & Student gender classification & \citep{traud2012social,lim2021large} & \href{https://github.com/CUAI/Non-Homophily-Large-Scale}{not stated} \\
shanghai & \href{https://github.com/LeonResearch/City-Networks}{PyG} & Transport & 10-quantile road-junction eccentricity classification & \citep{liang2026towards} & \href{https://github.com/LeonResearch/City-Networks}{not stated} \\
squirrel & \href{https://github.com/yandex-research/heterophilous-graphs}{Heteroph.} & Web & Page-traffic quantile classification & \citep{rozemberczki2021multi,platonov2023critical} & \href{https://github.com/yandex-research/heterophilous-graphs/blob/main/LICENCE.txt}{MIT} \\
tag\_bookchild & \href{https://huggingface.co/datasets/Graph-COM/Text-Attributed-Graphs}{HF TAG} & Commerce & Product category classification & \citep{wang2025generalization} & \href{https://huggingface.co/datasets/Graph-COM/Text-Attributed-Graphs}{Apache-2.0} \\
tag\_bookhis & \href{https://huggingface.co/datasets/Graph-COM/Text-Attributed-Graphs}{HF TAG} & Commerce & Product category classification & \citep{wang2025generalization} & \href{https://huggingface.co/datasets/Graph-COM/Text-Attributed-Graphs}{Apache-2.0} \\
tag\_citeseer & \href{https://huggingface.co/datasets/Graph-COM/Text-Attributed-Graphs}{HF TAG} & Scholarly & Scholarly topic or field classification & \citep{wang2025generalization,sen2008collective} & \href{https://huggingface.co/datasets/Graph-COM/Text-Attributed-Graphs}{Apache-2.0} \\
tag\_cora & \href{https://huggingface.co/datasets/Graph-COM/Text-Attributed-Graphs}{HF TAG} & Scholarly & Scholarly topic or field classification & \citep{wang2025generalization,sen2008collective} & \href{https://huggingface.co/datasets/Graph-COM/Text-Attributed-Graphs}{Apache-2.0} \\
tag\_cornell & \href{https://huggingface.co/datasets/Graph-COM/Text-Attributed-Graphs}{HF TAG} & Web & Web-page category classification & \citep{wang2025generalization,craven1998learning} & \href{https://huggingface.co/datasets/Graph-COM/Text-Attributed-Graphs}{Apache-2.0} \\
tag\_pubmed & \href{https://huggingface.co/datasets/Graph-COM/Text-Attributed-Graphs}{HF TAG} & Scholarly & Scholarly topic or field classification & \citep{wang2025generalization,namata2012query} & \href{https://huggingface.co/datasets/Graph-COM/Text-Attributed-Graphs}{Apache-2.0} \\
tag\_sportsfit & \href{https://huggingface.co/datasets/Graph-COM/Text-Attributed-Graphs}{HF TAG} & Commerce & Product category classification & \citep{wang2025generalization} & \href{https://huggingface.co/datasets/Graph-COM/Text-Attributed-Graphs}{Apache-2.0} \\
tag\_texas & \href{https://huggingface.co/datasets/Graph-COM/Text-Attributed-Graphs}{HF TAG} & Web & Web-page category classification & \citep{wang2025generalization,craven1998learning} & \href{https://huggingface.co/datasets/Graph-COM/Text-Attributed-Graphs}{Apache-2.0} \\
tag\_washington & \href{https://huggingface.co/datasets/Graph-COM/Text-Attributed-Graphs}{HF TAG} & Web & Web-page category classification & \citep{wang2025generalization,craven1998learning} & \href{https://huggingface.co/datasets/Graph-COM/Text-Attributed-Graphs}{Apache-2.0} \\
tag\_wikics & \href{https://huggingface.co/datasets/Graph-COM/Text-Attributed-Graphs}{HF TAG} & Web & Wikipedia page category classification & \citep{wang2025generalization,mernyei2020wiki} & \href{https://huggingface.co/datasets/Graph-COM/Text-Attributed-Graphs}{Apache-2.0} \\
tag\_wisconsin & \href{https://huggingface.co/datasets/Graph-COM/Text-Attributed-Graphs}{HF TAG} & Web & Web-page category classification & \citep{wang2025generalization,craven1998learning} & \href{https://huggingface.co/datasets/Graph-COM/Text-Attributed-Graphs}{Apache-2.0} \\
texas & \href{https://github.com/yandex-research/heterophilous-graphs}{Heteroph.} & Web & Web-page category classification & \citep{craven1998learning,pei2020geom} & \href{https://github.com/yandex-research/heterophilous-graphs/blob/main/LICENCE.txt}{MIT} \\
tolokers-2 & \href{https://zenodo.org/records/16895532}{PyG} & Social & Banned-worker classification & \citep{bazhenov2026graphland} & \href{https://zenodo.org/records/16895532}{Apache-2.0} \\
wiki & \href{https://pytorch-geometric.readthedocs.io/en/latest/generated/torch_geometric.datasets.AttributedGraphDataset.html}{PyG} & Web & Wikipedia page category classification & \citep{yang2020scaling} & \href{https://pytorch-geometric.readthedocs.io/en/latest/generated/torch_geometric.datasets.AttributedGraphDataset.html}{not stated} \\
wiki\_cs & \href{https://github.com/pmernyei/wiki-cs-dataset}{PyG} & Web & Wikipedia page category classification & \citep{mernyei2020wiki} & \href{https://github.com/pmernyei/wiki-cs-dataset/blob/master/LICENSE}{MIT} \\
wisconsin & \href{https://github.com/yandex-research/heterophilous-graphs}{Heteroph.} & Web & Web-page category classification & \citep{craven1998learning,pei2020geom} & \href{https://github.com/yandex-research/heterophilous-graphs/blob/main/LICENCE.txt}{MIT} \\
\bottomrule
\end{tabularx}
\end{table*}

\Cref{tab:dataset-catalog} documents the 51 retained tasks, providing each task's source, application domain, prediction target, citation, and upstream license information.
Links in the table allow readers to trace these entries to their original sources.

We assign application domains based on the meaning of the nodes, their connections, and the prediction targets.
Social datasets cover people, accounts, organizations, and workers, while scholarly datasets represent publications or authors connected through citations or coauthorship.
Web datasets concern pages and hyperlinks, and commerce datasets cover products, reviews, and users.
Transportation and finance datasets represent road networks and financial transactions, respectively.

For 49 of the 51 tasks, \method{} retrieves the data from the listed upstream sources and preprocesses them locally into the shared evaluation format.
The remaining two tasks, \texttt{deezer} and \texttt{lastfm\_asia}, originally relied on \texttt{graphmining.ai} download endpoints through PyG, but these endpoints were unavailable when the benchmark release was prepared.
To keep these tasks accessible and reproducible, we provide copies through a pinned Hugging Face release, together with their upstream provenance, available license information, and file checksums.

\subsection{Dataset Properties}
\label{app:dataset-definitions}

The RQ2 analysis compares performance across 22 overlapping subgroups defined by seven dataset properties.
These cover graph scale (node count), topology (average degree and adjusted homophily), features (dimension and nonzero fraction), and targets (class count and class imbalance).
\Cref{tab:property-definitions} defines each property, gives the subgroup boundaries, and reports the number of datasets in each subgroup. 
Each dataset belongs to one subgroup per property.
For example, a graph can have many nodes, low average degree, and binary targets at the same time.
Subgroups therefore overlap across properties, while those within a property are mutually exclusive.
The boundaries are fixed and shared across both label regimes.

Structural statistics are computed on a simple undirected graph with self-loops removed and duplicate edge orientations counted once.
Node count and feature dimension use the shared graph and feature representation.
Class counts and imbalance use valid labels, while adjusted homophily uses edges whose endpoints are both labeled.
For adjusted homophily, $h$ is the fraction of these edges joining nodes of the same class, and $h_0$ is the agreement expected from the class proportions at their endpoints.
The feature nonzero fraction is measured on $\min(N,4096)$ rows sampled uniformly without replacement, counting finite entries only, with a SHA256-derived seed based on 20260806.

\Cref{tab:additional-property-definitions} defines the remaining statistics used to describe dataset coverage in \Cref{fig:marginal-coverage,fig:joint-coverage}.
These include density, degree quantiles, hub concentration, feature-similarity AUROC, and label entropy.
They are retained as descriptive statistics but do not define the subgroups in \Cref{fig:gfm-groupwise-ranks}.

\begin{table*}[!ht]
\centering
\caption{\textbf{Property definitions and subgroup boundaries for RQ2.}
Seven properties define the 22 subgroups in \Cref{fig:gfm-groupwise-ranks}.
Each dataset belongs to one subgroup per property; \#data gives the subgroup size.}
\label{tab:property-definitions}
\vspace{6pt}
\begingroup
\small
\setlength{\tabcolsep}{0pt}
\renewcommand{\arraystretch}{1.15}
\newcommand{\ngproperty}[3]{\textbf{#1}\par\vspace{2pt}#2\par #3}
\newcommand{\ngbins}[1]{%
  \begin{tabular}[t]{@{}%
    >{\raggedright\arraybackslash}p{0.23\linewidth}%
    @{\hspace{4pt}}%
    >{\raggedright\arraybackslash}p{0.61\linewidth}%
    @{\hspace{4pt}}%
    >{\raggedleft\arraybackslash}p{\dimexpr0.16\linewidth-8pt\relax}@{}}%
  #1%
  \end{tabular}}
\begin{tabular}{@{}%
  >{\raggedright\arraybackslash}p{0.43\textwidth}%
  @{\hspace{14pt}}%
  >{\raggedright\arraybackslash}p{\dimexpr0.57\textwidth-14pt\relax}@{}}
\toprule
\textbf{Property and definition}
& \ngbins{\textbf{Subgroup} & \textbf{Boundary} & \textbf{\#data} \\} \\
\midrule
\ngproperty{Number of nodes}{$N=|V|$}{Number of graph nodes.}
& \ngbins{Tiny & $N\leq 2{,}000$ & 9 \\
Small & $2{,}000<N\leq 10{,}000$ & 14 \\
Medium & $10{,}000<N\leq 100{,}000$ & 19 \\
Large & $N>100{,}000$ & 9 \\} \\
\midrule
\ngproperty{Average degree}{$\bar d=2E/N$}{Mean number of neighbors per node.}
& \ngbins{Low & $\bar d\leq 5$ & 20 \\
Medium & $5<\bar d\leq 20$ & 18 \\
High & $\bar d>20$ & 13 \\} \\
\midrule
\ngproperty{Label homophily}{$h_{\mathrm{adj}}=(h-h_0)/(1-h_0)$}{Label agreement along edges, adjusted for chance.}
& \ngbins{$<0$ & $h_{\mathrm{adj}}<0$ & 8 \\
$0$--$0.5$ & $0\leq h_{\mathrm{adj}}<0.5$ & 15 \\
$\geq 0.5$ & $h_{\mathrm{adj}}\geq 0.5$ & 28 \\} \\
\midrule
\ngproperty{Number of features}{$F$}{Number of columns in the node-feature matrix.}
& \ngbins{Low & $F\leq 500$ & 15 \\
Medium & $500<F<5{,}000$ & 31 \\
High & $F\geq 5{,}000$ & 5 \\} \\
\midrule
\ngproperty{Feature nonzero fraction}{$s$}{Fraction of sampled finite feature entries that are nonzero.}
& \ngbins{Sparse & $s\leq 0.01$ & 16 \\
Medium & $0.01<s\leq 0.5$ & 17 \\
Dense & $s>0.5$ & 18 \\} \\
\midrule
\ngproperty{Number of classes}{$C$}{Number of nonempty classes among labeled nodes.}
& \ngbins{Binary & $C=2$ & 11 \\
3--10 & $3\leq C\leq 10$ & 32 \\
$>10$ & $C>10$ & 8 \\} \\
\midrule
\ngproperty{Class imbalance}{$r=\max_c n_c/\min_c n_c$}{Ratio of the largest to smallest nonempty class size.}
& \ngbins{Low & $r\leq 2$ & 15 \\
Medium & $2<r\leq 10$ & 23 \\
High & $r>10$ & 13 \\} \\
\bottomrule
\end{tabular}
\endgroup
\end{table*}

\begin{table*}[!ht]
\centering
\caption{\textbf{Additional statistics for describing dataset coverage.}
These describe dataset coverage but do not define the RQ2 subgroups.}
\label{tab:additional-property-definitions}
\vspace{6pt}
\begingroup
\small
\setlength{\tabcolsep}{6pt}
\renewcommand{\arraystretch}{1.15}
\begin{tabularx}{\textwidth}{@{}%
  >{\raggedright\arraybackslash}p{0.27\textwidth}%
  >{\raggedright\arraybackslash}X@{}}
\toprule
\textbf{Quantity} & \textbf{Definition} \\
\midrule
Edge count & $E$: number of unique undirected non-self-loop edges.
Reciprocal orientations are counted once. \\
\addlinespace[5pt]
Density & $\rho=2E/[N(N-1)]$ for $N\geq 2$; zero for $N<2$. \\
\addlinespace[5pt]
Degree quantiles & $q_{50}$ and $q_{99}$: median and 99th percentile of nonzero node degrees. \\
\addlinespace[5pt]
Hub concentration & $q_{99}/q_{50}$: ratio of the 99th-percentile to the median nonzero degree. \\
\addlinespace[5pt]
Feature-similarity AUROC & AUROC of cosine feature similarity for distinguishing same-label from different-label node pairs. Sampling is specified below. \\
\addlinespace[5pt]
Labeled-node count & $N_L$: number of nodes with valid target labels. \\
\addlinespace[5pt]
Normalized label entropy & $H(y)=-\sum_c p_c\log p_c/\log C$, where $p_c$ is the fraction of labeled nodes in class $c$. \\
\addlinespace[5pt]
Feature representation & \textit{Binary-like}: $b>0.999$.\newline
\textit{Dense}: $s>0.99$ and $b<0.01$.\newline
\textit{Mixed/sparse}: all other cases. \\
\bottomrule
\end{tabularx}
\par\vspace{6pt}
\begin{minipage}{\textwidth}
\small\raggedright
\textbf{Feature-similarity AUROC: measurement protocol.}
\begin{enumerate}
\setlength{\itemsep}{2pt}
\setlength{\parsep}{0pt}
\setlength{\topsep}{3pt}
\item \textit{Eligible nodes.} Use labeled nodes with finite, nonzero feature vectors.
\item \textit{Pair sampling.} Draw 20,000 same-label and 20,000 different-label pairs.
Balance same-label quotas across classes and different-label quotas across unordered class pairs.
\item \textit{Repeated draws.} Sample with replacement across pairs; the two nodes within each same-label pair must be distinct.
\item \textit{Random seed.} Derive the seed using SHA256 from 20260806 and the dataset name.
\end{enumerate}
\vspace{3pt}
\textbf{Feature representation.}
$b$ is the fraction of sampled finite feature entries equal to zero or one;
$s$ is the fraction that are nonzero.
These representation categories are descriptive and differ from the Sparse/Medium/Dense subgroups in \Cref{tab:property-definitions}.
\end{minipage}
\endgroup
\end{table*}

\subsection{Per-Dataset Statistics}
\label{app:dataset-statistics}

\Cref{tab:dataset-statistics,tab:dataset-target-statistics} report the seven grouping properties for every dataset, with each numerical value placed beside its subgroup.
The first table covers node count, average degree, and adjusted homophily; the second covers feature dimension, feature nonzero fraction, class count, and class imbalance.
For example, \texttt{coauthor\_cs} has 6,805 features and 15 classes, placing it in the High feature-dimension group and the $>10$ class-count group.
\Cref{tab:dataset-additional-statistics} reports the additional descriptive statistics.

\begin{table*}[!p]
\centering
\caption{\textbf{Structural values and subgroup assignments.}
Each property has a value column and a group column.
Groups follow \Cref{tab:property-definitions}; homophily group $0$--$0.5$ means $0\leq h_{\mathrm{adj}}<0.5$.
The displayed values are rounded.}
\label{tab:dataset-statistics}
\footnotesize
\setlength{\tabcolsep}{2.2pt}
\renewcommand{\arraystretch}{1.03}
\begin{tabular*}{\textwidth}{@{\extracolsep{\fill}}lrlrlrl@{}}
\toprule
& \multicolumn{2}{c}{\textbf{Number of nodes}}
& \multicolumn{2}{c}{\textbf{Average degree}}
& \multicolumn{2}{c}{\textbf{Label homophily}} \\
\cmidrule(lr){2-3}\cmidrule(lr){4-5}\cmidrule(lr){6-7}
\textbf{Dataset} & $N$ & Group & $\bar d$ & Group & $h_{\mathrm{adj}}$ & Group \\
\midrule
actor & 7,600 & Small & 7.02 & Medium & 0.003 & $0$--$0.5$ \\
amazon\_computer & 13,752 & Medium & 35.76 & High & 0.682 & $\geq 0.5$ \\
amazon\_photo & 7,650 & Small & 31.13 & High & 0.785 & $\geq 0.5$ \\
amazon\_ratings & 24,492 & Medium & 7.60 & Medium & 0.140 & $0$--$0.5$ \\
amherst41 & 2,235 & Small & 81.39 & High & 0.060 & $0$--$0.5$ \\
artnet-exp & 50,405 & Medium & 11.12 & Medium & 0.155 & $0$--$0.5$ \\
blogcatalog & 5,196 & Small & 66.11 & High & 0.272 & $0$--$0.5$ \\
chameleon & 890 & Tiny & 19.90 & Medium & 0.030 & $0$--$0.5$ \\
citation\_citeseer & 4,230 & Small & 2.52 & Low & 0.938 & $\geq 0.5$ \\
citeseer & 3,327 & Small & 2.74 & Low & 0.671 & $\geq 0.5$ \\
city-reviews & 148,801 & Large & 15.66 & Medium & 0.591 & $\geq 0.5$ \\
coauthor\_cs & 18,333 & Medium & 8.93 & Medium & 0.785 & $\geq 0.5$ \\
coauthor\_physics & 34,493 & Medium & 14.38 & Medium & 0.872 & $\geq 0.5$ \\
cora & 2,708 & Small & 3.90 & Low & 0.771 & $\geq 0.5$ \\
cora\_ml & 2,995 & Small & 5.45 & Medium & 0.749 & $\geq 0.5$ \\
cornell & 183 & Tiny & 3.03 & Low & -0.220 & $<0$ \\
cornell5 & 18,660 & Medium & 84.76 & High & 0.091 & $0$--$0.5$ \\
dblp & 17,716 & Medium & 5.97 & Medium & 0.679 & $\geq 0.5$ \\
deezer & 28,281 & Medium & 6.56 & Medium & 0.030 & $0$--$0.5$ \\
elliptic\_bitcoin & 203,769 & Large & 2.30 & Low & 0.516 & $\geq 0.5$ \\
facebook\_large & 22,470 & Medium & 15.20 & Medium & 0.821 & $\geq 0.5$ \\
flickr & 89,250 & Medium & 10.08 & Medium & 0.094 & $0$--$0.5$ \\
full\_cora & 19,793 & Medium & 6.41 & Medium & 0.556 & $\geq 0.5$ \\
genius & 421,961 & Large & 4.37 & Low & -0.053 & $<0$ \\
johnshopkins55 & 5,180 & Small & 72.04 & High & 0.097 & $0$--$0.5$ \\
la & 240,587 & Large & 2.84 & Low & 0.618 & $\geq 0.5$ \\
lastfm\_asia & 7,624 & Small & 7.29 & Medium & 0.856 & $\geq 0.5$ \\
london & 568,795 & Large & 2.66 & Low & 0.626 & $\geq 0.5$ \\
ogbn\_arxiv & 169,343 & Large & 13.67 & Medium & 0.588 & $\geq 0.5$ \\
paris & 114,127 & Large & 3.20 & Low & 0.570 & $\geq 0.5$ \\
penn94 & 41,554 & Medium & 65.56 & High & 0.021 & $0$--$0.5$ \\
pubmed & 19,717 & Medium & 4.50 & Low & 0.686 & $\geq 0.5$ \\
reed98 & 962 & Tiny & 39.11 & High & 0.022 & $0$--$0.5$ \\
shanghai & 183,917 & Large & 2.85 & Low & 0.613 & $\geq 0.5$ \\
squirrel & 2,223 & Small & 42.28 & High & 0.009 & $0$--$0.5$ \\
tag\_bookchild & 76,875 & Medium & 30.24 & High & 0.265 & $0$--$0.5$ \\
tag\_bookhis & 41,551 & Medium & 12.11 & Medium & 0.519 & $\geq 0.5$ \\
tag\_citeseer & 3,186 & Small & 2.65 & Low & 0.729 & $\geq 0.5$ \\
tag\_cora & 2,708 & Small & 3.90 & Low & 0.771 & $\geq 0.5$ \\
tag\_cornell & 191 & Tiny & 2.87 & Low & -0.225 & $<0$ \\
tag\_pubmed & 19,717 & Medium & 4.50 & Low & 0.686 & $\geq 0.5$ \\
tag\_sportsfit & 173,055 & Large & 17.45 & Medium & 0.851 & $\geq 0.5$ \\
tag\_texas & 187 & Tiny & 2.99 & Low & -0.294 & $<0$ \\
tag\_washington & 229 & Tiny & 3.19 & Low & -0.194 & $<0$ \\
tag\_wikics & 11,701 & Medium & 36.85 & High & 0.579 & $\geq 0.5$ \\
tag\_wisconsin & 265 & Tiny & 3.46 & Low & -0.169 & $<0$ \\
texas & 183 & Tiny & 3.05 & Low & -0.298 & $<0$ \\
tolokers-2 & 11,758 & Medium & 88.28 & High & 0.093 & $0$--$0.5$ \\
wiki & 2,405 & Small & 9.64 & Medium & 0.564 & $\geq 0.5$ \\
wiki\_cs & 11,701 & Medium & 36.85 & High & 0.579 & $\geq 0.5$ \\
wisconsin & 251 & Tiny & 3.59 & Low & -0.173 & $<0$ \\
\bottomrule
\end{tabular*}
\end{table*}

\begin{table*}[!p]
\centering
\caption{\textbf{Feature and target values and subgroup assignments.}
Each property has a value column and a group column.
$F$ is the number of features, $s$ is the feature nonzero fraction, $C$ is the number of classes, and $r$ is the class-imbalance ratio.
Groups follow \Cref{tab:property-definitions}; displayed values are rounded.}
\label{tab:dataset-target-statistics}
\footnotesize
\setlength{\tabcolsep}{2.2pt}
\renewcommand{\arraystretch}{1.03}
\begin{tabular*}{\textwidth}{@{\extracolsep{\fill}}lrlrlrlrl@{}}
\toprule
& \multicolumn{2}{c}{\textbf{Number of features}}
& \multicolumn{2}{c}{\textbf{Nonzero fraction}}
& \multicolumn{2}{c}{\textbf{Classes}}
& \multicolumn{2}{c}{\textbf{Class imbalance}} \\
\cmidrule(lr){2-3}\cmidrule(lr){4-5}\cmidrule(lr){6-7}\cmidrule(lr){8-9}
\textbf{Dataset} & $F$ & Group & $s$ & Group & $C$ & Group & $r$ & Group \\
\midrule
actor & 932 & Medium & 0.006 & Sparse & 5 & 3--10 & 2.30 & Medium \\
amazon\_computer & 767 & Medium & 0.343 & Medium & 10 & 3--10 & 17.73 & High \\
amazon\_photo & 745 & Medium & 0.344 & Medium & 8 & 3--10 & 5.86 & Medium \\
amazon\_ratings & 300 & Low & 1.000 & Dense & 5 & 3--10 & 8.49 & Medium \\
amherst41 & 1,193 & Medium & 0.005 & Sparse & 2 & Binary & 1.00 & Low \\
artnet-exp & 75 & Low & 0.712 & Dense & 2 & Binary & 9.00 & Medium \\
blogcatalog & 8,189 & High & 0.009 & Sparse & 6 & 3--10 & 1.19 & Low \\
chameleon & 2,325 & Medium & 0.005 & Sparse & 5 & 3--10 & 1.81 & Low \\
citation\_citeseer & 602 & Medium & 0.008 & Sparse & 6 & 3--10 & 1.49 & Low \\
citeseer & 3,703 & Medium & 0.009 & Sparse & 6 & 3--10 & 2.66 & Medium \\
city-reviews & 204 & Low & 0.110 & Medium & 2 & Binary & 7.27 & Medium \\
coauthor\_cs & 6,805 & High & 0.009 & Sparse & 15 & $>10$ & 35.05 & High \\
coauthor\_physics & 8,415 & High & 0.004 & Sparse & 5 & 3--10 & 6.33 & Medium \\
cora & 1,433 & Medium & 0.013 & Medium & 7 & 3--10 & 4.54 & Medium \\
cora\_ml & 2,879 & Medium & 0.018 & Medium & 7 & 3--10 & 4.44 & Medium \\
cornell & 1,703 & Medium & 0.055 & Medium & 5 & 3--10 & 5.12 & Medium \\
cornell5 & 4,735 & Medium & 0.001 & Sparse & 2 & Binary & 1.07 & Low \\
dblp & 1,639 & Medium & 0.003 & Sparse & 4 & 3--10 & 4.00 & Medium \\
deezer & 128 & Low & 1.000 & Dense & 2 & Binary & 1.26 & Low \\
elliptic\_bitcoin & 165 & Low & 1.000 & Dense & 2 & Binary & 9.25 & Medium \\
facebook\_large & 4,714 & Medium & 0.003 & Sparse & 4 & 3--10 & 2.07 & Medium \\
flickr & 500 & Low & 0.463 & Medium & 7 & 3--10 & 10.84 & High \\
full\_cora & 8,710 & High & 0.006 & Sparse & 70 & $>10$ & 61.87 & High \\
genius & 12 & Low & 0.128 & Medium & 2 & Binary & 4.00 & Medium \\
johnshopkins55 & 2,406 & Medium & 0.002 & Sparse & 2 & Binary & 1.21 & Low \\
la & 37 & Low & 0.321 & Medium & 10 & 3--10 & 1.00 & Low \\
lastfm\_asia & 7,842 & High & 0.050 & Medium & 18 & $>10$ & 98.25 & High \\
london & 37 & Low & 0.317 & Medium & 10 & 3--10 & 1.00 & Low \\
ogbn\_arxiv & 128 & Low & 1.000 & Dense & 40 & $>10$ & 942.10 & High \\
paris & 37 & Low & 0.326 & Medium & 10 & 3--10 & 1.00 & Low \\
penn94 & 4,814 & Medium & 0.001 & Sparse & 2 & Binary & 1.11 & Low \\
pubmed & 500 & Low & 0.101 & Medium & 3 & 3--10 & 1.92 & Low \\
reed98 & 745 & Medium & 0.008 & Sparse & 2 & Binary & 1.40 & Low \\
shanghai & 37 & Low & 0.314 & Medium & 10 & 3--10 & 1.00 & Low \\
squirrel & 2,089 & Medium & 0.007 & Sparse & 5 & 3--10 & 3.24 & Medium \\
tag\_bookchild & 3,072 & Medium & 1.000 & Dense & 24 & $>10$ & 304.68 & High \\
tag\_bookhis & 3,072 & Medium & 1.000 & Dense & 12 & $>10$ & 5842.00 & High \\
tag\_citeseer & 3,072 & Medium & 1.000 & Dense & 6 & 3--10 & 3.36 & Medium \\
tag\_cora & 3,072 & Medium & 1.000 & Dense & 7 & 3--10 & 4.54 & Medium \\
tag\_cornell & 3,072 & Medium & 1.000 & Dense & 5 & 3--10 & 4.88 & Medium \\
tag\_pubmed & 3,072 & Medium & 1.000 & Dense & 3 & 3--10 & 1.92 & Low \\
tag\_sportsfit & 3,072 & Medium & 1.000 & Dense & 13 & $>10$ & 137.91 & High \\
tag\_texas & 3,072 & Medium & 1.000 & Dense & 4 & 3--10 & 5.72 & Medium \\
tag\_washington & 3,072 & Medium & 1.000 & Dense & 5 & 3--10 & 11.78 & High \\
tag\_wikics & 3,072 & Medium & 1.000 & Dense & 10 & 3--10 & 9.08 & Medium \\
tag\_wisconsin & 3,072 & Medium & 1.000 & Dense & 5 & 3--10 & 12.20 & High \\
texas & 1,703 & Medium & 0.049 & Medium & 4 & 3--10 & 5.61 & Medium \\
tolokers-2 & 19 & Low & 0.668 & Dense & 2 & Binary & 3.58 & Medium \\
wiki & 4,973 & Medium & 0.130 & Medium & 17 & $>10$ & 45.11 & High \\
wiki\_cs & 300 & Low & 1.000 & Dense & 10 & 3--10 & 9.08 & Medium \\
wisconsin & 1,703 & Medium & 0.056 & Medium & 5 & 3--10 & 11.80 & High \\
\bottomrule
\end{tabular*}
\end{table*}

\begin{table*}[!p]
\centering
\caption{\textbf{Additional dataset statistics.}
These quantities describe dataset coverage but do not define the RQ2 subgroups.
$E$ counts undirected non-self-loop edges, $q_{50}$ and $q_{99}$ are nonzero-degree quantiles, Hub is $q_{99}/q_{50}$, and $N_L$ counts labeled nodes.
Feature AUROC and label entropy follow \Cref{tab:additional-property-definitions}.}
\label{tab:dataset-additional-statistics}
\footnotesize
\setlength{\tabcolsep}{2.2pt}
\renewcommand{\arraystretch}{1.03}
\begin{tabular*}{\textwidth}{@{\extracolsep{\fill}}lrrrrrrrr@{}}
\toprule
\textbf{Dataset} & $E$ & $q_{50}$ & $q_{99}$ & Hub & Density & \shortstack{Feature\\AUROC} & $N_L$ & \shortstack{Label\\entropy} \\
\midrule
actor & 26,659 & 4.00 & 51.00 & 12.75 & 9.23e-04 & 0.503 & 7,600 & 0.977 \\
amazon\_computer & 245,861 & 23.00 & 245.00 & 10.65 & 2.60e-03 & 0.542 & 13,752 & 0.813 \\
amazon\_photo & 119,081 & 22.00 & 201.32 & 9.15 & 4.07e-03 & 0.566 & 7,650 & 0.926 \\
amazon\_ratings & 93,050 & 5.00 & 31.00 & 6.20 & 3.10e-04 & 0.502 & 24,492 & 0.877 \\
amherst41 & 90,954 & 70.00 & 297.98 & 4.26 & 3.64e-02 & 0.500 & 2,032 & 1.000 \\
artnet-exp & 280,348 & 2.00 & 135.00 & 67.50 & 2.21e-04 & 0.521 & 50,405 & 0.469 \\
blogcatalog & 171,743 & 48.00 & 265.00 & 5.52 & 1.27e-02 & 0.652 & 5,196 & 0.999 \\
chameleon & 8,854 & 9.00 & 104.11 & 11.57 & 2.24e-02 & 0.516 & 890 & 0.984 \\
citation\_citeseer & 5,337 & 1.00 & 18.71 & 18.71 & 5.97e-04 & 0.590 & 4,230 & 0.995 \\
citeseer & 4,552 & 2.00 & 15.00 & 7.50 & 8.23e-04 & 0.606 & 3,327 & 0.979 \\
city-reviews & 1,165,415 & 4.00 & 196.00 & 49.00 & 1.05e-04 & 0.536 & 138,879 & 0.532 \\
coauthor\_cs & 81,894 & 6.00 & 47.00 & 7.83 & 4.87e-04 & 0.843 & 18,333 & 0.889 \\
coauthor\_physics & 247,962 & 10.00 & 77.00 & 7.70 & 4.17e-04 & 0.762 & 34,493 & 0.845 \\
cora & 5,278 & 3.00 & 19.00 & 6.33 & 1.44e-03 & 0.612 & 2,708 & 0.941 \\
cora\_ml & 8,158 & 3.00 & 32.00 & 10.67 & 1.82e-03 & 0.616 & 2,995 & 0.952 \\
cornell & 277 & 2.00 & 10.18 & 5.09 & 1.66e-02 & 0.630 & 183 & 0.880 \\
cornell5 & 790,777 & 64.00 & 343.41 & 5.37 & 4.54e-03 & 0.505 & 16,822 & 0.999 \\
dblp & 52,867 & 3.00 & 42.00 & 14.00 & 3.37e-04 & 0.532 & 17,716 & 0.885 \\
deezer & 92,752 & 4.00 & 38.00 & 9.50 & 2.32e-04 & 0.506 & 28,281 & 0.991 \\
elliptic\_bitcoin & 234,355 & 2.00 & 13.00 & 6.50 & 1.13e-05 & 0.617 & 46,564 & 0.461 \\
facebook\_large & 170,823 & 7.00 & 114.31 & 16.33 & 6.77e-04 & 0.548 & 22,470 & 0.976 \\
flickr & 449,878 & 6.00 & 67.00 & 11.17 & 1.13e-04 & 0.511 & 89,250 & 0.812 \\
full\_cora & 63,421 & 4.00 & 39.00 & 9.75 & 3.24e-04 & 0.662 & 19,793 & 0.943 \\
genius & 922,868 & 1.00 & 50.00 & 50.00 & 1.04e-05 & 0.537 & 421,961 & 0.722 \\
johnshopkins55 & 186,586 & 54.00 & 298.21 & 5.52 & 1.39e-02 & 0.504 & 4,762 & 0.993 \\
la & 341,523 & 3.00 & 4.00 & 1.33 & 1.18e-05 & 0.525 & 240,587 & 1.000 \\
lastfm\_asia & 27,806 & 4.00 & 55.77 & 13.94 & 9.57e-04 & 0.606 & 7,624 & 0.821 \\
london & 756,502 & 3.00 & 4.00 & 1.33 & 4.68e-06 & 0.539 & 568,795 & 1.000 \\
ogbn\_arxiv & 1,157,799 & 6.00 & 106.00 & 17.67 & 8.07e-05 & 0.628 & 169,343 & 0.811 \\
paris & 182,511 & 3.00 & 4.00 & 1.33 & 2.80e-05 & 0.507 & 114,127 & 1.000 \\
penn94 & 1,362,229 & 48.00 & 280.47 & 5.84 & 1.58e-03 & 0.502 & 38,815 & 0.998 \\
pubmed & 44,324 & 2.00 & 35.00 & 17.50 & 2.28e-04 & 0.639 & 19,717 & 0.965 \\
reed98 & 18,812 & 29.00 & 155.78 & 5.37 & 4.07e-02 & 0.506 & 865 & 0.980 \\
shanghai & 262,092 & 3.00 & 4.00 & 1.33 & 1.55e-05 & 0.572 & 183,917 & 1.000 \\
squirrel & 46,998 & 10.00 & 301.56 & 30.16 & 1.90e-02 & 0.513 & 2,223 & 0.950 \\
tag\_bookchild & 1,162,522 & 11.00 & 236.00 & 21.45 & 3.93e-04 & 0.701 & 76,875 & 0.749 \\
tag\_bookhis & 251,590 & 3.00 & 120.00 & 40.00 & 2.91e-04 & 0.662 & 41,551 & 0.550 \\
tag\_citeseer & 4,225 & 2.00 & 15.00 & 7.50 & 8.33e-04 & 0.708 & 3,186 & 0.971 \\
tag\_cora & 5,278 & 3.00 & 19.00 & 6.33 & 1.44e-03 & 0.737 & 2,708 & 0.941 \\
tag\_cornell & 274 & 2.00 & 10.00 & 5.00 & 1.51e-02 & 0.736 & 191 & 0.886 \\
tag\_pubmed & 44,324 & 2.00 & 35.00 & 17.50 & 2.28e-04 & 0.690 & 19,717 & 0.965 \\
tag\_sportsfit & 1,510,067 & 6.00 & 169.46 & 28.24 & 1.01e-04 & 0.792 & 173,055 & 0.775 \\
tag\_texas & 280 & 2.00 & 13.64 & 6.82 & 1.61e-02 & 0.779 & 186 & 0.839 \\
tag\_washington & 365 & 2.00 & 10.70 & 5.35 & 1.40e-02 & 0.683 & 229 & 0.816 \\
tag\_wikics & 215,603 & 13.00 & 259.00 & 19.92 & 3.15e-03 & 0.751 & 11,701 & 0.905 \\
tag\_wisconsin & 459 & 2.00 & 13.78 & 6.89 & 1.31e-02 & 0.721 & 265 & 0.816 \\
texas & 279 & 2.00 & 13.72 & 6.86 & 1.68e-02 & 0.646 & 182 & 0.838 \\
tolokers-2 & 519,000 & 30.00 & 826.86 & 27.56 & 7.51e-03 & 0.513 & 11,758 & 0.757 \\
wiki & 11,596 & 6.00 & 61.38 & 10.23 & 4.01e-03 & 0.805 & 2,405 & 0.879 \\
wiki\_cs & 215,603 & 13.00 & 259.00 & 19.92 & 3.15e-03 & 0.650 & 11,701 & 0.905 \\
wisconsin & 450 & 2.00 & 14.00 & 7.00 & 1.43e-02 & 0.705 & 251 & 0.814 \\
\bottomrule
\end{tabular*}
\end{table*}

\FloatBarrier

\subsection{Benchmark Comparison Criteria and Evidence}
\label{app:benchmark-comparison-evidence}

This subsection explains the entries in
Table~\ref{tab:benchmark-comparison}.
We first define the columns from left to right, then discuss the
benchmarks in the order shown in the table.
Assessments use the original papers and official documentation;
for raw results, we also inspect publicly available files in the
official releases.
A circle indicates full support, a triangle partial support,
and a cross no support established by these sources.

\mypar{Column definitions, from left to right.}
\begin{enumerate}
    \item \textbf{Number of datasets.}
    We count distinct node-classification tasks, excluding tasks
    such as graph classification and link prediction.

    \item \textbf{HPO control.}
    A circle requires specified, comparable hyperparameter-search
    conditions across the compared methods.
    A triangle indicates that tuning is performed, but the search
    conditions differ across methods or are not fully specified.
    A cross indicates that no controlled search is documented.

    \item \textbf{Validation selection.}
    A circle requires explicit validation-only selection at both
    stages: choosing the checkpoint within a training run and
    choosing the final hyperparameter configuration.
    A triangle indicates that only one stage, or an incomplete
    selection procedure, is documented.
    A cross indicates that validation-only selection is not established.

    \item \textbf{Data splits.}
    We report the number of train/validation/test partitions used
    in the evaluation.
    A set, such as $\{1,10\}$, means that the number varies by dataset.

    \item \textbf{Multi-metric evaluation.}
    Full support requires reporting more than one predictive metric
    for each evaluated node-classification task.
    Reporting accuracy on one dataset and AUROC on another does not
    qualify: each task is still evaluated using a single metric.
    Metrics available only in implementation code, without a reported
    multi-metric evaluation, do not establish this support.

    \item \textbf{Compute cost.}
    The three symbols refer, in order, to training, hyperparameter
    tuning, and inference.
    For each phase, a circle indicates direct measurements;
    a triangle indicates incomplete measurements, such as an
    aggregate time, a proxy, or a lower bound; and a cross indicates
    no reported measurement.
    A circle does not necessarily mean that costs are reported for
    every method--dataset pair.

    \item \textbf{Raw results.}
    A circle requires released run-level records covering the full
    comparison.
    Downloading existing records through a command or an external
    repository counts in the same way as downloading a file directly.
    A triangle indicates that records cover only part of the comparison.
    A cross indicates that no original run-level records are released.
    Code that generates records only after rerunning training or tuning
    does not count.
    Aggregate means and standard deviations also do not count,
    because they do not expose the individual runs.

    \item \textbf{Leaderboard.}
    Full support requires a public leaderboard that accepts external
    submissions.
    A static table of published results is not sufficient.
\end{enumerate}

\mypar{Evidence for each table row.}
The descriptions below follow the same column order:
dataset count; tuning and selection; splits and metrics;
computational cost; and released results and leaderboard.

\noindent\textbf{\citet{yang2016revisiting} (Planetoid).}
The evaluation covers four node-classification tasks.
It does not specify comparable hyperparameter-search conditions
across model families, but documents validation-based model selection.
Results use one standard partition and a single predictive metric
per task.
Training, tuning, and inference costs are not reported.
The \href{https://github.com/kimiyoung/planetoid}{official release}
provides datasets and runnable code, but no original run-level
evaluation records.
No public submission leaderboard is provided.

\noindent\textbf{\citet{shchur2018pitfalls} (Pitfalls).}
The evaluation covers eight node-classification datasets.
A common random-search framework provides HPO control.
Validation selection is partial because validation-only selection
is not established at both the checkpoint and final-configuration
stages for every compared result.
The evaluation uses 100 partitions and a single predictive metric
per task; none of the three cost phases is reported.
The \href{https://github.com/shchur/gnn-benchmark}{official framework}
can export individual-run records from its experiment database,
but the original database is not released.
Obtaining these records therefore requires rerunning experiments,
so raw results receive a cross.
There is no public submission leaderboard.

\noindent\textbf{\citet{pei2020geom} (Geom-GCN).}
The evaluation covers nine node-classification datasets,
with controlled hyperparameter tuning and validation-based selection.
Results use ten fixed partitions and a single predictive metric
per task.
Training time is measured, but tuning and inference times are
not reported separately.
The \href{https://github.com/bingzhewei/geom-gcn}{official repository}
provides result CSVs containing means and standard deviations.
Individual-run files are written to a local \texttt{runs} directory
only after retraining.
These aggregate files and reproduction commands do not constitute
released run-level records.
There is no public submission leaderboard.

\noindent\textbf{\citet{platonov2026fair} (Fair GFM Evaluation).}
The comparison covers four node-classification tasks,
with controlled tuning and validation-based selection.
It uses one reported partition and does not report multiple
predictive metrics per task.
Training, tuning, and inference costs are all measured.
The \href{https://github.com/yandex-research/gnn-fair-evaluation}{official release}
includes run-level GNN experiment reports, but directs readers to
separate model repositories for the GFM experiments.
The released records therefore cover only part of the comparison,
which gives raw results a triangle.
No public submission leaderboard is provided.

\noindent\textbf{\citet{you2020design} (GraphGym).}
The node-classification comparison covers 18 tasks.
Controlled design-space searches provide HPO control, but
validation-only selection at both stages is only partially
established across the reported analyses.
The evaluation reports three repetitions and a single predictive
metric per task.
Training runtime is directly measured, while tuning and inference
costs are reported incompletely.
The \href{https://github.com/snap-stanford/GraphGym}{official release}
describes its design-space outputs as raw results, but the distributed
CSVs aggregate results across random seeds.
Individual seed-level files require rerunning the experiments,
so raw results receive a cross.
There is no public submission leaderboard.

\noindent\textbf{\citet{hu2020open} (OGB).}
The comparison includes five node-classification tasks.
OGB provides a common evaluation framework and validation-based
model selection, but does not require comparable HPO budgets
across submitted methods.
Each task has one official split and a designated predictive metric,
rather than a multi-metric evaluation of the same task.
The three cost phases are not reported.
The \href{https://ogb.stanford.edu/}{public leaderboard}
accepts external submissions, but its submitted scores do not
provide the underlying individual-run records.
Thus, OGB receives a circle for the leaderboard and a cross
for raw results.

\noindent\textbf{\citet{lim2021large} (LINKX).}
The evaluation introduces seven node-classification tasks.
HPO control is partial because search conditions are
method-dependent; validation-based selection is documented.
Results use five partitions and a single predictive metric
per task.
Training-runtime information is incomplete, and separate tuning
and inference costs are not reported.
The \href{https://github.com/CUAI/Non-Homophily-Large-Scale}{official code}
writes aggregate scores to its result CSV, rather than releasing
the individual-run records underlying the comparison.
Raw results therefore receive a cross.
There is no public submission leaderboard.

\noindent\textbf{\citet{dwivedi2023benchmarking} (Benchmarking GNNs).}
The node-classification subset contains three tasks.
HPO control and validation selection are partial because the
search conditions and selection procedures are not fully
established under our criteria across all evaluated settings.
The evaluation uses one or 20 partitions depending on the dataset,
without a reported multi-metric comparison for each task.
Training runtime is directly reported, but separate tuning and
inference costs are not.
The \href{https://github.com/graphdeeplearning/benchmarking-gnns/blob/master/docs/03_run_codes.md}{official reproduction guide}
describes result files produced by running the experimental scripts;
it does not provide downloadable records from the original runs.
Raw results therefore receive a cross.
No public submission leaderboard is documented.

\noindent\textbf{\citet{platonov2023critical} (Heterophily).}
The evaluation covers five node-classification datasets.
Tuning conditions are only partially controlled across methods,
while validation-based selection is documented.
Results use ten partitions and a single predictive metric per task.
Training, tuning, and inference costs are not reported.
The \href{https://github.com/yandex-research/heterophilous-graphs/blob/main/utils.py}{official logger}
can save individual validation and test results to
\texttt{metrics.yaml}, but records from the original experiments
are not released.
Generating the file requires rerunning experiments, so raw results
receive a cross.
There is no public submission leaderboard.

\noindent\textbf{\citet{luo2024classic} (Classic GNNs).}
The comparison covers 18 node-classification datasets,
with controlled tuning and validation-based selection.
It uses one or ten partitions depending on the dataset and
does not report multiple predictive metrics per task.
None of the three cost phases is reported.
The \href{https://github.com/LUOyk1999/tunedGNN/blob/main/medium_graph/logger.py}{official logger}
prints individual-run scores during execution, but the saved
result CSV contains only their mean and standard deviation.
The original individual-run records are not released,
so raw results receive a cross.
There is no public submission leaderboard.

\noindent\textbf{\citet{bazhenov2026graphland} (GraphLand).}
The benchmark contains seven node-classification tasks.
Method-specific search conditions give partial HPO control,
while validation-based selection is documented.
Results use two or three partitions and a single predictive
metric per task.
Training-cost reporting is incomplete, and separate tuning and
inference costs are not reported.
The \href{https://github.com/yandex-research/graphland}{official workflow}
saves individual results locally when experiments are run,
but does not release the original run-level records.
Raw results therefore receive a cross.
There is no public submission leaderboard.

\noindent\textbf{\citet{yu2026evaluating} (GFM Benchmark).}
The comparison covers 15 node-classification tasks.
HPO control is partial, and the published protocol does not
establish validation-only checkpoint and final-configuration
selection.
Results use 50 reported partitions and multiple predictive metrics.
Training and inference costs are reported incompletely,
without a separate tuning-cost account.
The \href{https://github.com/smufang/GFMBenchmark}{official scripts}
generate experiment logs after pretraining and downstream evaluation
are rerun.
The automated download supplies datasets, splits, and checkpoints,
not the original result logs.
Raw results therefore receive a cross.
There is no public submission leaderboard.

\noindent\textbf{\method.}
Our comparison covers 51 node-classification tasks.
It uses controlled tuning and selects both checkpoints and final
configurations using validation data only.
Each label regime uses five shared partitions, with multiple
predictive metrics reported for each task.
Training, tuning, and inference costs are directly measured.
We release run-level records for the comparison and provide a public
leaderboard accepting external submissions.
The implementation and reporting procedures are described in
Appendices~\ref{app:datasets}--\ref{app:overhead-pareto}.
\section{Methods, Artifacts, and Optimization}
\label{app:methods}

This appendix specifies the methods and configurations used in the benchmark. It defines the method scope, identifies each implementation and pretrained artifact, describes target-dataset adaptation and preprocessing, documents the pretraining-overlap policy, and reports the optimization spaces and observed selection outcomes.

\subsection{Method Scope and GFM Definition}
\label{app:method-scope}

The benchmark contains 21 methods comprising one feature-only MLP, ten
GNNs, four graph Transformers, and six GFMs. We include methods that support
homogeneous single-graph node classification with the released node features,
provide a runnable implementation and sufficiently specified configuration or
pretrained artifact, use the shared graph, labels, masks, and evaluation
interface, and add a distinct design family. Unsupported or failed target
cells remain explicit. Checkpoint variants count as one method unless they
change the target-evaluation procedure.

Our evaluated GFMs use a shared pretrained model that can be applied to unseen graphs with different sizes, feature dimensions, and label spaces. We restrict the evaluation to target-dataset adaptation that keeps the pretrained parameters fixed.
We exclude methods that update the pretrained backbone on each target dataset,
address a different graph task, or lack a runnable and sufficiently specified
release. The registered Cora, Wisconsin, Product, and arXiv GraphAny
checkpoints therefore count as one method, for which we evaluate the arXiv
artifact.

\subsection{Implementations and Target-Dataset Adaptation}
\label{app:method-implementations}

Table~\ref{tab:method-implementations} identifies the implementation
provenance and evaluated configuration set for every method.
All 15 supervised methods run through benchmark-owned code rather than
the original authors' training scripts. Seven implementations wrap the named
PyG operators. The other eight are written directly in \method, including
the feature-only MLP and reimplementations of the published architectures.
The GFM rows separately identify official implementations and benchmark
adapters that load official pretrained artifacts.

\begin{table*}[!tbp]
\centering
\caption{\textbf{Implementation provenance and evaluated configuration sets.}
Benchmark implementation and benchmark reimplementation denote code maintained
within \method rather than execution of the original authors' repositories.
Benchmark wrapper denotes \method code built from the named PyG operator,
whereas the GFM rows explicitly distinguish official implementations from
adapters that load official checkpoints.}
\scriptsize
\setlength{\tabcolsep}{3.2pt}
\renewcommand{\arraystretch}{1.04}
\begin{tabularx}{\textwidth}{@{}ll>{\raggedright\arraybackslash}Xl@{}}
\toprule
Method & Family & Implementation used & Configuration set \\
\midrule
MLP & Feature-only & Benchmark implementation (PyTorch) & default + 200 sampled \\
APPNP & GNN & Benchmark wrapper using PyG \texttt{APPNP} & default + 200 sampled \\
FAGCN & GNN & Benchmark wrapper using PyG \texttt{FAConv} & default + 200 sampled \\
GAT & GNN & Benchmark wrapper using PyG \texttt{GATConv} & default + 200 sampled \\
GATv2 & GNN & Benchmark wrapper using PyG \texttt{GATv2Conv} & default + 200 sampled \\
GCN & GNN & Benchmark wrapper using PyG \texttt{GCNConv} & default + 200 sampled \\
GCNII & GNN & Benchmark wrapper using PyG \texttt{GCN2Conv} & default + 200 sampled \\
GPRGNN & GNN & Benchmark reimplementation & default + 200 sampled \\
GraphSAGE & GNN & Benchmark wrapper using PyG \texttt{SAGEConv} & default + 200 sampled \\
LINKX & GNN & Benchmark reimplementation & default + 200 sampled \\
SGC & GNN & Benchmark reimplementation & default + 200 sampled \\
NAGphormer & GT & Benchmark reimplementation of Hop2Token/PE & default + 200 sampled \\
NodeFormer & GT & Benchmark reimplementation of all-pair attention & default + 200 sampled \\
Polynormer & GT & Benchmark reimplementation of local-to-global path & default + 200 sampled \\
SGFormer & GT & Benchmark reimplementation of hybrid architecture & default + 200 sampled \\
G2T-FM & GFM & Official G2T-FM implementation & canonical \\
GraphAny & GFM & Benchmark adapter to official checkpoint & canonical arXiv artifact \\
GraphPFN & GFM & Official GraphPFN implementation & PCA off / 64 \\
GVT & GFM & Benchmark adapter to official RGVT checkpoint & encoder depths 1--8 \\
Node4All & GFM & Benchmark adapter to official CGT checkpoint & canonical MLP head \\
NodePFN & GFM & Official v1.0.0 implementation and checkpoint & default + 200 sampled \\
\bottomrule
\end{tabularx}
\label{tab:method-implementations}
\end{table*}

For every method, model and configuration selection use validation data, and test labels are reserved for final reporting. Table~\ref{tab:app-gfm-implementation-audit} describes whether each GFM uses labels as prediction context, fits analytical predictors, or trains a target-specific head.

\subsection{Pretrained Artifacts, Preprocessing, and Overlap}
\label{app:checkpoint-audit}

Table~\ref{tab:app-gfm-implementation-audit} records the released artifact used
for each GFM, which components remain frozen, and the target-dataset operation.

\begin{table*}[!t]
\centering
\caption{\textbf{Pretrained artifacts and target-dataset adaptation.}
The table identifies the released artifact actually evaluated and states which
components, if any, are trained on each target dataset.}
\small
\setlength{\tabcolsep}{3.5pt}
\renewcommand{\arraystretch}{1.05}
\begin{tabular}{@{}l>{\raggedright\arraybackslash}p{0.29\textwidth}>{\raggedright\arraybackslash}p{0.27\textwidth}>{\raggedright\arraybackslash}p{0.29\textwidth}@{}}
\toprule
Method & Artifact and source & Frozen / trainable & Target operation \\
\midrule
G2T-FM &
Official G2T-FM implementation with the TabPFNv2 classifier checkpoint &
TabPFNv2 frozen; no trainable head &
Constructs graph-derived tabular features and supplies training labels as
context. Uses \texttt{n\_epochs=0} and no fine-tuning. \\

GraphAny &
Official \texttt{graph\_any\_arxiv.pt} checkpoint &
Fusion MLP frozen; no trainable head &
Forms feature and propagated channels, analytically solves label-conditioned
linear channels from the training labels, and fuses their logits. \\

GraphPFN &
Official GraphPFN 1.3 release &
Complete pretrained model frozen; no trainable head &
Runs graph-aware label-conditioned inference and selects between PCA disabled
and 64 dimensions. \\

GVT &
Official RGVT \texttt{seed\_42.pth} checkpoint &
RGVT encoder frozen; target MLP head trained &
Computes representations at depths 1--8 and trains a separate
validation-selected head at each depth. \\

Node4All &
Official \texttt{cgt\_enc.pth} checkpoint &
CGT encoder frozen; target MLP head trained &
Computes one canonical representation and trains a validation-selected
target-specific MLP head. \\

NodePFN &
Official v1.0.0 \texttt{checkpoint\_}\allowbreak\texttt{epoch\_30.ckpt} &
Complete pretrained model frozen; no trainable head &
Uses training labels as context and selects dimensionality reduction,
component count, smoothing, and ensemble settings. \\
\bottomrule
\end{tabular}
\label{tab:app-gfm-implementation-audit}
\end{table*}

All methods receive labels remapped to contiguous nonnegative integers and
source features cast to \texttt{float32}. Invalid labels are removed from all
masks. Edges are coalesced, converted to a bidirectional view, and assigned
exactly one self-loop per node before method-specific transforms. MLP ignores
the edges. FAGCN and LINKX remove self-loops internally. NAGphormer constructs
Hop2Token neighborhoods and positional encodings, NodeFormer constructs
adjacency powers for relational bias, and Polynormer applies its staged
local-to-global path. GFM-specific reduction, structural descriptors, label
conditioning, and head training are summarized in
Table~\ref{tab:app-gfm-implementation-audit}.

We audit pretraining overlap by canonical graph identity rather than loader
name. A shared node--edge topology counts as overlap even when two loaders
provide different feature matrices. The evaluated GraphAny and GVT
checkpoints were pretrained on OGBN-Arxiv
\citep{zhao2025fully,lee2025view}. Their native results on
\texttt{ogbn\_arxiv} are therefore omitted, and common-universe analyses
replace only those two cells with the matched GCN-Default result. The other
four GFMs use synthetic pretraining tasks or graphs. G2T-FM uses synthetic
tabular tasks, GraphPFN and Node4All use synthetic attributed graphs, and
NodePFN uses synthetic Erd\H{o}s--R\'enyi and contextual block-model graphs
\citep{eremeev2025turning,eremeev2025graphpfn,lee2026node4all,choi2026learning}.
None has an exact real-target overlap among the 51 datasets. The
\texttt{cora}/\texttt{tag\_cora}, \texttt{pubmed}/\texttt{tag\_pubmed},
and \texttt{wiki\_cs}/\texttt{tag\_wikics} pairs share topology-level
identity, but none of the included GFMs was pretrained on these graph families.
Evaluation-only exposure in an original publication is not counted as
pretraining.

\subsection{Optimization Protocol, Search Spaces, and Outcomes}
\label{app:search-spaces}

For each dataset, label split, and experimental repeat, we select
each method's configuration independently.
This section explains how configurations are trained and selected,
which settings are searched, and how the resulting search outcomes
are summarized.

\mypar{Checkpoint and configuration selection.}
Both the checkpoint within each trial and the final configuration
across trials are selected using validation loss:
cross-entropy for multiclass tasks and binary cross-entropy for
binary tasks.
Ties are broken by training loss and then evaluation order.
Test performance is never used for either decision.

\mypar{Supervised training and tuning.}
Each supervised method is first evaluated with its default
configuration, recorded as trial~1.
Table~\ref{tab:search-spaces} lists the default hyperparameters
and validation-selection objectives for all evaluated methods.
We then sample up to 200 additional unique configurations from
each supervised method's predefined search space, giving at most
201 trials in total.
Search dimensions are sampled independently.
Duplicate configurations are detected using a sorted, serialized
configuration signature and resampled.
Exceeding the duplicate-resampling limit raises an error.

Training uses Adam, except for NAGphormer, which uses AdamW.
Each trial runs for at most 2,500 epochs unless the method specifies
a smaller limit.
The default early-stopping patience is 100; the search considers
patience values of $\{20,100\}$.

An additional trial that fails or produces non-finite results is
skipped and excluded from selection.
It is not replaced by another trial, so the budget allows up to
201 attempted configurations rather than guaranteeing 201
successful ones.
The default is executed before the error-handling procedure for
additional trials; a default failure can therefore stop the
evaluation for that method, dataset, split, and repeat.

\mypar{GFM adaptation, search, and inference settings.}
GFM searches vary only settings exposed by their target-evaluation
interfaces; they never update the pretrained backbone.
GVT and Node4All train prediction heads, while the other methods
use their respective inference or analytical adaptation procedures.
Early stopping applies to gradient-based training, not to frozen
in-context inference.
The settings below are implemented by the benchmark adapters.

\begin{itemize}
    \item \textbf{G2T-FM.}
    We use one canonical configuration without gradient updates.
    All training nodes are supplied as in-context labels.
    The inference batch size is~1,
    \texttt{seq\_len\_pred} is 1024, and the PEARL batch size is~8.

    \item \textbf{GraphAny.}
    We use one canonical configuration.
    The method processes nodes in batches of 100,000 and adapts
    analytically using the training labels.

    \item \textbf{GraphPFN.}
    We compare two PCA settings.
    Each evaluation uses a full-graph call with all training nodes
    supplied as labeled context.
    The requested ensemble size is 10, which the model increases
    automatically when required.

    \item \textbf{GVT.}
    We compare encoder depths from 1 to~8.
    For each depth, the frozen encoder processes the full graph,
    and a separate two-layer prediction head is trained.
    The head has hidden width 128, a ReLU activation, and no dropout.
    It uses Adam with learning rate $0.005$, zero weight decay,
    a maximum of 2,500 epochs, and patience 200.
    The checkpoint with the lowest validation CE or BCE is restored,
    with training loss breaking ties.
    These head settings are shared across all 510 stored GVT results.

    \item \textbf{Node4All.}
    We use one canonical configuration with a frozen encoder and
    a two-layer prediction head of hidden width 512.
    The method receives the full graph and processes features in
    chunks as needed, using a budget of 1,000,000 feature elements.
    The head uses Adam with learning rate $0.001$, zero weight decay,
    a maximum of 2,500 epochs, and patience 200.
    Only training labels are used to fit the head;
    validation loss selects its checkpoint.

    \item \textbf{NodePFN.}
    We evaluate the default and up to 200 additional preprocessing
    and inference configurations.
    The search considers ensemble sizes of $\{1,4,8,16,32\}$,
    either no dimensionality reduction or truncated SVD with
    $\{10,15,20,25,50\}$ components, and the exposed smoothing choices.
    All training nodes are used as context.
    All nontraining nodes are treated as queries and evaluated
    in batches of~32.
\end{itemize}

\begin{table*}[!ht]
\centering
\caption{\textbf{Default configurations and source-defined search spaces.}
The shared training row applies where a method row does not give
a different training setting. NodeFormer and Polynormer learning rates, weight
decays, and dropout values are categorical choices, not continuous or
log-uniform ranges.}
\label{tab:search-spaces}
\scriptsize
\setlength{\tabcolsep}{3pt}
\renewcommand{\arraystretch}{1.04}

\begin{tabularx}{\textwidth}{
    @{}
    >{\raggedright\arraybackslash}p{0.15\textwidth}
    >{\raggedright\arraybackslash}p{0.31\textwidth}
    >{\raggedright\arraybackslash}X
    @{}
}
\toprule
Method(s) & Default & Varied values \\
\midrule

Shared supervised training &
Adam; learning rate $10^{-2}$, weight decay $5\!\times\!10^{-4}$, 2,500 epochs,
patience 100 &
learning rate $\{10^{-3},5\!\times\!10^{-3},10^{-2}\}$; weight decay
$\{0,5\!\times\!10^{-5},5\!\times\!10^{-4}\}$; patience $\{20,100\}$ \\

APPNP &
hidden 64; feature dropout .5; $K=10$; $\alpha=.1$; propagation dropout 0 &
hidden $\{64,128,256,512\}$; feature dropout $\{0,.3,.5,.7\}$;
$K\in\{5,10,20\}$; $\alpha\in\{.1,.2,.5\}$; propagation dropout
$\{0,.3,.5\}$; shared training \\

FAGCN &
hidden 32; feature dropout .5; 2 layers; $\epsilon=.3$; edge dropout .5 &
hidden $\{32,64,128\}$; feature and edge dropout $\{0,.3,.5,.7\}$;
layers $\{1,2,3,4\}$; $\epsilon\in\{.1,.2,.3,.5\}$; shared training \\

GAT, GATv2 &
hidden 64; one head; feature dropout .5; attention dropout .2; 2 layers;
no normalization, residual, or pre-transform &
hidden $\{64,128,256,512\}$; heads $\{1,4\}$; feature dropout
$\{0,.3,.5,.7\}$; attention dropout $\{0,.2\}$; layers $\{1,2,4,8\}$;
normalization $\{\texttt{none},\texttt{layernorm}\}$; residual and
pre-transform booleans; shared training \\

GCN, GraphSAGE &
hidden 64; feature dropout .5; 2 layers; no normalization, residual, or
pre-transform &
hidden $\{64,128,256,512\}$; feature dropout $\{0,.3,.5,.7\}$;
layers $\{1,2,4,8\}$; normalization $\{\texttt{none},\texttt{layernorm}\}$;
residual and pre-transform booleans; shared training \\

GCNII &
hidden 64; feature dropout .6; 16 layers; $\alpha=.1$; $\theta=.5$;
shared weights; no normalization, residual, or pre-transform &
hidden $\{64,128,256,512\}$; feature dropout $\{0,.3,.5,.7\}$;
layers $\{4,8,16,32\}$; $\alpha\in\{.1,.2,.5\}$;
$\theta\in\{.5,1,1.5\}$; normalization, residual, and pre-transform choices;
shared training \\

GPRGNN &
hidden 64; feature dropout .5; propagation dropout 0; $K=10$;
$\alpha=.1$; PPR initialization &
hidden $\{64,128,256,512\}$; both dropouts $\{0,.3,.5,.7\}$;
$K\in\{5,10,15,20\}$; $\alpha\in\{.1,.2,.5,.9\}$;
initialization $\{\texttt{ppr},\texttt{sgc},\texttt{random}\}$; shared training \\

LINKX &
hidden 32; feature dropout .5; 2 total layers; one adjacency and one feature
layer &
hidden $\{32,64,128\}$; feature dropout $\{0,.3,.5,.7\}$; total layers
$\{1,2,3\}$; adjacency and feature layers $\{1,2\}$; shared training \\

MLP &
hidden 64; feature dropout .5; 2 layers; no normalization or identity skip &
hidden $\{64,128,256,512\}$; feature dropout $\{0,.3,.5,.7\}$;
layers $\{1,2,3,4\}$; normalization
$\{\texttt{none},\texttt{layernorm}\}$; identity-skip boolean; shared training \\

NAGphormer &
AdamW; hidden 512; 7 hops; 1 layer; 8 heads; dropout .1; PE dimension 15;
learning rate $10^{-3}$; weight decay $10^{-5}$ &
hidden $\{128,256,512\}$; hops $\{2,\ldots,20\}$; layers $\{1,\ldots,5\}$;
dropout $\{.1,.3,.5\}$; learning rate
$\{10^{-4},10^{-3},5\!\times\!10^{-3}\}$; weight decay
$\{10^{-5},10^{-4},5\!\times\!10^{-4}\}$ \\

NodeFormer &
hidden 32; 2 layers; 4 heads; dropout 0; 30 random features; 10 Gumbel
samples; link weight 1; bias order 2 and sigmoid transform; learning rate $10^{-3}$; weight decay $5\!\times\!10^{-3}$ &
hidden $\{32,64,128\}$; layers $\{2,3\}$; heads $\{1,2,4\}$; dropout
$\{0,.3,.5\}$; random features $\{30,50\}$; samples $\{5,10\}$; link
weight $\{.01,.1,1\}$; bias order $\{1,2,3\}$ and transform
$\{\texttt{sigmoid},\texttt{identity}\}$; activation and jumping-knowledge
booleans; learning rate $\{10^{-5},10^{-4},10^{-3},10^{-2}\}$;
weight decay $\{0,5\!\times\!10^{-4},5\!\times\!10^{-3},.05\}$ \\

Polynormer &
hidden 64; 7 local and 2 global layers; one head; $\beta=-1$; input,
model, and global dropout .15/.5/.5; 100 local epochs; learning rate
$10^{-3}$; weight decay $5\!\times\!10^{-4}$ &
hidden $\{64,128,256,512\}$; local layers $\{5,7,10\}$; global layers
$\{1,2,3,4\}$; heads $\{1,2,4,8\}$; $\beta\in\{-1,.1,.5,.9\}$;
input dropout $\{0,.15,.2,.5\}$; model dropout $\{.2,.3,.5,.7\}$;
global dropout $\{.3,.5\}$; pre-layernorm boolean; local epochs $\{100,200\}$;
learning rate $\{3\!\times\!10^{-5},5\!\times\!10^{-4},10^{-3}\}$;
weight decay $\{0,5\!\times\!10^{-5},5\!\times\!10^{-4}\}$ \\

SGC &
feature dropout 0; $K=2$ &
feature dropout $\{0,.2,.5\}$; $K\in\{1,\ldots,8\}$; shared training \\

SGFormer &
hidden 64; graph weight .5; transformer/GNN layers 1/2; both dropouts 0;
no initial-feature reuse &
hidden $\{64,128,256,512\}$; graph weight $\{.5,.8\}$; transformer and
GNN dropout $\{0,.3,.5\}$; GNN layers $\{1,2,4,8\}$; initial-feature-reuse
boolean; shared training \\

\midrule
G2T-FM &
TabPFNv2 backend; zero fine-tuning epochs &
one canonical target configuration \\

GraphAny &
evaluated arXiv checkpoint and canonical analytical adaptation &
one canonical target configuration \\

GraphPFN &
PCA enabled at 64 dimensions &
PCA disabled or 64 dimensions \\

GVT &
RGVT depth 8; hidden-128 ReLU head without dropout; Adam; learning rate .005; weight decay 0; patience 200 &
encoder depth $\{1,\ldots,8\}$; a separate head is trained at each depth \\

Node4All &
frozen CGT encoder; hidden-512 MLP head; learning rate $10^{-3}$;
weight decay 0; patience 200 &
one encoder/head configuration; validation selects the head epoch \\

NodePFN &
TSVD; 10 components; randomized solver; smoothing 2; ensemble 1 &
reduction $\{\texttt{none},\texttt{tsvd}\}$; components
$\{10,15,20,25,50\}$; TSVD solver
$\{\texttt{arpack},\texttt{randomized}\}$; smoothing $\{0,\ldots,4\}$;
ensemble $\{1,4,8,16,32\}$ \\
\bottomrule
\end{tabularx}
\end{table*}

\mypar{Reading the optimization outcomes.}
Table~\ref{tab:optimization-outcomes} summarizes how many trials
were attempted and completed, and which configurations were selected.
An evaluation is counted as successful when it has at least one
finite validation result and a stored output from the selected
configuration.

The \emph{default-selection rate} is the percentage of successful
evaluations that select trial~1.
The \emph{median trial identifier} summarizes the identifiers of
the selected trials.
The \emph{modal selected settings} give the most frequently selected
value of each method-specific search dimension.

\begin{table*}[!ht]
\centering
\caption{\textbf{Optimization execution and selected-configuration summaries.}
Trial records are successful/attempted totals. Cell coverage is measured
against 510 possible dataset--split--index cells per method. Default selected
is the percentage of successful cells whose selected trial identifier is 1.
For one-configuration methods this is necessarily 100\%. The last column gives
the modal selected values of method-specific search dimensions.}
\label{tab:optimization-outcomes}

\scriptsize
\setlength{\tabcolsep}{2.2pt}
\renewcommand{\arraystretch}{1.03}

\begin{tabularx}{\textwidth}{
    @{}
    >{\raggedright\arraybackslash}p{0.10\textwidth}
    r r r r
    >{\raggedright\arraybackslash}X
    @{}
}
\toprule
Method & Trial records & Cells & Default selected & Median ID
& Modal selected settings \\
\midrule

APPNP & 95,403/102,510 & 510/510 & 2.5\% & 101.5 &
hidden 512; feature drop .7; $K=5$; $\alpha=.5$; propagation drop 0 \\

FAGCN & 102,510/102,510 & 510/510 & 0.2\% & 104.5 &
hidden 128; feature drop .7; 4 layers; $\epsilon=.5$; edge drop .3 \\

GAT & 102,510/102,510 & 510/510 & 0.2\% & 97 &
hidden 64; 4 heads; feature/attention drop .3/.2; 1 layer;
layer norm; residual on; pre-transform off \\

GATv2 & 92,862/92,862 & 462/510 & 0.0\% & 99 &
hidden 64; 4 heads; feature/attention drop .3/.2; 1 layer;
layer norm; residual on; pre-transform off \\

GCN & 102,510/102,510 & 510/510 & 1.0\% & 102 &
hidden 512; feature drop .3; 2 layers; layer norm; residual on;
pre-transform off \\

GCNII & 100,500/100,500 & 500/510 & 3.4\% & 99.5 &
hidden 512; feature drop .7; 32 layers; $\alpha=.5$; $\theta=.5$;
layer norm; residual and pre-transform off \\

GPRGNN & 102,510/102,510 & 510/510 & 0.6\% & 106.5 &
hidden 512; feature drop .7; propagation drop 0; $K=5$;
$\alpha=.9$; random initialization \\

GraphSAGE & 102,510/102,510 & 510/510 & 0.4\% & 97 &
hidden 512; feature drop .3; 1 layer; layer norm; residual and
pre-transform off \\

LINKX & 102,510/102,510 & 510/510 & 0.2\% & 97.5 &
hidden 128; feature drop .7; 1 total, 1 adjacency, and 1 feature layer \\

MLP & 102,510/102,510 & 510/510 & 0.4\% & 101 &
hidden 512; feature drop .7; 1 layer; layer norm; identity skip off \\

NAGphormer & 99,286/100,500 & 500/510 & 8.2\% & 91.5 &
hidden 512; dropout .1; 7 hops; 1 layer \\

NodeFormer & 78,984/82,410 & 410/510 & 0.0\% & 109 &
hidden 128; dropout .5; 50 random features; 10 samples; 3 layers;
2 heads; link weight .1; bias order 3 with identity transform \\

Polynormer & 102,283/102,510 & 510/510 & 0.0\% & 100 &
hidden 512; 5 local and 4 global layers; 8 heads; $\beta=.9$;
input/model/global drop .5/.2/.5; 200 local epochs; pre-LN on \\

SGC & 102,510/102,510 & 510/510 & 3.5\% & 86.5 &
feature drop .5; $K=2$ \\

SGFormer & 102,510/102,510 & 510/510 & 0.0\% & 109 &
hidden 64; graph weight .8; transformer/GNN drop .5/.3;
8 GNN layers; initial-feature reuse off \\

\midrule

G2T-FM & 370/370 & 370/510 & 100.0\% & 1 &
canonical configuration \\

GraphAny & 500/500 & 500/510 & 100.0\% & 1 &
canonical arXiv configuration \\

GraphPFN & 860/1,020 & 510/510 & 84.7\% & 1 &
PCA enabled at 64 dimensions \\

GVT & 4,000/4,000 & 500/510 & 9.6\% & 4 &
encoder depth 1 \\

Node4All & 510/510 & 510/510 & 100.0\% & 1 &
canonical frozen encoder and MLP head \\

NodePFN & 78,156/90,450 & 450/510 & 0.2\% & 108 &
TSVD; 15 components; ARPACK solver; smoothing 0; ensemble 32 \\

\bottomrule
\end{tabularx}
\end{table*}

\FloatBarrier
\section{Evaluation and Computational Accounting}
\label{app:evaluation-accounting}

This appendix explains how we obtain the predictive scores and
computational costs reported in the paper.
We describe how data are split, how models are selected, how results
are combined across datasets, and what is included in each cost
measurement.
We then define the quantities used to compare predictive performance
against computational cost.

\subsection{Splits and Randomness}
\label{app:splits-randomness}

\mypar{Shared stratified splits.}
Each dataset is evaluated under 10/10/80 and 50/25/25
train/validation/test ratios, with five fixed stratified partitions
per ratio.
All methods load and validate the same stored partitions.
We first separate the training nodes, then divide the remaining
labeled nodes into validation and test sets, preserving class
proportions through stratification.

Only nodes with valid labels are eligible for splitting.
We initially require at least two nodes per class and increase
this threshold only if a valid stratified partition cannot be
constructed.
Nodes from excluded classes remain in the graph but are excluded
from all labeled masks; retained classes are remapped to consecutive
integer labels.
If no valid partition can be constructed, the run stops with an error.

\mypar{Randomness control.}
We use deterministically derived seeds for splitting, representation
construction, training, and hyperparameter search, with trial-specific
seeds for additional supervised configurations.
Split seeds depend on the dataset, label ratio, and experimental
index; training and search seeds are shared across ratios for a
fixed dataset and index.
GFM adapters follow their respective interfaces for inference
randomness.
Python, NumPy, and PyTorch random generators are seeded, although
some CUDA operations may prevent bitwise-identical results.

\subsection{Predictive Metrics and Model Selection}
\label{app:metric-selection}

\mypar{Validation selects the model; test data measure its performance.}
For trainable methods, we first select the checkpoint with the
lowest validation loss within each trial.
We then compare configurations using the same validation loss.
The loss is cross-entropy for multiclass tasks and binary
cross-entropy for binary tasks.
No test metric is used to select either the checkpoint or the
configuration.

The retrospective tuning analysis in
Appendix~\ref{app:tuning-robustness} follows the same separation:
it reports the test performance of validation-selected configurations
at different budgets, without using test performance to choose them.

\mypar{Primary and additional test metrics.}
Accuracy is the primary metric for multiclass tasks, and AUROC
is the primary metric for binary tasks.
The additional metrics describe other aspects of the same selected
predictions; they do not trigger separate model selection.
For binary tasks, class index~1 is the positive class, including
when computing positive-class precision.
Table~\ref{tab:appendix-predictive-metrics} lists the metrics and
their corresponding selection losses.

\begin{table*}[t]
\centering
\caption{\textbf{Primary and secondary predictive metrics.} Selection always minimizes validation cross-entropy or binary cross-entropy; primary denotes the headline test metric, not the selection objective.}
\label{tab:appendix-predictive-metrics}
\footnotesize
\setlength{\tabcolsep}{5pt}
\renewcommand{\arraystretch}{1.15}
\begin{tabularx}{\textwidth}{l l X X}
\toprule
task & primary & secondary & selection loss \\
\midrule
Multiclass & Accuracy & Macro-F1; cross-entropy & Validation cross-entropy \\
Binary & AUROC & Accuracy; class-1 precision; binary cross-entropy & Validation binary cross-entropy \\
\bottomrule
\end{tabularx}
\end{table*}

\subsection{Aggregation, Missing Results, and Uncertainty}
\label{app:aggregation-imputation}

\mypar{Mean ranks summarize relative performance across datasets.}
For each dataset, we first average a method's score over the five
partitions.
We then rank methods within that dataset, assigning average ranks
to ties.
Overall mean ranks weight each dataset equally.
A subgroup's mean rank is calculated in the same way, using only
the datasets belonging to that subgroup.

Figure~\ref{fig:gfm-groupwise-ranks} uses each dataset's primary
metric: AUROC for binary tasks and accuracy for multiclass tasks.
All 21 methods enter the ranking, although the figure displays
only GCNII, GPRGNN, and GraphPFN.
The 22 subgroups are defined by seven properties in
Appendix~\ref{app:dataset-definitions}.
Datasets can appear in several subgroups, but each subgroup is
summarized separately; subgroup results are not averaged again
across properties.

\mypar{Missing results are replaced only in the imputed comparisons.}
Comparisons over the full dataset collection require the same
methods to be represented on every dataset.
When a built-in method lacks a required result, the imputed
analysis uses the GCN Default result from the same dataset,
label ratio, and partition.
Affected results are marked as imputed.

Replacement occurs before computing ranks, pairwise outcomes,
best-error references, or bootstrap samples.
The pretraining-overlap policy also replaces GraphAny and GVT
results on \texttt{ogbn\_arxiv}.
User submissions are never imputed.
Table~\ref{tab:coverage-summary} reports coverage before replacement,
showing which methods require imputation.

\mypar{Elo summarizes pairwise wins rather than averaging raw scores.}
Elo combines results across datasets by asking which method wins
each pairwise comparison.
This avoids directly averaging accuracy and AUROC values from
different tasks.
For each dataset and partition, we compare test AUROC on binary
tasks and test accuracy on multiclass tasks.
The higher score wins; an exact tie contributes half a win.
Each partition-level comparison has weight $1/5$, so the five
partitions together give each dataset equal weight.

We fit the pooled outcomes with a logistic Bradley--Terry model
without an intercept.
The fit uses weak L2 regularization, with $C=10^6$, and at most
1,000 iterations.
The fitted method strengths are converted to Elo ratings
satisfying
\[
P(m>n)=\frac{1}{1+10^{(R_n-R_m)/400}}.
\]
Thus, a larger rating difference corresponds to a larger estimated
probability that one method outperforms the other.
We shift all ratings by the same amount so that GCN Default
has rating 1000.
Degenerate bootstrap fits use a deterministic iterative fallback.

Unlike mean-rank summaries, Elo compares methods within each
partition before pooling their outcomes.
Mean ranks instead rank the scores obtained after averaging
the five partitions.

\mypar{Confidence intervals and rank distributions show different variation.}
To estimate Elo uncertainty, we resample datasets with replacement
200 times using seed~0.
Each sampled dataset retains all five of its partitions.
The reported 95\% intervals use the 2.5th and 97.5th percentiles,
with the estimator's center correction and rating calibration.

Mean Improvability uses 100 dataset-bootstrap replicates with
seed~0, after averaging the values within each dataset.
By contrast, the interquartile ranges in the rank figures describe
how a method's rank varies across datasets.
They are not bootstrap confidence intervals.

\subsection{Computational Measurement}
\label{app:compute-environment}

The fitting phase differs by evaluation setting:
\begin{itemize}
    \item \textbf{Default supervised method.}
    The fitting time includes preprocessing, representation
    construction, and training the default model.

    \item \textbf{Tuned supervised method.}
    The fitting/search time includes the default and all executed
    HPO trials, their validation evaluations, and configuration
    selection.
    Time spent on failed or skipped trials remains included when
    the search continues.
    This is the cost of the entire search, not just the selected trial.

    \item \textbf{GFM.}
    The corresponding phase measures target-specific preprocessing
    and adaptation, including any head training or configuration
    search performed inside the fitting call.
\end{itemize}

\mypar{GPU synchronization and peak memory.}
CUDA is synchronized immediately before and after each timed
operation, so the wall-clock measurement includes completion
of the GPU work.
For memory, we reset the CUDA peak counter around the operation
and record the maximum allocated memory.

Workflow peak memory is the larger of the fitting/search/adaptation
peak and the prediction peak.
We take their maximum rather than their sum because these stages
run sequentially.
GraphAny fitting is CPU-only, so its CUDA fit-memory value is
not applicable, rather than zero.

\mypar{Hardware and software environment.}
All reported measurements were collected on an NVIDIA H200
with CUDA~12.1, cuDNN~90100, PyTorch~2.4.0, and Python~3.10.20.
Memory accounting covers the designated CUDA device and does not
automatically sum allocations across multiple devices.

\subsection{Mean Improvability, Mean Overhead, and Pareto Frontiers}
\label{app:overhead-pareto}

\mypar{Mean Improvability measures the remaining predictive gap.}
Mean Improvability asks how much of a configuration's error could
be removed by matching the best evaluated configuration on the
same dataset, label ratio, and partition.
Here, error is defined as one minus the primary test score:
AUROC for binary tasks and accuracy for multiclass tasks.
It is not the validation CE or BCE used for model selection.

For primary score $s_{m,d,q,i}$, let
\[
e_{m,d,q,i}=1-s_{m,d,q,i},
\qquad
e^\star_{d,q,i}=\min_{j\in\mathcal M}e_{j,d,q,i}.
\]
The corresponding Improvability is
\[
I_{m,d,q,i}=
\begin{cases}
0, & e_{m,d,q,i}\leq 10^{-12},\\
100\displaystyle\frac{e_{m,d,q,i}-e^\star_{d,q,i}}
{e_{m,d,q,i}}, & \text{otherwise}.
\end{cases}
\]
For example, an Improvability of 50\% means that matching the best
configuration would halve the error defined above.
Lower values are better.

We first average Improvability across the five partitions,
then average equally across datasets.
The comparison set $\mathcal M$ contains all configurations in
the analysis cohort.
The full Pareto comparison includes 36 configurations:
the default and tuned versions of 15 supervised methods,
plus six GFMs.

\mypar{Mean Overhead expresses cost relative to a common reference.}
Raw runtime and memory requirements differ greatly across datasets.
We therefore compare each configuration with the least expensive
eligible reference on the same dataset, label ratio, and
compute environment.
For resource $r$,
\[
O_{m,d,q}^{(r)}
=
\log_2\!\left(
\frac{C_{m,d,q}^{(r)}}
{\min_{j\in\mathcal R}C_{j,d,q}^{(r)}}
\right).
\]
Mean Overhead averages these log-ratios over eligible datasets.

On an individual dataset, an overhead of zero matches the
least expensive reference, and an overhead of one means twice
its cost.
A negative value means that the evaluated configuration is
cheaper than every eligible reference.
Across datasets,
$2^{\mathrm{Mean\ Overhead}}$
is the geometric-mean cost multiplier.

\mypar{The cost-reference pool is fixed.}
The reference pool $\mathcal R$ contains 32 configurations,
rather than all 36 evaluated configurations.
It excludes the default and tuned versions of FAGCN and LINKX,
and includes both endpoints of the other 13 supervised methods
and the six canonical GFMs.

The common-methods analysis uses the intersection of this fixed
pool with the configurations included in that analysis.
Evaluating a new method does not change the reference pool.

\mypar{A Pareto frontier identifies configurations that are not dominated.}
A configuration is Pareto-optimal if no other configuration is
at least as good in both Mean Improvability and Mean Overhead,
and strictly better in one of them.
The frontier therefore contains configurations for which better
predictive performance cannot be obtained at the same or lower
reported cost within the evaluated set.
Lower values are better on both axes.

\mypar{Intermediate tuning budgets combine stored predictions with estimated costs.}
Figure~\ref{fig:tuning-trajectories} evaluates budgets of
\[
b\in\{1,2,4,8,16,32,64,128,201\}.
\]
The predictive results come from stored trials.
Only the default and full-search workflow costs are measured
directly; intermediate costs are interpolated.
The trajectories are constructed as follows.

\begin{enumerate}
    \item \textbf{Construct a trial order.}
    For each method, dataset, label split, and partition,
    trial~1 remains the default configuration.
    Trial identifiers 2--201 are uniformly permuted in each of
    20 deterministic resamples.
    The SHA-256-derived seed includes the protocol, split, method,
    dataset, and partition.
    Within each resample, larger budgets extend the same trial order.

    \item \textbf{Select a configuration at each budget.}
    Budget $b$ includes the first $b$ attempted positions.
    Failed trials consume budget but cannot be selected.
    Among the eligible trials, we select the configuration with
    the lowest validation BCE for binary tasks or validation CE
    for multiclass tasks.
    Ties retain the configuration encountered first in that
    permutation.
    This tie rule is specific to the trajectory analysis;
    the primary HPO procedure follows
    Appendix~\ref{app:search-spaces}.

    \item \textbf{Read its stored test performance.}
    The selected configuration's test AUROC or accuracy determines
    its predictive result at that budget.
    The best-error reference used for Mean Improvability is computed
    from the cohort's complete configuration set and remains fixed
    across budgets.
    In the imputed analysis, a missing full-result record is replaced
    by the matched GCN Default trajectory.

    \item \textbf{Estimate its workflow cost.}
    Let $C_1$ and $C_{201}$ be the measured default and full-search
    workflow costs.
    We estimate the cost at budget $b$ by
    \[
    C_b=C_1+\frac{b-1}{200}(C_{201}-C_1).
    \]
    These values are interpolated between measured endpoints;
    they are not measurements of cumulative trial time.

    \item \textbf{Compute and display the frontier.}
    All 15 supervised methods and six GFM points enter the
    frontier calculation.
    A supervised trajectory is displayed only if at least one
    of its budget points lies on the resulting lower-left frontier.
\end{enumerate}

These trajectories compare aggregate predictive performance with
estimated relative workflow cost.
They are not a directly measured experiment that gives each method
the same wall-clock budget on each dataset.

This analysis is also distinct from the resampled predictive tuning
analysis in Appendix~\ref{app:tuning-robustness}, which uses a different
budget grid.
Table~\ref{tab:app-tuning-efficiency} supplements that analysis with
cost multipliers obtained from the same endpoint-interpolation approach.

\FloatBarrier

\section{Additional Dataset-Level and Configuration Results}
\label{app:complete-results}

This appendix reports per-dataset predictive outcomes, ranks all 36 evaluated configurations, and makes predictive coverage explicit.

\subsection{Dataset-Level Predictive Outcomes}
\label{app:complete-predictive-results}

Table~\ref{tab:per-dataset-primary-winners} reports the leading method on each of the 51 datasets under both label ratios. Candidates are the single report point of each method: the validation-selected tuned configuration for supervised methods and the canonical configuration for GFMs. Supervised defaults are not included. Multiclass
tasks use accuracy and binary tasks use AUROC. Each entry gives the mean and
standard deviation across the five shared partitions, and none of the winning
entries is imputed.

GFMs lead 15 datasets under 10/10/80 and 13 under 50/25/25. GraphPFN accounts
for 14 and 13 of these wins, respectively, while GraphAny contributes one
10/10/80 win. The table therefore shows both sides of the aggregate result:
GraphPFN can be the strongest method on individual datasets, but
supervised methods collectively lead most of the benchmark.

\begin{table}[H]
\centering
\caption{\textbf{Per-dataset primary-metric leaders in the 51-dataset cohort.} Leaders are selected among the single report point of each method: the validation-selected tuned configuration for supervised methods and the canonical configuration for GFMs. Values are mean $\pm$ standard deviation over five shared partitions. M uses multiclass accuracy and B uses binary AUROC. No imputed result is selected as a leader.}
\scriptsize
\setlength{\tabcolsep}{5pt}
\renewcommand{\arraystretch}{1.02}
\begin{tabular}{@{}llp{0.31\textwidth}p{0.31\textwidth}@{}}
\toprule
Dataset & Task & Best, 10/10/80 & Best, 50/25/25 \\
\midrule
actor & M (Acc.) & GraphPFN 38.15$\pm$0.56 & GraphPFN 39.64$\pm$0.34 \\
amazon\_computer & M (Acc.) & GraphPFN 91.09$\pm$0.27 & GraphPFN 93.04$\pm$0.36 \\
amazon\_photo & M (Acc.) & GraphPFN 94.95$\pm$0.33 & GraphPFN 96.25$\pm$0.47 \\
amazon\_ratings & M (Acc.) & GraphPFN 43.78$\pm$0.29 & LINKX 50.63$\pm$1.07 \\
amherst41 & B (AUROC) & LINKX 76.50$\pm$1.36 & LINKX 89.42$\pm$1.73 \\
artnet-exp & B (AUROC) & GraphPFN 84.60$\pm$0.33 & GraphPFN 86.75$\pm$0.27 \\
blogcatalog & M (Acc.) & GCNII 94.00$\pm$0.40 & GPRGNN 96.67$\pm$0.51 \\
chameleon & M (Acc.) & GraphPFN 46.63$\pm$2.17 & GraphPFN 51.03$\pm$3.22 \\
citation\_citeseer & M (Acc.) & GCNII 90.29$\pm$0.81 & GPRGNN 94.95$\pm$0.67 \\
citeseer & M (Acc.) & GPRGNN 73.11$\pm$0.51 & GraphSAGE 77.72$\pm$1.38 \\
city-reviews & B (AUROC) & GraphPFN 93.00$\pm$0.11 & Polynormer 93.92$\pm$0.20 \\
coauthor\_cs & M (Acc.) & GCNII 94.66$\pm$0.24 & GCNII 96.08$\pm$0.23 \\
coauthor\_physics & M (Acc.) & GCNII 96.46$\pm$0.11 & GCNII 97.17$\pm$0.15 \\
cora & M (Acc.) & GPRGNN 85.36$\pm$1.15 & GPRGNN 89.19$\pm$1.06 \\
cora\_ml & M (Acc.) & APPNP 86.78$\pm$0.23 & GPRGNN 90.17$\pm$0.84 \\
cornell & M (Acc.) & NodeFormer 65.58$\pm$3.87 & NodeFormer 80.43$\pm$5.54 \\
cornell5 & B (AUROC) & LINKX 83.00$\pm$0.47 & LINKX 91.65$\pm$0.49 \\
dblp & M (Acc.) & GPRGNN 84.80$\pm$0.31 & GPRGNN 85.83$\pm$0.39 \\
deezer & B (AUROC) & GraphPFN 70.74$\pm$0.53 & GraphPFN 73.32$\pm$0.69 \\
elliptic\_bitcoin & B (AUROC) & GraphPFN 98.22$\pm$0.20 & Polynormer 99.40$\pm$0.18 \\
facebook\_large & M (Acc.) & GCNII 93.20$\pm$0.10 & Polynormer 95.23$\pm$0.25 \\
flickr & M (Acc.) & GraphPFN 53.67$\pm$0.11 & Polynormer 56.92$\pm$0.36 \\
full\_cora & M (Acc.) & GPRGNN 66.84$\pm$0.45 & GPRGNN 72.79$\pm$1.19 \\
genius & B (AUROC) & GraphSAGE 92.06$\pm$0.12 & GraphSAGE 92.33$\pm$0.03 \\
johnshopkins55 & B (AUROC) & LINKX 79.23$\pm$1.69 & LINKX 89.11$\pm$0.61 \\
la & M (Acc.) & Polynormer 45.21$\pm$4.84 & Polynormer 67.22$\pm$5.48 \\
lastfm\_asia & M (Acc.) & FAGCN 87.01$\pm$0.67 & APPNP 89.90$\pm$0.69 \\
london & M (Acc.) & Polynormer 42.47$\pm$4.23 & GraphPFN 61.30$\pm$1.59 \\
ogbn\_arxiv & M (Acc.) & GCNII 71.62$\pm$0.44 & GCNII 74.18$\pm$0.33 \\
paris & M (Acc.) & GraphPFN 46.59$\pm$0.86 & GraphPFN 60.82$\pm$0.89 \\
penn94 & B (AUROC) & GPRGNN 87.13$\pm$0.61 & LINKX 92.89$\pm$0.27 \\
pubmed & M (Acc.) & GraphPFN 90.50$\pm$0.15 & GraphPFN 91.70$\pm$0.33 \\
reed98 & B (AUROC) & GraphAny 59.72$\pm$2.97 & LINKX 72.96$\pm$1.95 \\
shanghai & M (Acc.) & GCNII 46.60$\pm$3.01 & Polynormer 68.59$\pm$4.56 \\
squirrel & M (Acc.) & GraphPFN 43.97$\pm$0.56 & GraphPFN 46.19$\pm$1.20 \\
tag\_bookchild & M (Acc.) & APPNP 58.55$\pm$0.37 & GPRGNN 63.56$\pm$0.58 \\
tag\_bookhis & M (Acc.) & APPNP 85.19$\pm$0.20 & GCNII 86.78$\pm$0.49 \\
tag\_citeseer & M (Acc.) & GPRGNN 77.91$\pm$0.50 & FAGCN 81.93$\pm$1.02 \\
tag\_cora & M (Acc.) & FAGCN 87.34$\pm$0.88 & GPRGNN 90.10$\pm$1.13 \\
tag\_cornell & M (Acc.) & SGFormer 76.34$\pm$2.59 & GCNII 87.50$\pm$3.29 \\
tag\_pubmed & M (Acc.) & FAGCN 92.20$\pm$0.27 & GPRGNN 93.88$\pm$0.31 \\
tag\_sportsfit & M (Acc.) & GraphSAGE 93.05$\pm$0.09 & GPRGNN 94.47$\pm$0.18 \\
tag\_texas & M (Acc.) & SGFormer 85.73$\pm$2.61 & FAGCN 95.32$\pm$4.09 \\
tag\_washington & M (Acc.) & GPRGNN 79.67$\pm$2.54 & SGFormer 87.93$\pm$2.73 \\
tag\_wikics & M (Acc.) & GAT 85.07$\pm$0.56 & GraphPFN 88.24$\pm$0.44 \\
tag\_wisconsin & M (Acc.) & SGFormer 85.16$\pm$3.16 & SGFormer 92.54$\pm$2.36 \\
texas & M (Acc.) & NodeFormer 74.25$\pm$2.40 & GraphSAGE 87.39$\pm$4.96 \\
tolokers-2 & B (AUROC) & GraphPFN 85.83$\pm$0.36 & GraphPFN 87.67$\pm$0.34 \\
wiki & M (Acc.) & GCNII 78.76$\pm$1.44 & GCNII 85.81$\pm$1.62 \\
wiki\_cs & M (Acc.) & GAT 81.85$\pm$0.69 & GraphPFN 85.48$\pm$0.25 \\
wisconsin & M (Acc.) & NodeFormer 78.31$\pm$2.10 & NodeFormer 87.94$\pm$4.71 \\
\bottomrule
\end{tabular}
\label{tab:per-dataset-primary-winners}
\end{table}

\subsection{Configuration-Level Rankings}
\label{app:configuration-results}

Table~\ref{tab:configuration-ranking} expands the method comparison to all 36
default, tuned, and canonical configurations in Cohort V1. Rankings use
accuracy on multiclass datasets and AUROC on binary datasets. Reporting
default and tuned endpoints separately makes the contribution of optimization
visible and prevents a tuned method from being treated as interchangeable with
its default implementation.

Under 10/10/80, tuned supervised methods occupy the first eight
positions, and GraphPFN is the highest-ranked GFM in ninth. Under 50/25/25,
tuned GNNs occupy the first four positions and GraphPFN moves to fifth.
Optimization changes the ordering substantially. GPRGNN moves from ranks 15
and 16 in its default form to rank 2 under both label ratios, while GCNII moves
from ranks 19 and 27 to rank 1. Polynormer similarly moves from ranks 35 and 24
to ranks 12 and 8. These shifts explain why default-only comparisons would
understate the strongest supervised baselines.

\begin{table}[H]
\centering
\caption{\textbf{Primary-metric ranks of all 36 default, tuned, and canonical configurations in Cohort V1.}}
\label{tab:configuration-ranking}
\scriptsize
\setlength{\tabcolsep}{4.0pt}
\renewcommand{\arraystretch}{1}
\begin{tabular}{@{}lrrrr@{}}
\toprule
Configuration & \shortstack{Mean rank\\10/10/80} & \shortstack{Mean rank\\50/25/25} & \shortstack{Rank\\10/10/80} & \shortstack{Rank\\50/25/25} \\
\midrule
GCNII [hp\_tuned] & 6.4216 & 7.1667 & 1 & 1 \\
GPRGNN [hp\_tuned] & 6.8039 & 7.3333 & 2 & 2 \\
FAGCN [hp\_tuned] & 8.2353 & 8.4608 & 3 & 3 \\
GCN [hp\_tuned] & 9.9216 & 9.1373 & 4 & 4 \\
GAT [hp\_tuned] & 10.8431 & 10.7157 & 5 & 9 \\
GraphSAGE [hp\_tuned] & 11.2843 & 9.8627 & 6 & 7 \\
APPNP [hp\_tuned] & 11.5882 & 13.5686 & 7 & 11 \\
SGFormer [hp\_tuned] & 12.1373 & 9.7745 & 8 & 6 \\
GraphPFN [canonical] & 12.5 & 9.4412 & 9 & 5 \\
GATv2 [hp\_tuned] & 13.1569 & 11.2451 & 10 & 10 \\
GVT [canonical] & 13.3137 & 14.402 & 11 & 12 \\
Polynormer [hp\_tuned] & 13.6176 & 10.3137 & 12 & 8 \\
NodeFormer [hp\_tuned] & 14 & 15.7549 & 13 & 15 \\
NAGphormer [hp\_tuned] & 14.6373 & 15.1078 & 14 & 14 \\
GPRGNN [default] & 17.6373 & 17.7353 & 15 & 16 \\
LINKX [hp\_tuned] & 18.5294 & 14.902 & 16 & 13 \\
FAGCN [default] & 18.6471 & 21.1765 & 17 & 20 \\
NAGphormer [default] & 19.7941 & 18.7451 & 18 & 17 \\
GCNII [default] & 20.0294 & 24.5098 & 19 & 27 \\
Node4All [canonical] & 20.7843 & 20.9902 & 20 & 19 \\
SGC [hp\_tuned] & 20.9804 & 23.0784 & 21 & 22 \\
APPNP [default] & 21.0294 & 24.5294 & 22 & 28 \\
GraphSAGE [default] & 21.6765 & 20.5392 & 23 & 18 \\
SGFormer [default] & 22.3039 & 21.2843 & 24 & 21 \\
GCN [default] & 22.4216 & 23.5196 & 25 & 25 \\
MLP [hp\_tuned] & 22.8431 & 23.1765 & 26 & 23 \\
G2T-FM [canonical] & 23.4608 & 23.598 & 27 & 26 \\
NodeFormer [default] & 23.4902 & 24.5784 & 28 & 29 \\
NodePFN [canonical] & 25.2843 & 24.9216 & 29 & 31 \\
GATv2 [default] & 26.1275 & 25.2549 & 30 & 32 \\
MLP [default] & 26.1471 & 26.9608 & 31 & 34 \\
SGC [default] & 26.7353 & 30.1373 & 32 & 36 \\
GAT [default] & 26.7941 & 26.5196 & 33 & 33 \\
GraphAny [canonical] & 26.8235 & 29.3431 & 34 & 35 \\
Polynormer [default] & 28 & 23.451 & 35 & 24 \\
LINKX [default] & 28 & 24.7647 & 35 & 30 \\
\bottomrule
\end{tabular}
\end{table}

\FloatBarrier

\subsection{Predictive Coverage}
\label{app:run-status}

Thirteen of the 21 methods have native results for all 51 datasets under both
label ratios. Table~\ref{tab:coverage-summary} reports the remaining eight
methods, for which at least one dataset uses a GCN-Default replacement. A
dataset is counted as imputed when any of its five partitions is replaced, so
the table makes each method's native contribution to Cohort V1 explicit.

\begin{table}[H]
\centering
\caption{\textbf{Methods requiring at least one GCN-Default replacement in Cohort V1.} All unlisted methods have native results on all 51 datasets under both label ratios.}
\label{tab:coverage-summary}
\scriptsize
\setlength{\tabcolsep}{5pt}
\renewcommand{\arraystretch}{0.96}
\begin{tabular}{@{}lrrrr@{}}
\toprule
Method & \shortstack{Observed 10/10/80} & \shortstack{Imputed 10/10/80} & \shortstack{Observed 50/25/25} & \shortstack{Imputed 50/25/25} \\
\midrule
GATv2 & 46 & 5 & 46 & 5 \\
GCNII & 50 & 1 & 50 & 1 \\
NAGphormer & 50 & 1 & 50 & 1 \\
NodeFormer & 41 & 10 & 41 & 10 \\
G2T-FM & 44 & 7 & 30 & 21 \\
GraphAny & 50 & 1 & 50 & 1 \\
GVT & 50 & 1 & 50 & 1 \\
NodePFN & 48 & 3 & 42 & 9 \\
\bottomrule
\end{tabular}
\end{table}

\FloatBarrier

\end{document}